%% file: paper_final_vk.tex
\documentclass[twoside,11pt]{article}

\usepackage[abbrvbib]{jmlr2e}
 
\usepackage{url}            
\usepackage{booktabs}       
\usepackage{amsfonts}       
\usepackage{nicefrac}       
\usepackage{microtype}      
\usepackage{wrapfig,verbatim}

\usepackage{varioref}

\usepackage[cmex10,tbtags]{amsmath}
\DeclareMathOperator*{\argmax}{argmax}
\DeclareMathOperator*{\argmin}{argmin}
\usepackage{cases}
\usepackage{amsfonts}
\usepackage{bbm}

\newtheorem{protocol}{Protocol}
\usepackage{paralist,xparse}
\usepackage{amsfonts}
\usepackage{multirow} 
\usepackage{subfigure}
\usepackage{amssymb} 
\usepackage{scrextend}
\usepackage{float}
\usepackage{adjustbox}

\usepackage{algorithm}
\usepackage{algorithmic}
\usepackage{mathtools}
\usepackage{times}
\usepackage{pgfplots}
\usepackage{array}

\usepackage{enumitem}

\newcounter{assumindex}
\input{macrodefs.tex}

\usepackage{mathtools}

\DeclarePairedDelimiter\floor{\lfloor}{\rfloor}

\makeatletter
\def\@listiii{\leftmargin\leftmarginiii
               \labelwidth\leftmarginiii
               \advance\labelwidth-\labelsep
               \topsep\z@
               \parsep\z@
               \partopsep\z@
               \itemsep\topsep}
\makeatother

\usepackage{lastpage}
\jmlrheading{24}{2023}{1-\pageref{LastPage}}{10/20; Revised
12/22}{2/23}{20-1202}{Kunal Pattanayak and Vikram Krishnamurthy}
\ShortHeadings{Inverse Reinforcement Learning for Optimal Bayesian Stopping}{Pattanayak and Krishnamurthy}

\pgfplotsset{compat=1.16}
\begin{document}
\title{Necessary and Sufficient Conditions for Inverse Reinforcement Learning of Bayesian Stopping Time Problems
	\thanks{A short version of partial results has appeared in the Proceedings of the International  Conference on  Information  Fusion, July  2020.}
}
\author{
\name Kunal Pattanayak \email kp487@cornell.edu \\
       \addr Electrical and Computer Engineering\\
       Cornell University\\
       Ithaca, NY $14853$, USA
       \AND
       \name Vikram Krishnamurthy
       \email vikramk@cornell.edu \\
       \addr Electrical and Computer Engineering\\
       Cornell University\\
       Ithaca, NY $14853$, USA
       }

                
\editor{Andreas Krause}

\maketitle

\begin{abstract} This paper presents an inverse reinforcement learning~(IRL) framework for Bayesian stopping time problems. By observing the actions of a Bayesian decision maker, we provide a necessary and sufficient condition to identify if these actions are consistent with optimizing a cost function. In a Bayesian (partially observed) setting, the inverse learner can at best identify optimality wrt the observed strategies. Our IRL algorithm identifies optimality and then constructs set-valued estimates of the cost function.
To achieve this IRL objective, we use novel ideas from Bayesian revealed preferences stemming from microeconomics. We illustrate the proposed IRL scheme using two important examples of stopping time problems, namely, sequential hypothesis testing and Bayesian search. As a real-world example, we illustrate using a YouTube dataset comprising metadata from 190000 videos how the proposed IRL method predicts user engagement in online multimedia platforms with high accuracy. Finally, for finite datasets, we propose an IRL detection algorithm and give finite sample bounds on its error probabilities.

\end{abstract}

\acks{This research was funded in part the U. S. Army Research Office under grant W911NF-21-1-0093, National Science Foundation under grant CCF-2112457, and Air Force Office of Scientific Research under grant FA9550-22-1-0016}

\section{Introduction} \label{sec:introduction}
In a stopping time problem, a decision maker obtains noisy observations of a random variable (state of nature) $\state$  sequentially
over time. Based on the  observation history (sigma-algebra generated by the observations), the decision maker decides at each time whether to continue or stop. If the decision maker chooses the continue action, it pays a continuing cost and obtains the next observation.
If the decision maker chooses the stop action at a specific time, then the problem terminates, and the decision maker pays a stopping cost. In a {\em Bayesian} stopping time problem, the decision maker knows the prior distribution of state of nature $\state$ and the observation likelihood (conditional distribution of the observations) $p(y|\state)$ given the state $\state$, and uses this information to update its belief and choose its continue and stop actions.
Finally, in an {\em optimal} Bayesian stopping time problem, the decision maker chooses its continue and stop actions to minimize an expected cumulative cost function.

Inverse reinforcement learning (IRL) aims to estimate the costs/rewards of a decision maker by observing its actions and was first studied by \citet{NG00} and \citet{AB04}. This paper considers IRL for Bayesian  stopping time problems.
Suppose an inverse learner observes the actions of a decision maker  performing Bayesian sequential stopping in {\em multiple environments}. The decision maker has a fixed observation likelihood and observation cost, and incurs a different stopping cost in each environment\footnote{We refer the reader to \citet[Ch.\,3.5]{RUST94} and \citet{ROL22} for motivating the need to for multiple environments for identifiability of Markov decision processes (MDPs).}. The inverse learner does not know the realizations of the observation sequence nor the observation likelihood of the decision maker; the inverse learner only knows the true state $\state$ and observes the stopping action $\action$ of the decision maker. The two main questions we address are:
\begin{compactenum}
\item How can the inverse learner check if the actions of a Bayesian decision maker are consistent with optimal stopping? 
\item If the decision maker's actions are consistent with optimal stopping, how can the inverse learner estimate the stopping costs of the multiple environments?
\end{compactenum}

\subsection{Main results and context}

The key results in this paper are  summarized as follows:\\
{\bf 1. Inverse RL for Bayesian sequential stopping:} 
Theorem~\ref{thrm:NIAS_NIAC} in Sec.\,\ref{sec:BRP_RI} is our first key IRL result. Theorem~\ref{thrm:NIAS_NIAC} specifies a set of convex inequalities that are simultaneously necessary and sufficient for the actions of a Bayesian decision maker in multiple environments to be consistent with optimal stopping. If so, then Theorem~\ref{thrm:NIAS_NIAC} provides an algorithm for the inverse learner to generate set-valued estimates of the decision maker's costs in the multiple environments. Theorem~\ref{thrm:NIAS_NIAC} is especially useful in scenarios where the inverse learner has no knowledge of the decision maker's observation likelihood or observation sample paths, and yet can construct a set-valued estimate of the costs incurred by the decision maker.

\noindent{\bf 2. Inverse RL for SHT and Search: }
Sec.\,\ref{sec:SHT} and Sec.\,\ref{sec:Search} construct IRL algorithms for two specific examples of Bayesian stopping time problems, namely, Sequential Hypothesis Testing (SHT) and Search. The main results, Theorem~\ref{thrm:classic_SHT} and Theorem~\ref{thrm:Search} specify necessary and sufficient conditions for the decision maker's actions to be consistent with optimal SHT and optimal search, respectively. If the conditions hold, Theorems~\ref{thrm:classic_SHT} and \ref{thrm:Search} provide algorithms to estimate the incurred misclassification costs (for SHT) and search costs (for Bayesian search). In Sec.\,\ref{sec:SHT} for inverse SHT, we also propose an IRL algorithm to compute a point-estimate of the decision maker's costs. The point-estimate is computed by maximizing the regularized margin of the convex feasibility test for inverse SHT proposed in Theorem~\ref{thrm:classic_SHT} and estimates the misclassification costs with up to $95\%$ accuracy. Also, in Sec.\,\ref{sec:comparison}, we compare numerically the performance of the IRL algorithm in Theorem~\ref{thrm:NIAS_NIAC} with two existing IRL algorithms~\citep{CH11} in the literature. This numerical comparison highlights how the IRL approach in this paper complements the results of \cite{CH11}. \blue{Theorem~\ref{thrm:classic_SHT} achieves IRL when the inverse learner has partial information about the decision maker's costs.
}

\noindent 
{\bf 3. Illustration of Inverse RL for Bayesian stopping on Real-World Dataset:} One important use case of IRL is to extract preferences from expert human agents~\citep{LLL14,GJS16}. In Sec.\,\ref{sec:real-world}, we illustrate how our IRL algorithms extend to predicting human-level online multimedia user engagement using a massive YouTube dataset comprising video metadata from approximately $190000$ videos.\footnote{Although understanding YouTube commenting behavior was the main focus of our previous work~\citep{HKP20}, the inference methodology and numerical experiments in this paper are new; see Appendix~\ref{appdx:youtube} and \url{https://github.com/KunalP117/YouTube-Commenting-Analysis} for details.} From the set of costs that pass the convex feasibility test in Theorem~\ref{thrm:NIAS_NIAC} for optimal Bayesian stopping, we chose two point-valued IRL costs for IRL prediction, namely, max-margin IRL and entropy-regularized IRL. The main finding is that both  point estimates 
accurately predict YouTube commenting behavior. Also, we observe that the max-margin estimate yields a more accurate prediction compared to the entropy-regularized estimate (in terms of the chi-square and total variation distance).

\noindent{\bf 4. Sample Complexity for IRL}: 
In Sec.\,\ref{sec:finitesample}, we propose IRL detection tests for optimal stopping, optimal SHT and optimal search under finite sample constraints. Theorems~\ref{thrm:finite_robustness}, \ref{thrm:finite_SHT} and \ref{thrm:finite_Search} in Sec.\,\ref{sec:finitesample} comprise our sample complexity results that characterize the robustness of the detection tests by specifying Type-I and Type-II error bounds for the IRL detection tests. To the best of our knowledge, our finite sample complexity results for the IRL detector, namely, the sample size required to achieve a Type-I or Type-II error probability below a specified value for IRL, are novel. \vspace{0.1cm}\\

\noindent The proofs of all theorems are provided in the Appendix. For a practitioner's perspective, our key IRL algorithms are Theorems~\ref{thrm:NIAS_NIAC},~\ref{thrm:classic_SHT} and \ref{thrm:Search}, and finite sample complexity results for IRL error bounds are Theorems~\ref{thrm:finite_robustness},~\ref{thrm:finite_SHT} and \ref{thrm:finite_Search}. The MATLAB codes and the YouTube dataset for our real-world numerical experiment in Sec.\,\ref{sec:real-world} are available on Github and are completely reproducible.

\subsection{Identifiability. Why IRL for a decision maker in multiple environments?}
An important aspect of our IRL framework is that the inverse learner observes the decision maker in multiple environments.\footnote{The inverse learner in this paper can be viewed as a {\em passive} analyst that does not control the environment variables, that is, the agent's stopping costs. An interesting extension of this paper (for future work) is to consider an {\em active} inverse learner that  purposefully adapts the environment variables to minimize IRL detection errors.} The purpose of this section is to motivate this framework.

We consider a decision maker operating over $\numagents$ environments. In each environment, the decision maker solves a stopping time problem with a distinct stopping cost. The decision maker has a fixed observation likelihood (sensor accuracy) and sensing cost (operating cost), where both variables are invariant across multiple environments. Therefore, there are up to $\numagents$ distinct strategies exhibited by the decision maker, one for each environment. Let $J(\stoptime_\agent,\utilitysymbolagent{\agent})$ denote the expected cost incurred by the decision maker when it chooses stopping strategy $\stoptime_\agent$ in environment $\agent$ with stopping cost $\utilitysymbolagent{\agent}$.

Consider now the inverse learner that observes the decision maker. Assume that the inverse learner does not know the stopping costs $\utilitysymbolagent{\agent}$, but only observes\footnote{We are deliberately simplifying the IRL framework here for explanatory reasons. Our main result assumes the inverse learner only observes the actions of the decision maker, and not the strategy.} the set of strategies $\{\stoptime_\agent,\agent=1,2,\ldots,\numagents\}$. To achieve IRL, the inverse learner must first establish if the decision maker's strategy in each environment is consistent with minimizing an expected cost. Equivalently, the inverse learner must check if the expected cost incurred by the decision maker in environment $\agent$ by choosing strategy $\stoptime_\agent$ is less than that incurred by all other (infinitely many) stopping strategies. However, the inverse learner does not observe infinitely many strategies, but only $\numagents$ strategies. Given the decision maker's strategies in $\numagents$, each with a distinct stopping cost, the inverse learner's procedure to identify if the decision maker is optimal or not is defined below:\\
{\bf IRL identifiability of optimal stopping agent.} {\em Consider a Bayesian stopping agent that chooses strategy $\stoptime_\agent$ in environment $\agent$, over multiple environments $\agent=1,2,\ldots,\numagents$. Then, identifying an optimal Bayesian stopping time agent is equivalent to checking if the following inequalities have a feasible solution:
\begin{equation}\label{eqn:relopt_intuition}
     \text{There exists } \utilitysymbolagent{1},\utilitysymbolagent{2},\ldots,\utilitysymbolagent{\numagents}~\text{such that: } J(\stoptime_\agent,\utilitysymbolagent{\agent}) \leq     J(\stoptime_\agenttwo,\utilitysymbolagent{\agent}),~\forall~\agent,\agenttwo.
\end{equation}
Here, $J(\stoptime_\agent,\utilitysymbolagent{\agenttwo})$ is the decision maker's cumulative expected cost when the decision maker chooses strategy $\stoptime_\agent$ and incurs a stopping cost $\utilitysymbolagent{\agenttwo}$.}\\
The solution of the feasibility problem in \eqref{eqn:relopt_intuition} is the set-valued IRL estimate of the stopping costs incurred by the decision maker. The comparison in~\eqref{eqn:relopt_intuition} between the performance of the decision maker's strategy in each environment to the strategies chosen in all other (finitely many) environments is formalized in Lemma~\ref{lem:relativeoptimality} and is achieved by the inverse learner via the IRL procedure in Theorem~\ref{thrm:NIAS_NIAC}. We also refer the reader to the seminal work of \citet[Ch.\,3.5]{RUST94} and \citet{KIM21} on identifiability of MDPs for further justification of multiple environments. The above framework of a Bayesian stopping time agent operating in multiple environments arises in several applications; see Sec.\,\ref{sec:intro_applications} for details.


\subsection{Context. Bayesian revealed preference for IRL}
The formalism used in this paper to achieve IRL is {\em Bayesian revealed preferences} studied in microeconomics by~\citet{MR14,CM15} and \citet{CD15}; see Sec.\,\ref{sec:related_lit} for more details. This Bayesian revealed preference-based approach {\em complements} existing IRL results for partially observed Markov decision processes (POMDP) including \cite{CH11}. This paper considers a subset of POMDPs, namely, Bayesian stopping time problems. Due to the problem structure, we show that our IRL algorithms {\em do not} require knowledge of the observation likelihood of the decision maker and also do not require solving a POMDP. 

We now briefly discuss how the Bayesian revealed preference based IRL approach differs from classical IRL.
\begin{compactenum}
    \item The classical IRL frameworks~\citep{NG00,AB04} assume the observed agent is a reward maximizer (or equivalently, cost minimizer) and then seeks to estimate its cost function.
   The approach in this paper is more fundamental. We first {\em identify} if the decisions of a single decision maker in multiple environments are consistent with optimality and if so, we then generate set-valued estimates of the costs that are consistent with the observed decisions.
    \item Classical IRL  assumes complete knowledge of the decision maker's observation likelihood. We assume the inverse learner only knows the state of nature and the action chosen when the decision maker stops, and does not know its observation likelihoods or the sequence of observation realizations. \blue{Two important scenarios where this situation arises are:\\
    (i) {\em Multimedia Datasets.} 
    In online multimedia datasets such as the YouTube dataset analyzed in Sec.\,\ref{sec:real-world}, it is impossible to know the attention span (observation likelihood) of the online user. All that is available are the online user's actions (interactions such as comments and comment ratings) and the underlying state of nature (video metadata such as viewcount, thumbnail and video description); see also \cite{HKP20}.  \\
    (ii) {\em Adversarial Signal Processing.} In adversarial signal processing and sensing applications, it is not realistic for the inverse learner to know the model dynamics of the agent. An important example is IRL for radars~\citep{VK20}, where the radar is the adversary and so it is impossible to know its sensing modes (observation likelihood); however, the inverse learner records the electromagnetic waveforms (response) emitted by the radar.\\
    Additional applications where only the agent decisions are available for IRL (and not the observation likelihood) include consumer insights and advertisement design research, interpretable ML in smart healthcare and electronic warfare. These are discussed in Appendix~\ref{appdx:context}.}
    \item {\em Algorithmic Issues:} 
    In classical IRL~\citep{AB04}, the inverse learner  solves the Bayesian stopping time problem iteratively for various choices of the cost. This can be computationally prohibitive since it involves stochastic dynamic programming over a belief space which is  PSPACE hard \citep{PP87}.
    The IRL procedure in this paper \blue{does not require solving a POMDP and only requires testing for the feasibility of a set of convex inequalities.}
\end{compactenum}
\blue{For brevity, we discuss related  IRL literature and applications of IRL for Bayesian stopping problems in Appendix~\ref{appdx:context}.}

\section{Identifying optimal Bayesian stopping and reconstructing agent costs}\label{sec:BRP_RI} 
Our IRL framework  comprises a decision maker's actions in a stopping time problem over $\numagents$ environments, and an {\em inverse learner} that observes these actions. This section defines the IRL problem that the inverse learner faces  and then  presents two results regarding the inverse learner: \begin{compactenum}
\item
 {\em Identifying Optimal Stopping}. Theorem~\ref{thrm:NIAS_NIAC} below provides a necessary and sufficient condition for the inverse learner to identify if the Bayesian decision maker chooses its actions as the solution of an optimal stopping problem.
\item {\em IRL for Reconstructing Costs}.
 Theorem~\ref{thrm:NIAS_NIAC} is also constructive. It  shows that the continue and stopping costs of the Bayesian decision maker can be reconstructed by
 solving a convex feasibility problem.
\end{compactenum}
This section provides a complete IRL framework for Bayesian stopping time problems and sets the stage for subsequent sections where we formulate generalizations and examples. 

\subsection{Bayesian stopping agent}\label{sec:Stopping_Time}
A Bayesian stopping time agent  is parametrized by  the tuple \begin{equation}\label{eqn:IRL_tuple_stop} \stoptuple=(\stateset,\prior,\obsset,\actionset,\oprob,\stoptime)
 \end{equation}
 where
\begin{compactitem}
	    \item $\stateset=\{1,2,\ldots \numstates\}$ is a finite  set of states. 
	    \item At time $0$, the true state $\tstate\in\stateset$ is sampled from prior distribution $\prior$.
	    $\tstate$ is unknown to the agent.
	    \item $\obsset\subset\reals$ is the observation space. Given state $\tstate$, the observations $\obs \in \obsset$ have conditional probability density  $\oprob(\obs,\tstate)=\obslike$. 
	    \item $\actionset=\{1,2,\ldots\numactions\}$ is the finite set of stopping actions. 
	   \item Finally,  $\stoptime$ denotes the agent's stopping strategy.  The stopping strategy operates sequentially on a sequence of observations $\obs_1,\obs_2,\ldots$ as discussed below in Protocol~\ref{prtcl_decision}.
	   \end{compactitem}
     \begin{protocol}\label{prtcl_decision} Sequential Decision-making protocol: Assume the agent knows $\stoptuple$. \vspace{-0.4cm}\newline
    	    \begin{enumerate}
    	        \item Generate $\tstate\sim\prior$, at time $\timeinst=0$. Here $\tstate$ is not known to the agent.
    	        \item At time $\timeinst>0,$ agent records observation $\obs_{\timeinst}\sim\oprob(\cdot,\tstate)$. 
    	        \item {\em Belief Update:} 
    	      Let  $\mathcal{F}_{\timeinst}$ denote the sigma-algebra generated by  observations $ \{\obs_1,\obs_2,\ldots \obs_{\timeinst}\}$.
    	        The  agent updates its belief (posterior) $\belief_{\timeinst}(\state)=\prob(\tstate=\state|\mathcal{F}_{\timeinst}),\state\in\stateset$ using Bayes formula as
    	        \begin{equation}\label{eqn:HMM_stopping}
                 \belief_{\timeinst} = \frac{\oprob(y_{\timeinst})\belief_{\timeinst-1}}{\boldsymbol{1}'\oprob(y_{\timeinst})\belief_{\timeinst-1}},
                \end{equation}
                where $\oprob(\obs) = \operatorname{diag}(\{\oprob(\obs,\state),\state\in\stateset\})$.
    	        The belief $\belief_{\timeinst}$ is an $\numstates$-dimensional probability vector
    	        in the $\numstates-1$ dimensional unit simplex 
    	        \begin{equation}\label{eqn:simplex_stopping}
                \Delta(\stateset)\overset{\text{def.}}{=} \{\belief\in\mathbb{R}^{\numstates}_+:\boldsymbol{1}'\belief=1\}.
                \end{equation}
                \item  Choose action $\action_{\timeinst}=\stoptime(\belief_{\timeinst},\timeinst)$ from the set $\actionset\cup\{\text{continue}\}$. If $\action_{\timeinst}\in\actionset,$ then stop, else if $\action_{\timeinst}=\text{continue},$ set $\timeinst=\timeinst+1$ and go to Step 2.
    	    \end{enumerate}
    	    \end{protocol}
 The stopping  strategy $\stoptime$ is a (possibly randomized) time-dependent mapping from the agent's belief at time $\timeinst\in\mathbb{Z}^+$ to the set $\actionset\cup\{\text{continue}\}$ and belongs to $\stoptimeset$, the set of admissible stopping strategies:\blue{
 \begin{equation}\label{eqn:variable_stopping_policy}
\stoptimeset = \{ \stoptime:\Delta(\stateset)\times \posintegers\rightarrow \actionset\cup\{\text{continue}\}\}.
 \end{equation}} We define the random variable  $\funcstop$ as the time when the agent stops and takes a stop action from $\actionset$.
\begin{equation}\label{eqn:def_stop_time}
    \funcstop =\inf\{\timeinst\geq0|~\stoptime(\belief_{\timeinst},\timeinst)\neq\{\text{continue}\}\}.
\end{equation}
Clearly, the set $\{\funcstop=\timeinst\}$ is measurable wrt $\mathcal{F}_t$, the sigma-algebra generated by  observations $ \{\obs_1,\obs_2,\ldots \obs_{\timeinst}\}$. Hence, the random variable $\funcstop$ is adapted to the filtration $\{\mathcal{F}_\timeinst\}_{\timeinst\geq0}$. In the following sub-section, we will introduce costs for the agent's stop and continue actions. We will use $\funcstop$ for expressing the expected cumulative cost of the agent.

To summarize, a Bayesian stopping agent is parameterized by $\stoptuple$ and operates according to Protocol~\ref{prtcl_decision}. Several decision problems such as SHT and sequential search fit this formulation.

\subsection{Optimal Bayesian  stopping  agent in multiple environments} \label{sec:optimal_stopping_time}
So far we have defined a  Bayesian stopping agent.
Our main IRL result is to identify if a \blue{Bayesian stopping agent's behavior in a set of environments $\agentset$ is {\em optimal}. The purpose of this section is to define optimal Bayesian stopping~\citep{BS15} in multiple environments.
}
For identifiability reasons (see assumption \ref{asmp:IRL2} below) we require at least two environments ($\numagents \geq 2$).

\blue{An optimal Bayesian stopping agent in multiple environments is defined by the tuple} 
\begin{equation} \label{eqn:IRL_tuple_compact}
\optstoptuple=(\stoptuple,\agentset,\runcost,\utilitysymbolset,\stoptimeset^\ast).
\end{equation}
In (\ref{eqn:IRL_tuple_compact}),
\begin{compactitem}
\item $\agentset$ is the set of $\numagents$ environments.
\item The parameters $\stateset,\obsset,\actionset,\prior,\obslikesymbol$ in $\stoptuple$~(\ref{eqn:IRL_tuple_stop}) and continue cost $\runcost$ (defined below) are the same for all environments in $\agentset$.
\item $\runcost = \{\runcostinst_{\timeinst}\}_{\timeinst\geq 0}$, $\runcostinst_\timeinst(\state)\in\reals^{+}$ is the continue cost incurred in any environment $\agent\in\agentset$ at time $\timeinst$ given state $\tstate=\state$.
\item $\utilitysymbolset=\{\utilityagent{\agent},\state\in\stateset,\action\in\actionset,\agent\in\agentset\}$, $\utilityagent{\agent}<\infty$ is the cost for taking stop action $\action$ when the state $\tstate=\state$ in the $\agent^{\text{th}}$ environment.
\item   \blue{$\optstoptimeset=\{\optstoptime_{\agent},\agent\in\agentset\}$}
is the set of {\bf optimal} stopping strategies of the Bayesian stopping agent over the set of environments $\agentset$, where the optimality is defined in Definition~\ref{def:absoptimality} below.
In environment $\agent$, the Bayesian stopping agent employs its stopping strategy \blue{$\stoptime_{\agent}^\ast,\agent\in\agentset$} and operates according to Protocol~\ref{prtcl_decision}.
\end{compactitem}
\begin{definition}[Optimal Stopping Strategy]\label{def:absoptimality} For each environment $\agent\in\agentset$, strategy \blue{$\optstoptime_{\agent}$} is optimal for stopping cost $\utilityagent{\agent}$ iff the following conditions hold:
\begin{align}
 \optstoptime_{\agent}(\belief,\funcstop)&= \argmin_{\action\in\actionset} \belief'\utilityvec_{\agent,\action},\label{eqn:opt_stop_act}\\
 \netobjfun(\optstoptime_{\agent},\utilitysymbolagent{\agent})&=\inf_{\stoptime\in\stoptimeset}\netobjfun(\stoptime,\utilitysymbolagent{\agent})\label{eqn:opt_stop_time},\vspace{-0.1cm}
\end{align}
Recall $\boldsymbol{\stoptime}$~\eqref{eqn:variable_stopping_policy} denotes the set of all stopping strategies.
Also $\netobjfun(\stoptime,\utilitysymbolagent{\agent})$ is the expected cumulative cost defined as:\vspace{-0.1cm}
\begin{equation}\label{eqn:opt_stop_objfun}
\begin{split}
&\netobjfun(\stoptime,\utilitysymbolagent{\agent})=\grosspayoff(\stoptime,\utilitysymbolagent{\agent}) + \sumruncostsymbol(\stoptime),\text{ where}\\
&\grosspayoff(\stoptime,\utilitysymbolagent{\agent})=\E_{\stoptime}\left\{ \belief_{\funcstop}'\utilityvec_{\agent,\stoptime(\belief_{\funcstop},\funcstop)}\right\},~\sumruncostsymbol(\stoptime) = \E_{\stoptime}\bigg\{ \sum_{\timeinst=0}^{\funcstop-1} \belief_\timeinst'\runcostinstvec_{\timeinst}\bigg\},~\stoptime\in\stoptimeset.
\end{split}
\end{equation}
$\E_{\stoptime}$ denotes expectation parametrized by $\stoptime$ wrt the probability measure induced by $\obs_{1:\funcstop}$. Also, $\utilityvec_{\action}$, $\runcostinstvec_{\timeinst}$ are the stopping and continue\footnote{Since the continue cost is a positive real, the stopping time $\funcstop$~\eqref{eqn:def_stop_time} is finite a.s. \label{ftnt:unobserved}} cost vectors, respectively, vectorized over states $\state\in\stateset$.
\end{definition}
Definition~\ref{def:absoptimality} is standard  for the optimal strategy in a sequential stopping problem~\citep{VK16}. The optimal strategy naturally decomposes into two steps: choosing whether to continue or stop according to (\ref{eqn:opt_stop_time}); and if the decision is to stop, then  choose a specific stopping action from $\actionset$ according to
(\ref{eqn:opt_stop_act}).
The optimal stopping strategies $\stoptime_{\agent},\agent\in\agentset$ that satisfy the conditions (\ref{eqn:opt_stop_act}), (\ref{eqn:opt_stop_time}) can be obtained by solving a stochastic dynamic programming problem~\citep{VK16}. It is a well-known result~\citep{LVJ87} that the set of beliefs for which it is optimal to stop is convex.

\subsubsection*{Relation to Bayesian contextual bandits}
\blue{For readers familiar with the multi-armed bandit problem, optimal Bayesian stopping can be viewed as an instance of the {\em partially-observed regularized contextual Bayesian bandit problem}; {\em contextual}~\citep{AG13} since the agent faces multiple ground truths $\state$ (context), {\em partially observed}~\citep{KW09} since the agent observes a sequence of noisy measurements of the underlying context $\state$, {\em Bayesian}~\citep{HKZ22} since the agent minimizes its expected cumulative cost per context averaged over all contexts sampled from a prior distribution $\prior$,
and {\em regularized}~\citep{FBP19} since the agent minimizes the sum of expected stopping cost and a regularization term, namely, the expected continue cost. Loosely speaking, this paper addresses the problem of IRL for partially-observed regularized contextual bandits. Although our IRL results are introduced in subsequent sections, we remark here that there is ample scope to extend the results in this paper to typical RL decision frameworks that allow underlying state transitions. At a high level, this can be made possible by constructing feasibility tests in terms of the state-occupancy measure induced by the decision maker's policy in multiple environments.
}


\subsection{IRL for inverse optimal stopping. Main result}
\label{sec:IRL_Stopping}
We now discuss  an inverse learner-centric view of the Bayesian stopping time problem and the main IRL result.
%
Suppose the inverse learner observes the actions of a Bayesian stopping agent in $\numagents$ environments, where each environment is characterized by the stopping costs incurred by the agent. Suppose the agent  performs several  independent trials of Protocol \ref{prtcl_decision} in all $\numagents$ environments. 
We make the following assumptions about the  inverse learner performing IRL.

\begin{enumerate}[label=(A\arabic*)]
\item \label{asmp:IRL1}The inverse learner knows the dataset
\begin{equation}\label{eqn:IRL_tuple}
\begin{split}
  \datainf & =   (\prior,\actselectset),\text{ where }
\actselectset =\{\actselectagent{\agent},\state\in\stateset,\action\in\actionset,\agent\in\agentset\}.
\end{split}
\end{equation}
 In \eqref{eqn:IRL_tuple}, $\actselectagent{\agent}$ is the Bayesian stopping agent's conditional probability of choosing stop action $\action$ at the stopping time given state~$\tstate=\state$ in the $\agent^{\text{th}}$ environment. We  call $ \actselectagent{\agent}$
as the agent's {\em action selection policy}. \\ Note that:\\
(i) The inverse learner does not know the stopping times; it only has access to the conditional density of which stop action $\action$ was chosen given the true state $\tstate$.\\
\blue{(ii) We assume the decision maker visits all states in the support of the prior pmf $\prior$~\eqref{eqn:IRL_tuple} infinitely often. In Sec.\,\ref{sec:finitesample}, we address the case where the decision maker visits the states finitely often and provide IRL performance guarantees via finite sample complexity.}
\item \label{asmp:IRL2} 
\blue{Dataset $\datainf$} is generated by a Bayesian agent acting in at least $\numagents\geq 2$ environments, where each environment has distinct stopping costs.
\setcounter{assumindex}{\value{enumi}}
\end{enumerate}

Both assumptions are discussed below  after the main theorem, but let us make some preliminary remarks at this stage.
\ref{asmp:IRL1} implies the inverse learner observes the
stopping actions chosen by a Bayesian stopping agent in a finite number ($\numagents$) of environments, where the agent performs an infinite number of independent trials of Protocol~\ref{prtcl_decision} in each environment; see discussion in Sec.\,\ref{sec:disc_assumpt} for asymptotic interpretation.
In Sec.\,\ref{sec:finitesample} we will consider finite sample effects  where the inverse learner observes the agent performing a finite number of independent trials of Protocol~\ref{prtcl_decision}. Assumption \ref{asmp:IRL2} is necessary for the inverse optimal stopping problem to be well-posed.\\

\blue{Let $\stoptime_\agent$ denote the policy chosen by the agent in the $\agent^\text{th}$ environment, and $\boldsymbol{\stoptime}_{\agentset}=\{\stoptime_\agent,\agent\in\agentset\}$ denote the set of chosen strategies.}\footnote{Recall that $\stoptime$ is a generic variable of a stopping policy, $\boldsymbol{\stoptime}$ is the space of admissible policies, $\stoptime^\ast_\agent$ is the optimal policy in environment $\agent$ and $\stoptime_\agent$ is a realization of the agent's policy.} The finite assumption on $|\agentset|$ in \ref{asmp:IRL1} imposes a restriction on our IRL task of identifying optimality of a Bayesian stopping agent formalized below:
\begin{lemma}[IRL identifiability of optimal Bayesian stopping agent.]\label{lem:relativeoptimality} 
Given the dataset $\dataset_\numagents$~\eqref{eqn:IRL_tuple}, the inverse learner can identify an
optimal Bayesian stopping agent~\eqref{eqn:IRL_tuple_compact} acting in $\numagents$ environments if and only if \eqref{eqn:opt_stop_act} and the following relaxation of \eqref{eqn:opt_stop_time} holds:
\begin{equation}\label{eqn:relopt_stop_time_continue_cost}
\grosspayoff_{\agent,\agent}  +    \sumruncostsymbol_\agent \leq  \grosspayoff_{\agenttwo,\agent}  +    \sumruncostsymbol_\agenttwo,~\forall \agent,~\agenttwo\in\agentset,~\agent\neq\agenttwo.
\end{equation}
In \eqref{eqn:relopt_stop_time_continue_cost}, $\grosspayoff_{\agenttwo,\agent}=\grosspayoff(\stoptime_\agenttwo,\utilitysymbol_\agent)$ is the expected stopping cost and $\sumruncostsymbol_\agent = \sumruncostsymbol(\stoptime_\agent)$ is the expected cumulative continue cost  for the policy $\stoptime_\agent$ chosen in environment $\agent,~\agent\in\agentset$.
\end{lemma}
\blue{The proof of Lemma~\ref{lem:relativeoptimality} is in Appendix~\ref{appdx:lemrelopt}. Lemma~\ref{lem:relativeoptimality} formalizes the IRL identification procedure of the inverse learner in~\eqref{eqn:relopt_intuition}.} Since the inverse learner only observes the agent's actions from $\numagents$ strategies chosen by the stopping agent, the best the inverse learner can do is check if $\stoptime_\agent$ is optimal 
for environment $\agent$ out of the finite strategies in $\boldsymbol{\stoptime}_\agentset$. \blue{Indeed, the expected stopping cost $\grosspayoff_{\agenttwo,\agent}$ is a function of the policy $\stoptime_{\agenttwo}$. However, in Appendix~\ref{appdx:NIAS_NIAC}, we show how the expected stopping cost can be expressed only in terms of the observed variables in $\dataset_\numagents$, namely, the action selection policies $\{p_\agent(\action|\state)\}_{\agent=1}^\numagents$ of the agent induced by the stopping strategies $\{\stoptime_\agent\}_{\agent=1}^\numagents$.}
This is precisely what Theorem~\ref{thrm:NIAS_NIAC} below achieves when the inverse learner has access to the agent's action selection policies.

\hspace{-5cm}
\blue{
\begin{table}[t]
    \centering
    \begin{tabular}{|m{2.2cm}|m{3.75cm}|m{3.75cm}|m{3.75cm}|}\hline
       & $\sumruncostagent{}$ unknown & $\sumruncostagent{}\in\mathcal{C}$ convex in $p(\action|\state)$ & $\sumruncostagent{}\in\mathcal{C}$ non-convex in $p(\action|\state)$ \\ \hline
     Identifiability &  Absolute Optimality & Absolute Optimality  & Relative Optimality \\ \hline
     Conditions & \eqref{eqn:opt_stop_act},~\eqref{eqn:opt_stop_time} in Def.\,\ref{def:absoptimality} & \eqref{eqn:opt_stop_act},~\eqref{eqn:opt_stop_time} in Def.\,\ref{def:absoptimality} & \eqref{eqn:opt_stop_act},~\eqref{eqn:relopt_stop_time_continue_cost} in Lemma~\ref{lem:relativeoptimality} \\ \hline
     IRL Example & $--$  & Inverse Optimal Stopping with Entropic Running Cost & Inverse SHT (Sec.\,\ref{sec:SHT}) \\ \hline
     Reconstruction & Convex reconstruction \eqref{eqn:reconstructed-RI-cost} & Convex reconstruction \eqref{eqn:reconstructed-RI-cost} & Reconstructed cost for a finite set of strategies/ \\ \hline
    \end{tabular}
    \caption{IRL Identifiability of Optimal Bayesian stopping.}
    \label{tab:identifiability}
\end{table}
}

\noindent {\em Remarks:}\\
\blue{(1) If the analyst does not know {\em a priori} the structure of the expected continue cost in \eqref{eqn:opt_stop_objfun}, then the IRL identifiability can be generalized from testing for relative optimality~\eqref{eqn:opt_stop_act},~\eqref{eqn:relopt_stop_time_continue_cost} to testing for absolute optimality~\eqref{eqn:opt_stop_act},~\eqref{eqn:opt_stop_time} in Definition~\ref{def:absoptimality}. Specifically, we show a certain reconstruction of the expected continue cost (see \eqref{eqn:reconstructed-RI-cost} in Appendix~\ref{appdx:NIAS_NIAC}) ensures if relative optimality~\eqref{eqn:relopt_stop_time_continue_cost} holds, then absolute optimality~\eqref{eqn:opt_stop_time} holds.\\
(2) In contrast to remark (1) above, if the analyst does know a functional form of the expected continue cost, IRL identifiability cannot be improved from testing for relative optimality. One example is IRL for inverse SHT discussed in Sec.\,\ref{sec:SHT} below where the expected continue cost is known to be the expected stopping time of the agent. On a deeper and more subtle level, knowledge of the structure of the expected continue cost imposes an implicit constraint on the reconstructed cost. Ensuring the reconstructed expected continue cost~\eqref{eqn:reconstructed-RI-cost} in Appendix~\ref{appdx:NIAS_NIAC} satisfies this implicit constraint is non-trivial and beyond the scope of this paper.} 

We now present our first main IRL result.  The result specifies a set of inequalities that, given the inverse learner's specifications in assumptions~\ref{asmp:IRL1} and \ref{asmp:IRL2}, are simultaneously {\em necessary} and {\em sufficient} for the inverse learner to identify a Bayesian stopping agent's actions to be optimal in the sense of Lemma~\ref{lem:relativeoptimality}. \blue{For readability, we provide the exact expressions for the feasibility inequalities introduced below after the main theorem.}
\begin{theorem}[IRL for inverse Bayesian optimal stopping~\citep{CD15}]\label{thrm:NIAS_NIAC}
Consider the inverse learner with dataset $\datainf$ (\ref{eqn:IRL_tuple})
obtained from a Bayesian stopping agent's actions over $\numagents$ environments.
 Assume  \ref{asmp:IRL1} and \ref{asmp:IRL2} hold. Then: \newline
{\em 1.} \underline{Identifiability}: The inverse learner can identify if the dataset $\datainf$ is generated by an optimal Bayesian stopping agent, i.e.\,, \eqref{eqn:opt_stop_act} and \eqref{eqn:opt_stop_time}; see Lemma~\ref{lem:relativeoptimality}.\\
{\em 2.} \underline{Existence}:
There exists an optimal stopping agent parameterized by tuple $\optsearchtuple$~\eqref{eqn:IRL_tuple_compact}
, if and only if there exists a feasible solution to the following convex (in stopping costs) inequalities: 
	\begin{align}
	&\text{Find } \utilityagent{\agent}\in\reals_+~\forall \agent\in\agentset \text{ s.t.}\nonumber\\
	&\operatorname{NIAS}(\datainf,\{\utilityagent{\agent},\state\in\stateset,\action\in\actionset,\agent\in\agentset\}) \leq 0\label{eqn:NIAS_thrm_CD},\\
	&\operatorname{NIAC}(\datainf,\{\utilityagent{\agent},\state\in\stateset,\action\in\actionset,\agent\in\agentset\}) \leq 0 \label{eqn:NIAC_thrm_CD}.
	\end{align}
	The NIAS  (No Improving Action Switches) and NIAC (No Improving Action Cycles) inequalities are defined in (\ref{eqn:NIAS_def}), (\ref{eqn:NIAC_def}) below, and are convex in the stopping cost $\utilityagent{\agent},\agent\in\agentset$.\newline
	{\em 3.} \underline{Reconstruction of costs}:\\
{\em (a)} \blue{If the inverse learner knows the agent's expected continue cost $\sumruncostagent{\agent}$ for all environments $\agent$, the set-valued IRL estimate of the agent's stopping costs is the set of all feasible costs $\{\utilityagent{\agent},\agent\in\agentset\}$ that satisfy the NIAS~\eqref{eqn:NIAS_thrm_CD}, NIAC~\eqref{eqn:NIAC_thrm_CD} and SUMCOST inequalities below:}
\begin{equation}
\label{eqn:feasibilitycontinuecost}
    \sumcost(\datainf,\{\utilityagent{\agent},\sumruncostagent{\agent},\agent\in\agentset\})\leq 0,
\end{equation}
\blue{where $\sumcost$ is defined in (\ref{eqn:sumcostfeasible}), and $\sumruncostagent{\agent}$ is the expected cost of the Bayesian stopping agent in environment $\agent$.}\\
\blue{{\em (b)} Suppose the inverse learner knows the agent's stopping costs, and the NIAS~\eqref{eqn:NIAS_thrm_CD} and NIAC~\eqref{eqn:NIAC_thrm_CD} inequalities are feasible. Then, the set-valued IRL estimate of the agent's expected continue cost is given by the set of all feasible costs $\sumruncostagent{\agent}$ that satisfy the SUMCOST inequality~\eqref{eqn:feasibilitycontinuecost}. Also, if the inverse learner knows the agent's expected continue cost is convex, then the SUMCOST inequality structure permits a convex reconstruction of the cost outlined in Definition~\ref{def:NIAC_ineq}.\hfill \qedsymbol
}
\end{theorem}

Theorem~\ref{thrm:NIAS_NIAC} is proved in  Appendix~\ref{appdx:NIAS_NIAC}. It says that identifying if a set $\agentset$ comprising  stopping actions of a Bayesian stopping agent in multiple environments is optimal and then reconstructing the costs incurred in the environments is equivalent to solving a convex feasibility problem. Theorem~\ref{thrm:NIAS_NIAC} provides a constructive procedure for the inverse learner to  generate set valued estimates of the stopping cost $\utilityagent{\agent}$ and expected cumulative continue cost $\sumruncostagent{\agent}$ for all environments $\agent\in\agentset$. Algorithms for convex feasibility such as interior points methods~\citep{BD04} can be used to check feasibility of~\eqref{eqn:NIAS_thrm_CD} and \eqref{eqn:NIAC_thrm_CD} (defined in \eqref{eqn:NIAS_def} and \eqref{eqn:NIAC_def} below) and construct a feasible solution.

The inequalities NIAS, NIAC and SUMCOST denoted abstractly in Theorem~\ref{thrm:NIAS_NIAC} are defined below:
\blue{\begin{definition}[NIAS, NIAC and SUMCOST inequalities] \label{def:NIAC_ineq}
Given dataset $\datainf$, stopping costs\\ $\{\utilityagent{\agent},\agent\in\agentset\}$ and expected continue costs $\{\sumruncostagent{\agent},\agent\in\agentset\}$: 
\begin{align}
    & \operatorname{NIAS}: \sum_{\state\in\mathcal{X}}p_{\agent}(\state|\action)(\utilityagent{\agent}-\utilitysymbolagent{\agent}(\state,b)) \leq 0,\forall \action,\agent.\label{eqn:NIAS_def}\\
    & \operatorname{NIAC}:  \sum_{\agent\in\widehat{\agentset}}
     \E_{\action\sim \sum_{\state}\prior(\state)p_{\agent}(\cdot|\state)}\left\{\min_{\action'\in\actionset}\E_{\state\sim p_{\agent}(\cdot|\action)}\{\utilityagent{\agent}-\utilityagentaltact{\agent+1}\}\right\}
      \leq 0,\nonumber\\
    & \text{for any subset of indices } \widehat{\agentset}
    \subseteq\agentset, \text{ where } \agent_k + 1 = \agent_{k+1} \text{ if } k<l \text{ and } \agent_l + 1 = \agent_1. \label{eqn:NIAC_def}\\
    &\sumcost:~
    \E_{\state\sim\prior,\action\sim p_{\agent}(\cdot|\state)}\{\utilityagent{\agent}\} + \sumruncostagent{\agent} \leq\E_{\action\sim p_{\agenttwo}(\action)}\{\min_{\action'\in\actionset} \E_{\state\sim p_{\agenttwo}(\cdot|\action)}\{\utilityagentaltact{\agent}\}\} + \sumruncostagent{\agenttwo},\nonumber\\
    & \forall~\agent,\agenttwo\in\agentset.\label{eqn:sumcostfeasible}
    \end{align}
    \blue{{\bf Reconstruction of expected cumulative continue cost}. If NIAS, NIAC and SUMCOST inequalities defined above have a feasible solution, the following convex reconstruction of the agent's expected continue cost is consistent with optimal Bayesian stopping~\eqref{eqn:opt_stop_act},~\eqref{eqn:opt_stop_time}, a stronger condition compared to relative optimality~\eqref{eqn:opt_stop_act},~\eqref{eqn:relopt_stop_time_continue_cost}:} 
    \begin{align}
    \widehat{\sumruncostagent{}}(\stoptime) & = \max_{\agent=1,2,\ldots,\numagents}~\left\{
     \sumruncostagent{\agent} + \grosspayoff_{\agent,\agent} - \grosspayoffsurr(\stoptime,\utilitysymbolagent{\agent})
     \right\},\text{ where}\label{eqn:reconstructed-RIcost-theorem}\\
     \grosspayoffsurr(\stoptime,\utilitysymbolagent{\agent}) & =\sum_{\action\in\actionset}\left(\sum_{\state\in\stateset}p_{\stoptime}(\action|\state)\prior(\state)\right)\min_{b\in\actionset}\sum_{\state\in\stateset} p_{\stoptime}(\state|\action)\utilitysymbolagent{\agent}(\state,b),~\text{and}\label{eqn:grosspayoffsurr-theorem}\\
     \grosspayoff_{\agent,\agent} & = \sum_{\state\in\stateset,\action\in\actionset} \prior(\state)p_\agent(\action|\state)\utilityagent{\agent}\label{eqn:grosspayoff-theorem}
\end{align}
\blue{The above reconstruction assumes the agent's mapping from the sequence of observations $\obs_{1:\funcstop(\stoptime)}$ to the space of actions is one-to-one, and is valid if and only if the agent's expected cumulative continue cost is convex.}
\end{definition}
}

Let us now provide an intuitive explanation for the abstract inequalities of Theorem~\ref{thrm:NIAS_NIAC}.\\
\underline{NIAS~\eqref{eqn:NIAS_thrm_CD}}: NIAS applies to each of the $\numagents$ environments in $\agentset$. NIAS checks if, for every environment, the agent chooses the optimal stop action given its stopping belief and stopping strategy.\\
\underline{NIAC~\eqref{eqn:NIAC_thrm_CD}}: NIAC checks for optimality of the agent's stopping strategies in $\numagents$ environments. Since the stopping agent chooses its strategies in a finite number ($\numagents$) environments, NIAC checks if the agent's strategy in the $\agent^\text{th}$ environment performs at least as well as the strategies of the agent in all other environments given the environment's stopping cost $\utilityagent{\agent}$, for all $\agent\in\agentset$. If so, it constructs a feasible set of stopping costs in the $\numagents$ environments so that the chosen strategies are consistent with an optimal stopping agent.\\
\underline{SUMCOST~\eqref{eqn:feasibilitycontinuecost}}: If the Bayesian agent is an optimal stopping agent (NIAS and NIAC have a feasible solution), SUMCOST constructs a set of feasible expected continue costs incurred by the Bayesian agent in the multiple environments. The feasibility of NIAS and NIAC ensures that the SUMCOST inequalities have a feasible solution. In \eqref{eqn:sumcostfeasible}, the RHS term is the expected cumulative cost of the agent in environment $n$ given the stopping costs in environment $m$. The feasibility inequality~(\ref{eqn:sumcostfeasible}) checks for feasible expected cumulative continue costs so that the agent's stopping strategies in $\agentset$ are identified as optimal by the inverse learner, i.e.\,, \eqref{eqn:relopt_stop_time_continue_cost} is satisfied. \blue{The reconstructed cost $\widehat{\sumruncostagent{}}$ \eqref{eqn:reconstructed-RIcost-theorem} is a convex interpolation of expected stopping costs and feasible scalars $\sumruncostagent{\agent}$~\eqref{eqn:sumcostfeasible} such that conditions~\eqref{eqn:opt_stop_act} and \eqref{eqn:opt_stop_time} for optimal Bayesian stopping hold; see Appendix~\ref{appdx:NIAS_NIAC} for a detailed discussion. We remark that the reconstruction in \eqref{eqn:reconstructed-RIcost-theorem} is only valid when (a) the inverse learner has no information about the agent's observation likelihood, and (b) the inverse learner does not know the agent's expected continue cost. In Table~\ref{tab:identifiability}, we highlight the subtle issues underpinning IRL identifiability for optimal Bayesian stopping in more detail. In Sec.\,\ref{sec:SHT} below, we discuss IRL for optimal Bayesian stopping when the inverse learner knows the agent's expected continue cost; hence, the reconstruction \eqref{eqn:reconstructed-RIcost-theorem} is no more required for achieving IRL.
}

\subsection{Discussion of Theorem~\ref{thrm:NIAS_NIAC}}\label{sec:disc-theorem-1}
We now discuss the implications of Theorem~\ref{thrm:NIAS_NIAC} and contextualize the NIAS and NIAC feasibility inequalities (\ref{eqn:NIAS_thrm_CD}), (\ref{eqn:NIAC_thrm_CD}) of Theorem~\ref{thrm:NIAS_NIAC}.
 
\noindent\textbf{(i) Necessity and Sufficiency.}\\ The NIAS and NIAC conditions~(\ref{eqn:NIAS_thrm_CD}), (\ref{eqn:NIAC_thrm_CD}) are necessary and sufficient for the inverse learner to identify an optimal stopping agent. This makes Theorem~\ref{thrm:NIAS_NIAC} a remarkable result. If no feasible solution exists, then the dataset~$\datainf$ cannot be rationalized by an optimal Bayesian stopping agent. Also, if there exists a feasible solution, then the dataset~$\datainf$ must be generated by an optimal stopping agent in multiple environments (Lemma~\ref{lem:relativeoptimality}).

\noindent\textbf{(ii) Set valued estimate vs point estimate.}\\ An important consequence of Theorem~\ref{thrm:NIAS_NIAC} is that the reconstructed utilities are set-valued estimates rather than point valued estimates even though the dataset $\datainf$ has $\numtrials\rightarrow\infty$ samples. Estimating the costs from the solution of a cost minimization problem is an ill-posed problem. Put differently, all points in the feasible set of rationalizing costs explain the dataset $\datainf$ equally well.

\noindent\textbf{(iii) Consistency of Set-Valued Estimate.}\\ \blue{The NIAS and NIAC inequalities are both necessary and sufficient for optimal Bayesian stopping.} The necessity implies that the true stopping costs and expected continue costs incurred by the agent are feasible wrt the convex NIAS and NIAC inequalities. Hence, the IRL procedure is consistent in that the set-valued estimator contains the true generating model.

\noindent\textbf{(iv) Context: NIAS and NIAC.}\\ The inequalities (\ref{eqn:opt_stop_act}), (\ref{eqn:relopt_stop_time_continue_cost}) for the inverse learner to identify an optimal stopping agent can be written in abstract notation as (\ref{eqn:NIAS_start}), (\ref{eqn:NIAC_start}), respectively, in terms of the variables $\{\utilitysymbol_\agent,\sumruncostsymbol_\agent\}_{\agent=1}^\agentset$:
\begin{align}
    &\operatorname{NIAS}(\{\{\obslikestoppingagent{\agent},\state\in\stateset\},\utilitysymbolagent{\agent},\agent\in\agentset\},\prior)\leq 0\label{eqn:NIAS_start},\\
    &\operatorname{NIAC}^\ast(\{\{\obslikestoppingagent{\agent},\state\in\stateset\},\utilitysymbolagent{\agent},\sumruncostagent{\agent},\agent\in\agentset\},\prior)\leq 0\label{eqn:NIAC_start}.
\end{align}
The inverse learner in our setup does not know the agent's observation sequences $\{\obs_{1:\funcstop},\agent\in\agentset\}$, observation likelihood $\oprob$ or the continue cost $\runcost$. \blue{Hence, as shown in Appendix~\ref{appdx:NIAS_NIAC}, the best the inverse learner can do is check for the feasibility of the NIAC~\eqref{eqn:NIAC_def} that does not depend on $\sumruncostsymbol_\agent$.} Otherwise, the IRL task is equivalent to using optimality equations (\ref{eqn:opt_stop_act}), (\ref{eqn:relopt_stop_time_continue_cost}) expressed abstractly as NIAS and NIAC$^\ast$ above to reconstruct the costs. 
Eq.\,\ref{eqn:NIAS_thrm_CD} and \ref{eqn:NIAC_thrm_CD}  in Theorem~\ref{thrm:NIAS_NIAC} specialize to \eqref{eqn:NIAS_start} and \eqref{eqn:NIAC_start} by replacing the action selection policy $\actselectagent{\agent}$ with the unknown likelihood $\obslikestoppingagent{\agent}$. Put differently, \eqref{eqn:NIAS_thrm_CD} and \eqref{eqn:NIAC_thrm_CD} defined in \eqref{eqn:NIAS_def},~\eqref{eqn:NIAC_def} can be viewed as surrogates of the feasibility conditions~\eqref{eqn:NIAS_start} and \eqref{eqn:NIAC_start}, respectively. However, as discussed in the proof in Appendix~\ref{appdx:NIAS_NIAC}, the action selection policy $\actselectagent{\agent}$ \blue{suffices for both necessity and sufficiency of Bayes optimality~\eqref{eqn:NIAS_start}, \eqref{eqn:NIAC_start} in spite of being a Blackwell noisy measurement of $\obslikestoppingagent{\agent}$. Also, observe the NIAC inequality~\eqref{eqn:NIAC_def} is independent of $\sumruncostsymbol_{\agent}$ and expressed only in terms of stopping costs $\utilitysymbol_\agent$. However, as shown in Appendix~\ref{appdx:NIAS_NIAC}, the feasibility of both inequalities \eqref{eqn:NIAC_start} and \eqref{eqn:NIAC_def} are equivalent. 
Finally, in some examples of stopping time problems such as SHT discussed in Sec.\,\ref{sec:SHT}, the inverse learner knows the agent's expected cumulative continue cost and hence, can use the NIAC$^\ast$ inequality as is to identify optimality and achieve IRL.}

\noindent \blue{
{\em NIAS and NIAC with $\eps$-feasibility.} One trivial solution that satisfies both NIAS and NIAC inequalities in Theorem~\ref{thrm:NIAS_NIAC} is the degenerate cost of all zeros. Such degeneracy is common in IRL literature due to the fundamental ill-posedness of the inverse optimization problem. In practice, one can ensure only non-trivial solutions pass the NIAS and NIAC feasibility inequalities by introducing a margin constraint:
\begin{equation}\label{eqn:niasc_margin}
    \operatorname{NIAS}(\cdot)\leq -\eps,~\operatorname{NIAC}(\cdot) \leq -\eps,~\eps>0.
\end{equation}
Margin constraints for ensuring non-degenerate solutions to feasibility tests are common practices in IRL~\citep{RAT06}. In complete analogy, using the $\epsilon$ restriction of \eqref{eqn:niasc_margin}, we can ensure only non-trivial informative costs pass the NIAS and NIAC feasibility test of Theorem~\ref{thrm:NIAS_NIAC}.
}

\noindent\textbf{(v) Private and Public Beliefs.}\\  The stopping belief $\belief_{\funcstop}$ in (\ref{eqn:opt_stop_time}) can be interpreted as the {\em private belief} evaluated by the agent after measuring $\obs_{1:\funcstop}$ in the sense of Bayesian  social learning~\citep{VK16,CHAM04}. Since $\belief_{\funcstop}$ is unavailable to the inverse learner, it uses the {\em public belief} $p(\state|\action)$ as a result of the agent's stop action to estimate its incurred costs.

\noindent \blue{\textbf{(vi) IRL for stopping agent whose observation likelihood changes with the environment.} For notational convenience, we assume the Bayesian agent's observation likelihood is fixed across different environments. However, in Appendix~\ref{appdx:NIAS_NIAC}, we discuss under what conditions the inverse learner can achieve IRL when the Bayesian agent's observation likelihoods change with the environment. We provide a specific example of the agent continue cost, namely, the entropic continue cost that facilitates the inverse learner to achieve IRL for different agent observation likelihoods in different environments. The agent's stopping cost in this case is a logistic function in terms of its action selection policy; the logistic function also arises in Max-Entropy IRL~\citep{ZB08}. This resemblance is not surprising; the agent in \citep{ZB08} maximizes its cumulative expected reward subject to a bound on the mutual information between the prior and  the distribution of beliefs induced by its policy. The objective function in \eqref{eqn:opt_stop_time} where $\sumruncostagent{}$ is the mutual information between the prior and the stopping belief is simply the Lagrangian form of the objective the agent aims to optimize in \citet{ZB08}. The IRL problem for agents that Maximize their expected terminal rewards with a mutual information penalty has also been studied in the Bayesian revealed preference literature by \citet{CD19}.
}

\noindent\textbf{(vii) IRL for boundedly-rational forward learner.
}\\
\blue{
For general POMDPs, it is difficult\footnote{\cite{PP87} show that solving partially observed Markov decision processes are in general PSPACE hard. The SHT and Search problems discussed in this paper are special cases where the optimal
stopping strategy is stationary due to the problem structure and characterized as a threshold policy in the belief space.} for a Bayesian sequential decision maker to compute the optimal policy $\stoptime^\ast$ in \eqref{eqn:opt_stop_act},~\eqref{eqn:opt_stop_time}. We say that a strategy $\hat{\stoptime}$ is $\epsilon$-optimal if the following condition holds:
\begin{equation}\label{eqn:eps_stopping}
\text{$\epsilon$-optimal Bayesian stopping: }J(\hat{\stoptime}) - J(\stoptime^\ast) \leq \epsilon,~\text{for some }\epsilon\geq 0.
\end{equation}
Eq.\,\ref{eqn:eps_stopping} arises when the forward learner uses sub-optimal procedures for solving the POMDP such
as approximate value iteration, open loop feedback and finite state controllers. When both the stopping cost and the expected continue cost are free variables like in Theorem~\ref{thrm:NIAS_NIAC}, detecting $\epsilon$-optimality is non-identifiable and a difficult task. However, if either the stopping cost or the expected continue cost, (such as in the case of SHT discussed in Sec.\,\ref{sec:SHT}) is known to the inverse learner, one can identify $\epsilon$-optimality based on the feasibility of the IRL inequalities. We briefly discuss identification of $\epsilon$-optimality after Theorem~\ref{thrm:classic_SHT}; a general framework is beyond the scope of this paper and the subject of  future work. Indeed, more precise knowledge of the agent's sub-optimality allows the inverse learner to achieve IRL; see \cite{irl_subopt} for a discussion on how to achieve IRL when the inverse learner has access to a ranked set of forward learner's decision trajectories, ranked according to the extent of sub-optimality in each trajectory.}

\noindent \blue{(viii) {\bf No knowledge of observation likelihood by the inverse learner.} This paper assumes the inverse learner has no knowledge of the agent's observation likelihood. The sufficiency proof of Theorem~\ref{thrm:NIAS_NIAC} exploits this zero-knowledge assumption and posits that the inverse learner can thus assume a one-to-one mapping from the space of observation sequences $\obs_{1:\funcstop(\stoptime)}$ to the space of stopping actions. Indeed, one can show that if the instantaneous continue cost has an entropic form, for example, the Shannon-Gibbs entropy, R{\'e}nyi entropy or Tsallis entropy, the optimal mapping from observation sequences to stopping actions is one-to-one due to the strongly concave nature of these costs; see \citet{CD19} for a discussion of IRL for entropic costs. 
}

\blue{
}

\noindent \blue{(ix) {\bf Partial knowledge of agent costs.} If the Bayesian agent's instantaneous continue cost is zero, then it is optimal to never stop sensing, i.e., the agent observe infinitely many samples and the posterior belief approaches the Dirac delta function centered at the state $\state$\footnote{It follows from Bernstein-von Mises theorem~\citep{LEC53} that, under mild smoothness conditions, the agent's posterior belief converges asymptotically to a normal distribution centered around the maximum likelihood estimate with covariance $\lim_{\timeinst\rightarrow\infty} (\timeinst~I(\state))^{-1}$, where $I$ denotes the Fisher information matrix.}. Hence, the optimal $p_\agent(\action|\state)$ has non-zero weights if and only if $\action\in\argmin_{\action'\in\actionset}~\utilitysymbolagent{\agent}(\state,\action')$. Then checking for optimal Bayesian stopping with zero running cost is equivalent to identifying feasible stopping costs that satisfy the following condition:
\begin{equation}
    p_{\agent}(\action|\state) \neq 0 \iff \action\in\argmin_{\action'\in\actionset}~\utilitysymbolagent{\agent}(\state,\action'). 
\end{equation}
Sec.\,\ref{sec:SHT} considers the case where the instantaneous continue cost is a constant, hence the cumulative expected continue cost is proportional to the expected stopping time of the agent. If the inverse learner knows the expected continue cost, IRL is  achieved by checking for the existence of feasible stopping costs that satisfy the NIAS~\eqref{eqn:NIAS_def} and SUMCOST~\eqref{eqn:sumcostfeasible} inequalities with $\sumruncostagent{\agent}$ set to the agent's expected continue cost in environment $\agent$.
}

\noindent \blue{(x) {\bf IRL with $\eps$-feasibility.} If neither the stopping costs nor the expected continue costs are known to the inverse learner, the NIAS, NIAC and SUMCOST inequalities are trivially feasible by choosing the degenerate solution of constant costs. In this case it makes sense to construct  the inverse learner's non-trivial IRL cost estimate  as the set of feasible costs $\{\utilityagent{\agent},\sumruncostagent{\agent},\agent\in\agentset\}$ that are $\epsilon$-feasible wrt the NIAC, NIAC and SUMCOST inequalities:
\begin{align}
    &\bullet~\text{Choose feasibility margins }\epsilon_{NIAS},~\epsilon_{NIAC},~\epsilon_{SUMCOST}\geq 0,~\text{ not all zero}.\nonumber\\
    &\bullet~\text{Construct the set-valued IRL estimate as the set of all tuples }\{\utilityagent{\agent},\sumruncostagent{\agent},\agent\in\agentset\}~\text{that satisfy}\nonumber\\
    & \operatorname{NIAS}(\cdot)\leq \epsilon_{NIAS},~\operatorname{NIAC}(\cdot)\leq \epsilon_{NIAC}~\text{ and }\operatorname{SUMCOST}(\cdot)\leq \epsilon_{SUMCOST}.\label{eqn:epsilon-feasibility-theorem}
\end{align}
}

\subsection{Discussion of \ref{asmp:IRL1} and \ref{asmp:IRL2}}
\label{sec:disc_assumpt}
\noindent {\bf \ref{asmp:IRL1}}: To motivate \ref{asmp:IRL1}, suppose for each environment $m \in \agentset$, the inverse learner records the Bayesian stopping agent's true state $\tstatedata$, stopping action $\actiondata$ and stopping time  $\stoptimedata$ over $\trial=1,2,\ldots,\numtrials$ independent trials. Then the  pmf $\actselectagent{\agent}$ in (\ref{eqn:IRL_tuple}) is  the limit pmf of the empirical pmf $\actselectagentemp{\agent}$  as the number of trials $\numtrials\rightarrow\infty$ defined as:
\begin{equation}\label{eqn:actionselection}
\actselectagentemp{\agent} =\frac{ \sum_{\trial=1}^\numtrials\mathbbm{1}\{\tstatedata=\state,\actiondata=\action\} }{ \sum_{\trial=1}^\numtrials\mathbbm{1}\{\tstatedata=\state\}}.
\end{equation}

Specifically, since for each $\agent\in\agentset$ the sequence $\{\tstatedata,\actiondata\}$ is i.i.d for $\trial=1,2,\ldots \numtrials$, by Kolmogorov's strong law of large numbers, as the number of trials $\numtrials\rightarrow\infty$, $\actselectagentemp{\agent}$ converges with probability $1$ to the  pmf $\actselectagent{\agent}$. In the remainder of the paper (apart from Sec.\,\ref{sec:finitesample}), we will work with the asymptotic dataset $\datainf$ for IRL.
In Sec.\,\ref{sec:finitesample} we analyze the effect of finite sample size $K$ on the inverse learner using concentration inequalities.\vspace{0.1cm}\\

\noindent {\bf \ref{asmp:IRL2}}: \ref{asmp:IRL2} is necessary for the  identification of an optimal stopping agent~(Lemma~\ref{lem:relativeoptimality}) to be well-posed. Suppose \ref{asmp:IRL2} does not hold. Then, for $\numagents=2$ and true stopping costs $\utilitysymbol_1=\utilitysymbol_2$, we have $\actselectagent{1}=\actselectagent{2}$ in $\datainf$. This implies the set of feasible solutions $(\sumruncostagent{1},\sumruncostagent{2})$ for the feasibility inequality (\ref{eqn:sumcostfeasible}) is the set $\{(\sumruncostagent{1},\sumruncostagent{2}): \sumruncostagent{1}=\sumruncostagent{2},\sumruncostagent{1},\sumruncostagent{2}\in\reals_+\}$ and is hence, unidentifiable. 
\footnote{The condition $\numagents=1$ (or equivalently, $\numagents=2$ with equal stopping costs) is analogous  to  probing an agent with the same probe vector in classical revealed preferences~\citep{AF67,VAR12}. The obtained dataset of probes and responses can be rationalized by any concave, locally non-satiated, monotone utility function thus leading to loss of identifiability of the agent's utilities.}


\subsection{Outline of proof of Theorem~\ref{thrm:NIAS_NIAC}}
The proof of Theorem~\ref{thrm:NIAS_NIAC} in Appendix~\ref{appdx:NIAS_NIAC} involves two main ideas.
The first key idea is to specify a fictitious likelihood $ \prob_{\stoptime}(\obsfict|\state)$ parametrized by the stopping strategy so that given strategy $\stoptime$, observation likelihood $\oprob$ and prior $\prior$, the observation trajectory $\obs_{1:\funcstop}$ of the stopping time problem yields an identical stopping belief $\belief_{\stoptime}$, i.e., 
\begin{equation*}
    \prob(\obsfict|\state,\stoptime) = \prob\left(\{\obs_{1:\funcstop}\}:\belief_{\funcstop}=\belief|\state\right).
\end{equation*}
A more precise statement is given in (\ref{eqn:att_fun_basic}).
In other words,  a one-step Bayesian update using the likelihood
$ \prob(\obsfict|\state,\stoptime)$
 is equivalent to the multi-step Bayesian update (\ref{eqn:HMM_stopping}) of the state till the stopping time.
This idea is shown in Fig.\,\ref{fig:equivalance_stopping}. Recall that the cumulative expected cost of the agent comprises two components, the stopping cost and  cumulative continue cost. A useful property of this fictitious likelihood is that it is a sufficient statistic for the expected stopping cost  $\grosspayoff(\cdot)$.
\begin{figure}
	\centering
	\includegraphics[width=0.4\textwidth]{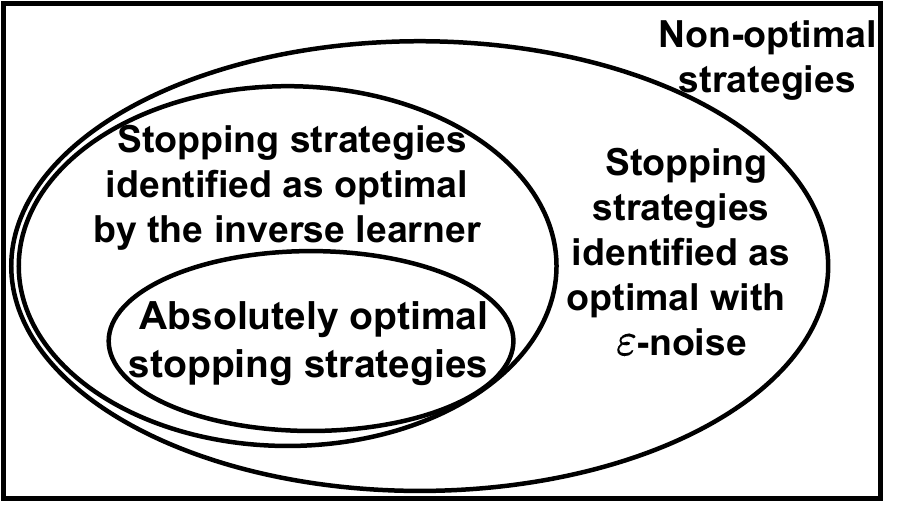}
	\caption{Given tuple $(\stoptuple,\runcost,\{\utilityagent{\agent},\agent\in\agentset\})$, the set of stopping strategies (\ref{eqn:relopt_stop_time_continue_cost}) of a stopping agent identified as an optimal stopping agent by the inverse learner (Lemma~\ref{lem:relativeoptimality}) contains the stopping strategies of an absolutely optimal agent defined by (\ref{eqn:opt_stop_act}), (\ref{eqn:opt_stop_time}). Such strategies can be obtained by small perturbations of the absolute optimal strategies such that the Bayesian stopping agent's strategy in each environment still performs better than that chosen by the agent in any other environment in $\agentset$. Like Sec.\,\ref{sec:BRP_RI}, Sec.\,\ref{sec:SHT} and \ref{sec:Search}, deals with identifying such optimal strategies for the SHT and search problems. In Sec.\,\ref{sec:finitesample}, we will detect if the agent's strategies corrupted by noise (due to finite sample constraints) belong to the set of strategies identified as optimal strategies by the inverse learner.}
	\label{fig:rel_abs_opt}
\end{figure}

The second main idea is to formulate the agent's expected cumulative cost using the observed action selection policy $\actselect$ of the agent instead of the unobserved fictitious likelihood $\obslikestoppingagent{\agent}$ that determines the expected stopping cost. $\actselectagent{\agent}$~(\ref{eqn:IRL_tuple}) is a stochastically garbled (noisy) version of $\obslikestoppingagent{\agent}$. We use this concept to formulate the NIAS and NIAC inequalities whose feasibility given $\datainf$ is necessary and sufficient for identifying an optimal stopping by a Bayesian stopping agent in multiple environments. 

 Showing that feasibility of the NIAS and NIAC inequalities (\ref{eqn:NIAS_thrm_CD}), (\ref{eqn:NIAC_thrm_CD}) is a necessary condition for the stopping strategies chosen by the Bayesian stopping agent to be optimal, \eqref{eqn:opt_stop_act}, \eqref{eqn:relopt_stop_time_continue_cost} is straightforward. The key idea in the sufficiency proof is to note that the elements of the garbling matrix that maps the fictitious observation likelihood to the action selection policy is unknown to the inverse learner. Hence, the inverse learner can arbitrarily assume $\actselectagent{\agent}$ to be an accurate measurement of $\obslikestoppingagent{\agent}$. We then show that for a feasible set of viable stopping costs $\{\utilityagent{\agent},\sumruncostagent{\agent},\agent\in\agentset\}$ that satisfy the NIAS and NIAC inequalities, there exist a set of positive reals $\{\sumruncostsymbol_{\agent},\agent\in\agentset\}$ that satisfy (\ref{eqn:opt_stop_act}), (\ref{eqn:opt_stop_time}) with the expected cumulative continue cost incurred by the agent in the $\agent^{\text{th}}$ environment set to $\sumruncostsymbol_{\agent}$.
 
The NIAS and NIAC inequalities are convex in the stopping costs $\utilitysymbolagent{\agent},\agent\in\agentset$. The inverse learner can solve for these convex feasibility constraints to obtain a feasible solution. Thus, we have a constructive IRL procedure for reconstructing the stopping and expected cumulative continue costs for the inverse optimal stopping time problem.

\begin{figure}[ht]
	\centering
	\includegraphics[width=0.75\textwidth]{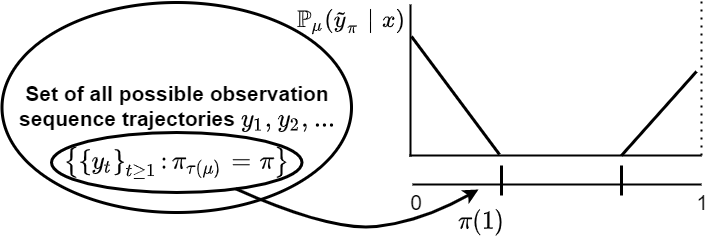}
	\caption{Schematic illustration of first main idea of proof of Theorem \ref{thrm:NIAS_NIAC} for the case when $\numstates=2$. The key idea is to construct a fictitious observation likelihood $\prob_{\stoptime}(\obsfict|\state)$ for compact representation of the agent's expected stopping cost. The probability of generating the fictitious observation $\obsfict$ is equal to the probability of a sequence of observations yielding a stopping belief $\belief$ for a given stopping time $\stoptime$.
	}
	\label{fig:equivalance_stopping}
\end{figure}
\subsection{Summary} This section has laid the groundwork for IRL of a Bayesian stopping time agent.
Specifically,  we  discussed the dynamics of the Bayesian stopping time agent in a single environment \eqref{eqn:IRL_tuple_stop} and multiple environments~\eqref{eqn:IRL_tuple_compact}.  We then described the IRL problem that the  inverse learner aims to solve. Theorem \ref{thrm:NIAS_NIAC} gave a necessary and sufficient condition for a Bayesian stopping time agent to be identified as an optimal stopping agent when its decisions in multiple environments are observed by the inverse learner. The agent's stopping cost in each environment can be estimated by solving a convex feasibility problem. Theorem~\ref{thrm:NIAS_NIAC} forms the basis of the IRL framework in this paper. \blue{Next, we develop IRL results for $2$ examples of stopping time problems, namely, sequential hypothesis testing and Bayesian search.}

\section{Example 1. IRL for sequential hypothesis testing (SHT)}
\label{sec:SHT}
We now discuss our first example of IRL for an optimal Bayesian stopping time problem, namely, {\em inverse}  Sequential Hypothesis Testing (SHT). Our  main result below 
(Theorem~\ref{thrm:classic_SHT}) specifies a necessary and sufficient condition for  IRL in  SHT. \blue{The SHT problem is a special case of the optimal Bayesian stopping problem discussed in Sec.\,\ref{sec:optimal_stopping_time} since the continue cost $\runcostinst_\timeinst$~\eqref{eqn:IRL_tuple_stop} is a constant for all time $\timeinst$ in the SHT problem. For our IRL task, the continue cost can be chosen as $1$ WLOG.
}
\subsection{Sequential hypothesis testing (SHT) Problem}\label{sec:SHT_back}
Let $\obs_{1},\obs_{2},\ldots$ be a sequence of i.i.d observations. Suppose the Bayesian agent knows that the pdf  of $\obs_i$ is either $p(\obs|\state=1)$ or $p(\obs|\state=2)$. The aim of classical SHT  is to decide sequentially on whether $\state=1$ or $\state=2$ by minimizing a combination of the continue (measurement) cost and misclassification cost. In analogy to Sec.\,\ref{sec:optimal_stopping_time}, we now define a set of SHT environments in which a Bayesian stopping agent operates. 

\begin{definition}[Optimal SHT in multiple environments]
 The set $\agentset$ of optimal SHT in multiple environments is a special case of optimal stopping in multiple environments $\optstoptuple$~\eqref{eqn:IRL_tuple_compact} with:
\begin{compactitem}
	\item $\stateset=\{1,2\}$, $\mathcal{Y}\subset \mathbb{R}$ , $\actionset=\stateset$.
	\item $\runcost=\{\runcostinst_{\timeinst}\}_{\timeinst\geq 0},~\runcostinst_{\timeinst}(\state)=\runcostinst\in\reals^+,~\forall \state\in\stateset$ is the constant continue cost.
	\item $\{\stoptime_{\agent},\agent\in\agentset\}$ are the SHT stopping strategies chosen by the Bayesian agent over $\numagents$ SHT environments defined below.
	\item $\utilityagent{\agent}$ is the stopping cost incurred by the agent in the $\agent^{\text{th}}$ SHT environment  parametrized by misclassification costs $(\mc_{\agent,1},\mc_{\agent,2})$.
	\begin{equation*}
	    \utilityagent{\agent} = \begin{cases}
	    \mc_{\agent,1}, & \text{ if } \state=1,\action=2,\\
	    \mc_{\agent,2}, & \text{ if } \state=2,\action=1,\\
	    0, & \text{ if } \state=\action\in\{1,2\}.
	    \end{cases}
	\end{equation*}
	\end{compactitem} 
\end{definition}
The SHT stopping strategies in the above definition satisfy the optimality conditions in Definition~\ref{def:absoptimality} and
can be computed using stochastic dynamic programming~\citep{VK16}. The solution for $\stoptime_\agent$ for the $\agent^{\text{th}}$ SHT environment is well-known~\citep{LVJ87} to be a stationary policy with the following threshold rule parameterized by scalars $\alpha_{\agent},\beta_{\agent}\in(0,1)$:
	\begin{align}\stoptime_{\agent} (\belief) = \begin{cases}
	\text{choose action }2, & \mbox{if } 0\leq \belief(\state=2) \leq \beta_{\agent} \\
	\mbox{continue,} & \mbox{if } \beta_{\agent} < \belief(\state=2) \leq \alpha_{\agent} \\
	\text{choose action } 1, & \mbox{if } \alpha_{\agent}<\belief(\state=2) \leq 1.
	\end{cases}
	\label{eqn:opt_policy}
	\end{align}
{\em Remark:}
Since the SHT  dynamics can be parameterized by $c,\mc_1,\mc_2$, we can set $c=1$
without loss of generality since the optimal policy is unaffected. \blue{Also, the expected cumulative continue cost of the agent is simply the expected stopping time of the agent.}

\subsection{IRL for inverse SHT. Main assumptions}\label{sec:IRL_SHT}
Suppose the inverse learner observes the actions of  a Bayesian stopping agent in $\numagents$  SHT environments. In addition to assumptions \ref{asmp:IRL2}, we assume the following about the inverse learner performing IRL for identifying an SHT agent:
\begin{enumerate}[label=(A\arabic*)]
\setcounter{enumi}{\value{assumindex}}
\item \label{asmp:SHT1} The inverse learner has the dataset 
\begin{equation}
\datainf(SHT)=(\datainf,\{\sumruncostagent{\agent},\agent\in\agentset\}),
    \label{eqn:dataset_SHT}
\end{equation}
where $\datainf$ is defined in (\ref{eqn:IRL_tuple}), $\sumruncostagent{\agent}=\mathbb{E}_{\stoptime_{\agent}}\{\funcstop\}$ is the expected continue cost incurred by the Bayesian agent in the  $\agent^\text{th}$ environment.
\item \label{asmp:SHT2} The stopping strategies $\{\stoptime_{\agent},\agent\in\agentset\}$ are stationary strategies characterized by the threshold structure in (\ref{eqn:opt_policy}).
\item \label{asmp:SHT3} There exist reals $\delta_1,~\delta_2\in(0,1)$ such that the following conditions are satisfied:
  \begin{align} 
  (i)&~\beta_{\agent}\leq \delta_1\leq \delta_2\leq\alpha_{\agent},~\forall \agent\in\agentset,~ (ii)&\delta_1/(1-\delta_1) \leq \mc_{\agent,1}/\mc_{\agent,2} \leq \delta_2/(1-\delta_2),\nonumber
 \end{align}
 where $\alpha_{\agent},\beta_{\agent}$ are the threshold values of the stationary strategy $\stoptime_{\agent}$ chosen by the Bayesian agent in environment $\agent$.
\setcounter{assumindex}{\value{enumi}}
\end{enumerate}


{\em Remarks}: (i) Assumption \ref{asmp:SHT1} specifies additional information the inverse learner has for performing IRL for SHT by recording the agent decisions over $\numtrials\rightarrow\infty$ independent trials. Since the continue cost is 1,  the expected cumulative continue cost is simply the expected stopping time of the agent. The inverse learner obtains an a.s.\ consistent  estimate of the expected stopping time by computing the sample average of the $\numtrials$  stopping times. \blue{Since the expected continue cost is simply the expected stopping time of the agent and known to the inverse learner, it is no more a feasible variable in the feasibility equations \eqref{eqn:NIAC_def}. This yields a smaller feasibility set for the stopping costs.}\\
(ii) Assumption \ref{asmp:SHT2} comprises partial information the inverse learner has about the stopping strategies chosen by the agent and its observation likelihood. \blue{Since the optimal stopping strategy is well-known to have a threshold structure~\cite{LVJ87}, the inverse learner only needs to compare the expected cost incurred from threshold policies to check for optimality and achieving IRL.\\
(iii) Assumption \ref{asmp:SHT3} ensures the expected stopping cost of the SHT agent $\grosspayoff(\stoptime_{\agent},\utilitysymbol)$~\eqref{eqn:opt_stop_time} that depends on the unobserved strategy $\stoptime_\agent$ can be expressed in terms of the induced action selection policy $\actselectagent{\agent}$ for any stopping cost $\utilitysymbol$, i.e.\,,
    $\grosspayoff(\stoptime_{\agent},\utilitysymbolagent{\agenttwo})=  \E_{p_\agent(\action)} \{\E_{\state\sim p_\agent(\cdot|\action)}\{\utilitysymbol(\state,\action)\}\}$. 
}

\subsection{IRL for inverse SHT. Main result}
Our main result below specifies a set of linear inequalities that are necessary and sufficient for the Bayesian agent's actions observed by the inverse learner to be identified as that of an optimal SHT agent (Lemma~\ref{lem:relativeoptimality}). Any feasible solution constitutes a viable  SHT misclassification cost for the $\numagents$ SHT environments in which the Bayesian agent operates.

\begin{theorem}[IRL for inverse SHT] \label{thrm:classic_SHT}
    Consider the inverse learner with dataset $\datainf(\SHT)$ (\ref{eqn:dataset_SHT}) obtained from a Bayesian agent taking actions in $\numagents$ SHT environments. Assume \ref{asmp:IRL2} holds. Then:\\
    {\em 1.} \underline{Identifiability}: The inverse learner can identify if the dataset $\datainf(\SHT)$ is generated by an optimal SHT agent (Lemma~\ref{lem:relativeoptimality}).\\ 
	{\em 2.} \underline{Existence}: There exists an optimal SHT agent parameterized by tuple $\optsearchtuple$~\eqref{eqn:IRL_tuple_compact}, if and only if there exists a feasible solution to the following \blue{convex}~(in stopping costs) inequalities: 
	\begin{align} 
	\text{Find }&\utilityagent{\agent}> 0,\utilitysymbolagent{\agent}(\state,\state)=0,~\forall \state,\action\in\stateset,~\agent\in\agentset\text{ s.t.}\nonumber\\
	\operatorname{NIAS}:&\sum_{\state\in\mathcal{X}}p_{\agent}(\state|\action)(\utilityagent{\agent}-\utilitysymbolagent{\agent}(\state,b)) \leq 0,\forall \action,\actiontwo,\agent.\nonumber\\
	\operatorname{NIAC}^\ast:&\left(\sum_{\state,\action}\prior(\state)\actselectagent{\agent}\utilityagent{\agent} + \sumruncostagent{\agent}\right) -\left(\sum_{\action}p_{\agenttwo}(a)\min_{\actiontwo}\sum_{x}p_{\agenttwo}(x|a)\utilitysymbolagent{\agent}(\state,\actiontwo)
	+\sumruncostagent{\agenttwo}\right)\leq 0,\nonumber\\
	&\forall m,n\in\agentset,m\neq n.\label{eqn:NIAC_ast_def}
	\end{align}
	\blue{(Recall that $C_m = \E_{\stoptime_{\agent}}\{\funcstop\}$ is known to the inverse learner, and hence is not a free variable)}.\\
	{\em 3.} \underline{Reconstruction}: The set-valued IRL estimates of the SHT misclassification costs\\ $\{\mc_\agent,m\in\agentset\}$ are defined below where  $\mc_\agent = (\mc_{1,m},\mc_{2,m})$: \vspace{-0.2cm}
	\begin{align*}
	    &\mc_{1,\agent} = \utilitysymbolagent{\agent}(1,2),~\mc_{2,\agent}=\utilitysymbolagent{\agent}(2,1)~\forall \agent\in\agentset,
	\end{align*}
	where $\{\utilityagent{\agent},\agent\in\agentset\}$ is any feasible solution to the NIAS and NIAC$^{\ast}$ inequalities.\qedsymbol
\end{theorem} 
Theorem~\ref{thrm:classic_SHT} is a special instance of Theorem~\ref{thrm:NIAS_NIAC} for identifying an optimal stopping agent operating in multiple environments. \blue{The NIAC$^\ast$ resembles SUMCOST~\eqref{eqn:sumcostfeasible} with the only difference that $\sumruncostagent{\agent}$ is the expected stopping time of the agent in environment $\agent$ instead of being a feasible variable like in \eqref{eqn:sumcostfeasible}. We note that since the expected stopping time is non-convex in the agent's action selection policy $\actselectagent{\agent}$, the inverse learner cannot use the convex reconstruction procedure of \eqref{eqn:reconstructed-RIcost-theorem} to estimate the expected stopping time for any other policy.} \\
{\em Remarks:} \\
1. \blue{Inverse SHT is an IRL task with partially specified costs: out of the continue and stopping costs, the continue cost incurred by the Bayesian agent is already known to the inverse learner. As a consequence, the feasibility test for identifying an optimal SHT agent imposes tighter restrictions (fewer feasible variables) compared to identifying optimal stopping in Theorem~\ref{thrm:NIAS_NIAC} and avoids degenerate feasible solutions that trivially satisfy the inequalities~\eqref{eqn:NIAC_ast_def} of Theorem~\ref{thrm:classic_SHT}.} \\
\blue{2. {\em IRL for Multi-state SHT.} Theorem~\ref{thrm:classic_SHT} is independent of the number of states $\numstates$. When $\numstates>2$, IRL for inverse SHT comprises estimating the misclassification costs $\{\mc_{\agent,\state,\action},~\state\neq\action,\state,\action\in\numstates\}$, and is achieved by solving the feasibility inequalities \eqref{eqn:NIAS_def} and \eqref{eqn:NIAC_ast_def} of Theorem~\ref{thrm:classic_SHT}.\footnote{\blue{Since the state and environment index suffice to denote the misclassification cost when $\numstates=2$, the subscript `$\action$' is dropped from the misclassification cost notation in Lemma~\ref{lem:relativeoptimality} for notational clarity.}}}\\
\blue{
3. {\em Inverse SHT for boundedly-rational forward learner.} In Sec.\,\ref{sec:disc-theorem-1}, we discussed the concept of $\epsilon$-optimality for a forward learner. Below, we briefly discuss how the NIAS and NIAC$^\ast$ feasibility inequalities of Theorem~\ref{thrm:classic_SHT} can identify if an agent performs $\epsilon$-optimal SHT when the inverse learner knows the agent's expected continue cost.\\
If NIAS and NIAC$^\ast$~\eqref{eqn:NIAC_ast_def} are feasible, then one cannot say if the dataset $\datainf$~\eqref{eqn:dataset_SHT} is generated from an absolutely optimal Bayesian agent~(Definition~\ref{def:absoptimality}) or an $\epsilon$-optimal Bayesian agent~\eqref{eqn:eps_stopping}. However, if $\datainf$ fails the feasibility test~\eqref{eqn:NIAC_ast_def} of Theorem~\ref{thrm:classic_SHT}, then it is clear $\datainf$ results from an $\epsilon$-optimal Bayesian agent, where a bound on $\epsilon$ can be obtained by finding the minimum relaxation needed for passing the feasibility test~\eqref{eqn:NIAC_ast_def}:
\begin{equation}\label{eqn:detect_eps_optimality}
    \min_{\epsilon_{\text{relax}}\geq 0} \epsilon_{\text{relax}},~\text{ such that }~\operatorname{NIAS}(\datainf,\{\utilityagent{\agent}\})\leq \epsilon_{\text{relax}},~\operatorname{NIAS}(\datainf,\{\utilityagent{\agent}\})\leq \epsilon_{\text{relax}}.
\end{equation}
The $\epsilon$-relaxation in \eqref{eqn:detect_eps_optimality} arises frequently in microeconomic theory in robustness tests to measure how far an economic agent is from satisfying economics-based rationality. Some examples of widely used robustness measures in economics literature include the Houtman index (HM-Index)~\citep{HM85}, Afriat measure~\citep{AFT72} and Varian measure~\citep{VAR91}. }

\subsection{Numerical example illustrating IRL for inverse SHT}
\label{sec:Example_SHT}
\blue{We now present a toy numerical example for inverse SHT with $3$ SHT environments and $3$ states. The  aim of this example is to illustrate the consistency property of Theorem~\ref{thrm:classic_SHT}. That is, that the true misclassification costs lie in the set of feasible costs computed by the inverse learner by solving the convex feasibility test of  Theorem~\ref{thrm:classic_SHT}.}

{\em SHT environments.} We consider $\numagents=3$ SHT environments with:
\begin{compactitem}
    \item {\em Prior} $ \prior = [0.5~0.5]'$.
    \item {\em Observation likelihood}: $p(\obs|\state=1)=\mathcal{N}(1,2)$, $p(\obs|\state=2)= \mathcal{N}(-1,2)$,
where $\mathcal{N}(\mu,\sigma^2)$ denotes the normal distribution with mean $\mu$ and variance $\sigma^2$.
\item {\em Misclassification costs}:\\
Environment 1: $(\mc_{1,1},\mc_{1,2})=(2,2.5)$, 
Environment 2: $(\mc_{2,1},\mc_{2,2})=(4,3)$, 
Environment 3: $(\mc_{3,1},\mc_{3,2})=(6,6)$.
\end{compactitem}

{\em Inverse Learner specification.} Next we  consider the inverse learner.
We generate $\numtrials=10^5$ samples for the $3$ SHT environments using the above parameters. Recall from Theorem~\ref{thrm:classic_SHT} that the inverse learner uses the dataset $\datainf(\SHT)$ to perform IRL for inverse SHT, where $\datainf(\SHT)$ is defined as: \vspace{-0.25cm}
\begin{equation}
\label{eqn:example_SHT_dataset}
    \datainf(\SHT) = (\prior, (\hat{p}_{\agent}(\action|\state),\sum_{\trial=1}^{K}\stoptimedata/\numtrials),\agent\in\{1,2,3\}),
\end{equation}
where $\numtrials=10^5$, the second and third terms are the empirically calculated action selection policy and expected stopping time for SHT environments $\agent$ from the $10^5$ generated samples. \blue{We denote the action selection policy in \eqref{eqn:example_SHT_dataset} as $\hat{p}_\agent(\action|\state)$ and not $p_\agent(\action|\state)$ since the numerical example uses an empirical estimate.}
\begin{figure}[t]
\centering
\includegraphics[width=0.45\textwidth]{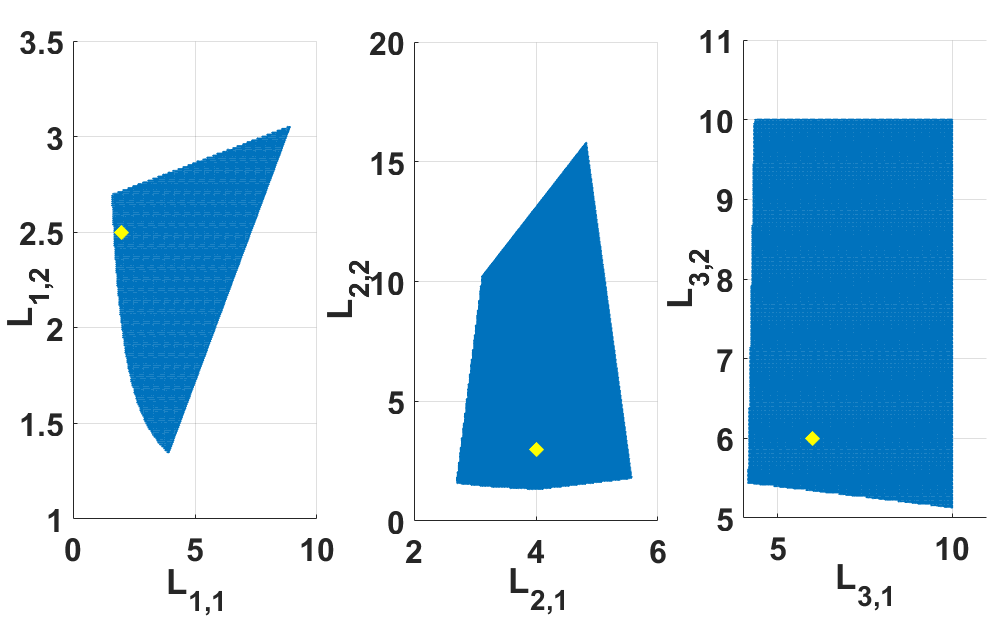}
\caption{Inverse SHT numerical example with parameters specified in Sec.\,\ref{sec:Example_SHT}.
The key observation is that the true misclassification costs (yellow points) lie in the feasible set~(blue region) of costs computed via Theorem~\ref{thrm:classic_SHT}. This follows from the necessity proof of Theorem~\ref{thrm:classic_SHT} which says if the Bayesian agent is an optimal SHT agent, then the true costs lie in the feasible set of costs that satisfy the NIAS and NIAC$^{\ast}$ inequalities.
Highlighting the advantage of the set-valued estimate of our IRL algorithm, we note that all points in the blue feasible region rationalize the observed stop actions of the Bayesian agent equally well. \blue{Indeed, the feasible region shrinks with the number of environments $\numagents$.}}
\label{fig:results}
\end{figure}

{\em IRL Result.} The inverse learner performs IRL by using the dataset $\datainf(\SHT)$ (\ref{eqn:example_SHT_dataset}) to solve the linear feasibility problem in Theorem~\ref{thrm:classic_SHT}. The result of the feasibility test is shown in Fig.\,\ref{fig:results}. The blue region is the set of feasible misclassification costs for each SHT environment. 
The feasible set of costs is $\{(\mc_{\agent,1},\mc_{\agent,2}),\agent\in\{1,2,3\}\}\subseteq \mathbb{R}^6_+$.
Fig.\,\ref{fig:results} displays the feasible misclassification costs for a single environment keeping the costs for the other two environments fixed at their true values. \blue{The need to fix costs for the other two environments for plotting the set of feasible costs is only for visualization purposes. It is not possible to plot a $6$ dimensional point (vector of estimated misclassification costs for $3$ SHT environments) on the $2$-d plane.}

The true misclassification costs for each SHT environment are highlighted by a yellow point. The key observation is  that these true costs belong to the set of feasible costs (blue region) computed via Theorem~\ref{thrm:classic_SHT}. Thus, Theorem~\ref{thrm:classic_SHT} successfully performs IRL for the SHT problem and the set of feasible misclassification costs can be reconstructed as the solution to a linear feasibility problem. Also, all points in the set of misclassification costs explain the SHT dataset 
equally well.

\blue{
\subsection{Numerical example. Regularized max-margin IRL for inverse SHT.}\label{sec:example_regul_IRL_SHT}
}
\blue{We now present a numerical example for inverse SHT involving $\numagents = 100$ environments where we compute a point-valued IRL estimate of the SHT misclassification costs. This inference task is in contrast to the set-valued IRL flavor considered thus far in the paper.}
\blue{Given dataset $\datainf(\SHT)$~\eqref{eqn:example_SHT_dataset}, we compute a point estimate $\mc^\ast$ of misclassification costs \blue{that maximizes the $\mathcal{L}_2$-regularized margin of the NIAC$^\ast$ feasibility inequalities of Theorem~\ref{thrm:classic_SHT}. The point estimate $\mc^\ast$ is inspired by max-margin IRL methods in the literature
\citep{AB04,RAT06} and defined as:}
\begin{align}\label{eqn:regul_irl_sht}
&\mc^\ast = \argmin_{\mc}
\sum_{\agent,\agenttwo=1,\agent\neq\agenttwo}^{\numagents} \operatorname{Margin}_{\datainf(\SHT)}(m,n,\mc)  - \lambda \|\mc\|_2^2,\\
&\operatorname{Margin}_{\datainf(\SHT)}(m,n,\mc)  = \left(G(\hat{p}_{\agenttwo},\mc_{\agent}) + \hat{\sumruncostagent{\agenttwo}}\right) - \left(G(\hat{p}_{\agent},\mc_{\agent})  + \hat{\sumruncostagent{\agent}}\right) ,
\label{eqn:reg_irl_def}
\end{align}
\blue{where $G(\hat{p},\mc_\agent)$ is the expected misclassification cost for SHT with action selection policy $\hat{p}$ and misclassification costs $\mc_\agent$, and $\hat{\sumruncostsymbol}_\agent=\sum_{\trial=1}^{K}\stoptimedata/\numtrials)$ is the agent's expected continue cost in environment $\agent$ computed empirically from $\numtrials$ independent trials. In simple terms, \eqref{eqn:reg_irl_def} is the difference in expected cumulative cost between action policies $\hat{p}_\agent$ and $\hat{p}_\agenttwo$ for a fixed misclassification cost $\mc_\agent$.} The objective function in~\eqref{eqn:regul_irl_sht} is the $\mathcal{L}_2$-norm regularized margin with which the {\em candidate} SHT misclassification costs pass the NIAC$^\ast$ convex feasibility test of \eqref{eqn:NIAC_ast_def}.
In \eqref{eqn:regul_irl_sht}, $\lambda>0$ is a tunable regularization parameter and $G(\cdot)$ is the expected misclassification cost defined in \eqref{eqn:relopt_stop_time_continue_cost}. Setting $\lambda$ to $0$ yields the max-margin IRL estimate of the stopping agent's misclassification costs and lies 
within the feasible set of costs generated by Theorem~\ref{thrm:classic_SHT}. The other extreme is setting $\lambda$ to $\infty$ which results in $\mc^\ast=0$.\vspace{0.2cm}\\}

\begin{figure*}\label{fig:regul_inv_sht}
    \centering
    \includegraphics[width=0.45\linewidth]{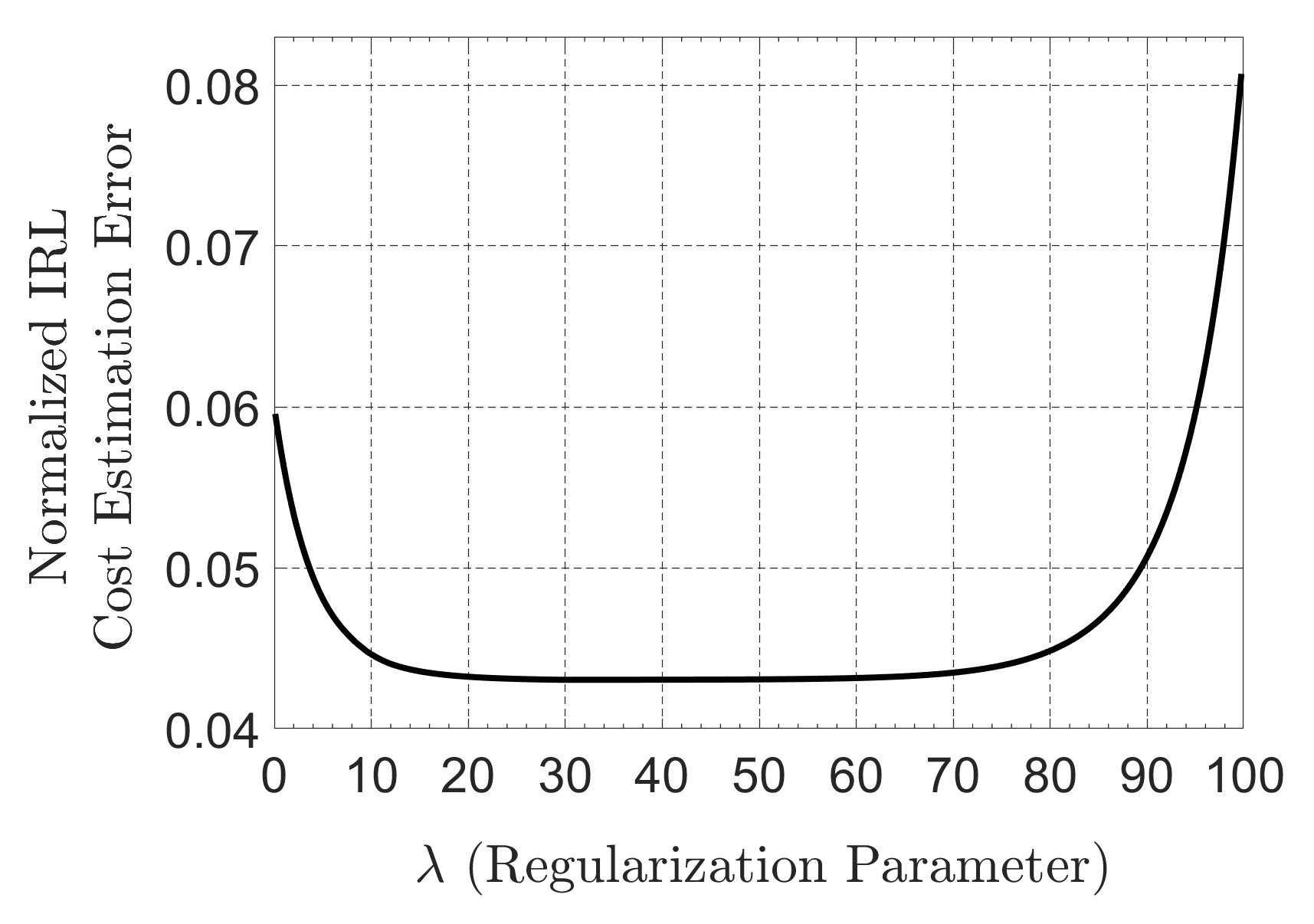}
    \caption{Inverse SHT numerical example for $100$ SHT environments, with parameters specified in Sec.\,\ref{sec:example_regul_IRL_SHT}. The main takeaway is that regularized max-margin IRL for inverse SHT~\eqref{eqn:regul_irl_sht} can estimate the misclassification costs incurred by the stopping agent in the $100$ SHT environments with up to $95\%$  accuracy by varying the regularization parameter $\lambda$ in \eqref{eqn:regul_irl_sht}.}
\end{figure*}

\blue{The following numerical example illustrates regularized IRL~\eqref{eqn:regul_irl_sht} for inverse SHT.\\
{\em SHT environments.} We consider $\numagents=100$ SHT environments with:
\begin{compactitem}
    \item {\em Prior} $ \prior = [1/4~~1/4~~1/4~~1/4]'$ (The state space is now $\stateset=\{1,2,3,4\}$).
    \item {\em Observation likelihood}: $p(\obs|\state=1)=\mathcal{N}(-2,8)$,~ $p(\obs|\state=2)= \mathcal{N}(0,8)$,\\
 , $p(\obs|\state=3)= \mathcal{N}(2,8)$ and $p(\obs|\state=4)= \mathcal{N}(4,8)$.
\item {\em Misclassification costs}: The misclassification costs $\mc=\{\mc_{\agent,\state,\action}\}$ in the $\numagents$ environments is uniformly sampled from the interval $[4,10]^{\numagents\times\numstates\times(\numstates-1)}$.
\end{compactitem}\vspace{0.1cm}
{\em Inverse Learner Specification}: The inverse learner aggregates the dataset $\datainf(\SHT)$ according to the procedure described in \eqref{eqn:example_SHT_dataset} by generating $\numtrials=10^7$ independent trials for the SHT agent in all $\numagents=100$ environments. Then, the inverse learner computes the regularized max-margin IRL estimate $\mc^\ast$ by solving the optimization problem~\eqref{eqn:regul_irl_sht}.\vspace{0.1cm}\\
{\em IRL Results:} The inverse learner performs IRL by using the dataset $\datainf(\SHT)$ to solve the optimization problem \eqref{eqn:regul_irl_sht}. Recall the dataset $\datainf(\SHT)$ is generated by observing the actions of an SHT agent in multiple environments with misclassification costs $\mc$. Figure~4 shows the estimation error $\|\mc^\ast-\mc\|_2/\|\mc|\|_2$ of the inverse learner's IRL estimate $\mc^\ast$~\eqref{eqn:regul_irl_sht} computed by the inverse learner as the regularization parameter $\lambda$ in \eqref{eqn:regul_irl_sht} is varied. The error is normalized wrt the $\mathcal{L}_2$-norm of the true misclassification costs in multiple environments incurred by the Bayesian agent whose actions comprise $\datainf(\SHT)$. }

\blue{The least estimation error obtained by varying $\lambda$ over the interval $[0,100]$ was observed to be $0.042$. In other words, the point IRL estimate obtained by solving the optimization problem \eqref{eqn:regul_irl_sht} can estimate the true misclassification costs of the SHT environments with up to $95\%$ accuracy. 
Indeed, the estimation accuracy increases with the number of environments at the cost of greater computation resources. Second, we observed that the error starts increasing sharply from $\lambda\sim 75$. This is expected since the regularization term in \eqref{eqn:regul_irl_sht} dominates the margin term at large values of $\lambda$.\vspace{-0.2cm}
}

\subsection{Performance Comparison. IRL for Inverse SHT and existing IRL methods for POMDPs}
\label{sec:comparison}
\blue{In this section, we compare the IRL performance of Theorem~\ref{thrm:classic_SHT} for inverse SHT against two well-known algorithms for IRL of POMDPs, namely, Max-Margin between Values (MMV)~\citep[Alg.\ 4]{CH11} and Max-Margin between Feature Expectations (MMFE)~\citep[Alg.\ 5]{CH11}. We compare the performance of MMV and MMFE algorithms against max-margin inverse SHT~\eqref{eqn:regul_irl_sht} with regularization parameter $\lambda$ set to 0.}

\blue{Recall from \eqref{eqn:dataset_SHT} that our inverse SHT result of Theorem~\ref{thrm:classic_SHT} requires state-terminal action pairs of the SHT agent over several independent trials and the expected stopping time of the SHT agent. 
In comparison, MMV and MMFE do not require the expected stopping time, but instead require complete knowledge of: (a) the observation likelihood of the Bayesian agent, and (b) the beliefs of the SHT agent at every time step. Moreover, MMV and MMFE  require a POMDP solver for IRL. }

\blue{To compare the performance of our IRL scheme~\eqref{eqn:regul_irl_sht} against MMV and MMFE, we perform two sets of numerical experiments with different specifications of the agent's observation likelihood:\\
{\em Case 1: Perfect Knowledge of SHT Model Dynamics.} MMV and MMFE have perfect knowledge of the SHT agent's observation likelihood.\\
{\em Case 2: Misspecified SHT Model Dynamics.} MMV and MMFE have misspecified knowledge of the SHT agent's observation likelihood. For environment $\agent$ the observation likelihood $p_\agent(\obs|\state)$ is misspecified to be the agent's action policy $p_\agent(\action|\state)$.
}
\subsubsection*{Experimental Setup}
\blue{\noindent For our numerical experiments, we consider $\numagents=4$ SHT environments with:
\begin{compactitem}
    \item {\em Prior} $ \prior = [1/2~~1/2]'$~(The state space is $\stateset=\{1,2\}$).
    \item {\em Observation likelihood}: $p(\obs|\state=1)=\mathcal{N}(+2,4)$,~ $p(\obs|\state=2)= \mathcal{N}(-2,4)$,
\item {\em Misclassification costs}: The misclassification costs $\mc=\{\mc_{\agent},\agent\in\agentset\}$ in the $\numagents$ environments are uniformly sampled from the interval $[5,25]$ for all states and actions in $\stateset$. Recall that we assume the continue cost is set to $1$ WLOG.
\end{compactitem}
}
\vspace{0.1cm}
\blue{
For every environment $\agent=1,2,3,4$, we computed $\mc_{\agent,\operatorname{MMV}},~\mc_{\agent,\operatorname{MMFE}}$ and $\mc_{\agent,\operatorname{Margin}}$, the point-valued IRL estimate of the agent's misclassification cost from MMV, MMFE and max-margin inverse SHT~(defined in \eqref{eqn:regul_irl_sht} with regularization parameter $\lambda=0$), respectively. For estimated misclassification cost $\mc_{\agent,\operatorname{est}}\in\{\mc_{\agent,\operatorname{MMV}},~\mc_{\agent,\operatorname{MMFE}},~\mc_{\agent,\operatorname{Margin}}\}$ with true cost $\mc_\agent$ and chosen stopping strategy $\stoptime_\agent$~(Lemma~\ref{lem:relativeoptimality}), the normalized IRL estimation error is defined as:
\begin{equation}\label{eqn:IRL_estimation_error}
    \text{IRL Estimation Error} = \frac{ \vert J(\stoptime_\agent,\mc_{\agent}) - J(\stoptime_\agent,\mc_{\agent,\operatorname{est}}) \vert }{J(\stoptime_\agent,\mc_{\agent})},
\end{equation}
where $J(\cdot)$ is the expected cumulative cost defined in \eqref{eqn:opt_stop_time}.}

\begin{figure}
    \centering
    \includegraphics[width = 0.8\columnwidth]{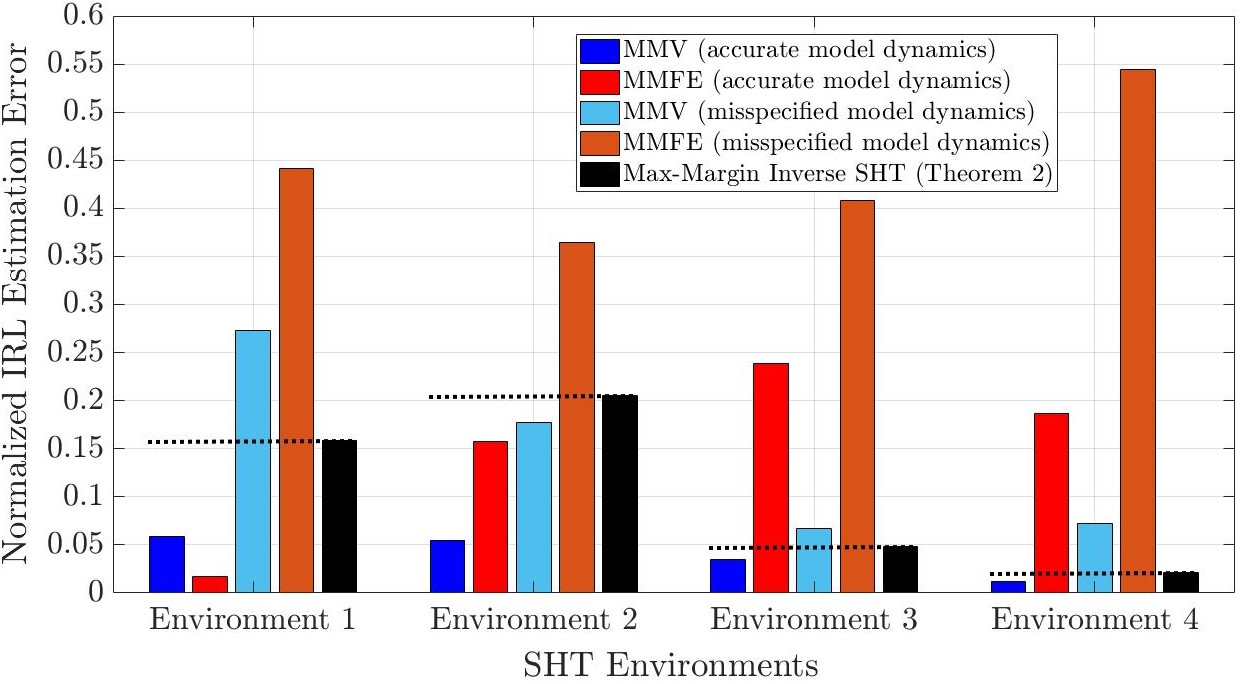}
    \caption{Inverse SHT Performance Comparison. Max-Margin NIAS-NIAC Test of Theorem~\ref{thrm:classic_SHT} versus MMV and MMFE~\citep{CH11}}    \label{fig:comparison_value_inv_SHT}
\end{figure}

\blue{Our experimental results are displayed in Fig.\,\ref{fig:comparison_value_inv_SHT}. Our results show that our proposed IRL algorithm yields a lower IRL estimation error~\eqref{eqn:IRL_estimation_error} than MMV and MMFE algorithms when model dynamics are misspecified. We observe that, on average, our max-margin IRL algorithm yields  $60\%$ lower estimation error compared to MMV and MMFE algorithms with misspecified model dynamics, and yields $27\%$ higher estimation error compared to MMV and MMFE algorithms with accurate model dynamics. 
}

\subsubsection*{Key Findings}
\blue{\noindent Our key findings from the numerical experiments\footnote{All our numerical results are completely reproducible and can be accessed from the GitHub repository \url{https://github.com/KunalP117/YouTube-Commenting-Analysis}} can be summarized as:
\begin{compactitem}
    \item For the case of perfect knowledge of model dynamics (case 1), we observed that the MMV and MMFE algorithms of \cite{CH11} perform better than max-margin IRL~\eqref{eqn:regul_irl_sht}, and yield
    approximately $\mathbf{27}\%$ lower IRL estimation error compared to max-margin IRL. This is expected since both MMV and MMFE have access to private information the forward learner uses for decision-making and hence generates a more accurate IRL estimate.
    \item When the model dynamics are misspecified  (case 2), our max-margin IRL algorithm outperforms both MMV and MMFE algorithms and yields approximately $\mathbf{60}\%$ lower IRL estimation error compared to MMV and MMFE.
\end{compactitem}
Indeed, when no assumptions are placed on the underlying POMDP structure like in \cite{CH11}, achieving IRL requires perfect knowledge of the model dynamics. Hence, MMV and MMFE fail when model dynamics are misspecified.
}

\subsubsection*{Perspective}
\blue{
Cases 1 and 2 highlight the fact that our approach is complementary to that of \cite{CH11}. \cite{CH11} achieve IRL where the model dynamics are perfectly specified (case 1). In comparison, our IRL methods yield necessary and sufficient methods for optimal Bayesian stopping when no knowledge of model dynamics is provided to the inverse learner. 
}

\subsection{Summary} 
Theorem~\ref{thrm:classic_SHT} specified necessary and sufficient conditions for identifying an optimal SHT agent acting in multiple environments. These conditions constitute a linear feasibility program that the inverse learner can solve to estimate SHT misclassification costs of the environments. 
The IRL task of solving the inverse SHT problem is more structured than the inverse optimal stopping problem in Sec.\,\ref{sec:BRP_RI}, \blue{since the agent's costs are partially known (expected continue cost is known) to the inverse learner. Hence, the feasible set of costs generated using Theorem~\ref{thrm:classic_SHT} is smaller than that generated by Theorem~\ref{thrm:NIAS_NIAC} for the inverse SHT problem. We also proposed an IRL algorithm for point-valued estimation of the environments' misclassification costs and illustrated its performance in Sec.\,\ref{sec:example_regul_IRL_SHT}. Our key finding is that this point-valued IRL algorithm reconstructs the misclassification costs with up to $95\%$ accuracy. Recall from Sec.\,\ref{sec:intro_applications} that an online user in multimedia platforms can be viewed as a Bayesian agent performing SHT. In the context of online multimedia platforms, the continue and stopping cost of the SHT agent can be viewed as the online user's sensing cost (attention to visual cues) and preference for viewing the online content, respectively. Hence, the numerical example in Sec.\,\ref{sec:example_regul_IRL_SHT} can be viewed as an IRL methodology to reconstruct an online user's preferences for advertisements/movie thumbnails by observing his/her actions in multiple environments (webpages). \blue{We illustrate this claim in Sec.\,\ref{sec:real-world} with an IRL analysis on a real-world dataset. Finally, in Sec.\,\ref{sec:comparison} we compared our inverse SHT algorithm to two existing algorithms in the literature for IRL for POMDPs, namely, MMV and MMFE~\cite{CH11}. Our key observation was that our inverse SHT algorithm outperforms MMV and MMFE in scenarios where the inverse learner has limited information about the forward learner, i.e.\,, the learner's model dynamics are misspecified.}
}

\section{Example 2. IRL for inverse search}\label{sec:Search}
In this section, we present a second example of IRL for an optimal Bayesian stopping time problem, namely, {\em inverse} Bayesian Search. In the  search problem, a Bayesian agent sequentially searches over a set of target locations until a static (non-moving) target  is found. The optimal search problem is a special case of a Bayesian multi-armed bandit problem, \blue{and also of the optimal Bayesian stopping problem discussed in Sec.\,\ref{sec:optimal_stopping_time} since the continue cost~\eqref{eqn:IRL_tuple_stop} is the cost of searching a location and the stopping cost is $0$ in the Bayesian search problem. Our IRL task in this section will be to estimate the search costs.
}

The optimal search problem is a modification of the sequential stopping problem in Sec.\,\ref{sec:BRP_RI} with the following changes:
\begin{compactitem}
    \item There is only $1$ stop action but multiple continue actions, namely, which of the 
    $\numstates$ locations to search at each time. We will call the continue actions as search actions, or simply, actions.
    \item The observation likelihood $\oprob$ depends both on the true state $\tstate$ and the continue action $\action$.
\end{compactitem}
Suppose an inverse learner observes the decisions of a Bayesian search agent over $\numagents$ search environments. The aim of the inverse  search problem is to identify if the search actions of the agent are optimal and if so, estimate their search costs. Our IRL result for Bayesian search (Theorem~\ref{thrm:Search} below)  gives a necessary and sufficient condition for identifying an optimal search agent (formalized in Lemma~\ref{lem:relativeoptimality_search} below) as equivalent to the existence of a feasible solution to a set of linear inequalities. 

\subsection{Optimal Bayesian search agent in multiple search environments}
Suppose an agent searches for a target  location $\state\in \stateset$. When the agent chooses action
$\action \in \stateset $ to search location $\action$,  it obtains an observation $\obs$. Assume the agent knows the set of conditional pmfs of $\obs$, namely,  $\{p(\obs|\tstate=\state),\state\in\{1,2,\ldots \numstates\}\}$. The aim of optimal search is to decide sequentially which location to search at each time to  minimize the cumulative search cost until the target is found.

We  define an optimal Bayesian search agent in $\numagents$ search environments as
\begin{equation}
\label{eqn:tuple_search}
    \optstoptuple = (\stateset,\prior,\obsset, \actionset,\revprobset, \{\searchcostsymbol_{\agent},\stoptime_{\agent},\agent\in\agentset\})
\end{equation}
where
\begin{compactitem}
    \item $\stateset = \{1,2,\ldots \numstates\}$ is a finite set of states (target locations). \item At time $0$, the true state $\tstate\in\stateset$ is sampled from prior pmf $\prior$. This location $\state$ is not known to the agent but is known to the inverse learner (performing IRL).
    \item $\obsset = \{0,1\}$, where $\obs=1$ (found)  and $\obs=0$ (not found) after searching a location.
    \item The set of actions $\actionset = \stateset$, $\action\in\actionset$ is the location searched by the agent. \item The Bayesian agent in search environments $\agent$ incurs instantaneous cost $\searchcostagent{\agent}>0$ for searching location $\action$.
    \item $\revprobset=\{\revprob,\action\in\actionset\}$, $\revprob$ is the reveal probability for location $a$, i.e.\,, the probability that the target is found when the agent  searches the target location ($\state=\action$) in search environment $\agent\in\agentset$. $\revprobset$ characterizes the action dependent observation likelihood $\oprob(\obs,\state,\action)$.
    \begin{equation}
      \oprob(\obs,\state,\action) = p(\obs|\state,\action) = \begin{cases}
      \revprob, &\obs=1,\state=\action\\
      1-\revprob, & \obs=0,\state=\action\\
      1,&\obs=0,\state\neq\action.
      \end{cases}
    \end{equation}
    For IRL identifiability, we assume that the reveal probabilities are the same for all search environments in $\agentset$.
    \item $\{\stoptime_{\agent},\agent\in\agentset\}$ are the optimal search strategies of the Bayesian agent over all environments in $\agentset$, when the agent operates sequentially on a sequence of observations $\obs_1,\obs_2,\ldots$ as discussed below in Protocol~\ref{prtcl_decision_search}.
\end{compactitem}

\begin{protocol}
\label{prtcl_decision_search}
Sequential Decision-making protocol for Search:
\begin{enumerate}
    \item Generate $\tstate\sim\prior$ at time $\timeinst=0$.
    \item At time $\timeinst\geq1$, agent records observation $\obs_{\timeinst}\sim\oprob(\cdot,\action_{\timeinst-1},\tstate)$.
    \item If $\obs_{\timeinst}=1$, then stop. Otherwise, if $\obs_{\timeinst}=0$:\\
    (i) Update belief $\belief_{\timeinst-1}\rightarrow \belief_{\timeinst}$ (described below). \\
    (ii) For search policy $\stoptime$, agent takes action $\action_{\timeinst}=\stoptime(\belief_{\timeinst})$. (Note the first action is taken at time $\timeinst=0$, while the first observation is at $\timeinst=1$).\\
    (iii) Set $\timeinst=\timeinst+1$ and go to Step $2$.
\end{enumerate}
{\em Belief Update:} Let $\mathcal{F}_{\timeinst}$ denote the sigma-algebra generated by the action and observation sequence $\{\action_1,\obs_1,\ldots\action_{\timeinst},\obs_{\timeinst}\}$. The agent updates its belief $\belief_{\timeinst}=\prob(\tstate=\state|\mathcal{F}_t),~\state\in\stateset$ using Bayes formula as
\begin{equation}
\label{eqn:beliefupdate_search}
    \belief_{\timeinst} = \frac{\oprob(\obs_{\timeinst},\action_{\timeinst-1})\belief_{\timeinst-1}}{\mathbf{1}'\oprob(\obs_{\timeinst},\action_{\timeinst-1})\belief_{\timeinst-1}},
\end{equation}
where $\oprob(\obs,\action)=\operatorname{diag}(\{\oprob(\obs,\state,\action),\state\in\stateset\})$. The belief $\belief_{\timeinst}$ is an $\numstates-$dimensional probability vector belonging to the $(\numstates-1)$ dimensional unit simplex (\ref{eqn:simplex_stopping}).
\end{protocol}
{\em Remark:} The search agent's stopping region is simply the set of distinct vertices of the $\numstates-1$ dimensional unit simplex.

We define the random variable $\funcstop$ as the time when the agent stops (target is found).
\begin{equation}
    \funcstop = \inf~\{\timeinst>0|~\obs_{\timeinst}=1\}
\end{equation}
Clearly, the set $\{\funcstop=\timeinst\}$ is measurable wrt $\mathcal{F}_t$, hence, the random variable $\funcstop$ is adapted to the filtration $\{\mathcal{F}_\timeinst\}_{\timeinst\geq0}$. Below, we define the optimal search strategies $\{\stoptime_{\agent},\agent\in\agentset\}$.

\begin{definition}[Search strategy optimality]\label{def:absoptimality_search}
The optimal search strategy $\stoptime_\agent$ of the Bayesian agent operating according to Protocol~\ref{prtcl_decision_search} in environment $\agent\in\agentset$ that minimizes the agent's cumulative expected search cost 
is well known \citep{VK16} to be a stationary policy as defined below:  
\begin{align}
    & \netobjfun(\stoptime_{\agent},\searchcostsymbol_{\agent}) = \min_{\stoptime}\netobjfun(\stoptime,\searchcostsymbol_{\agent}) =  \E_{\stoptime}\left\{\sum_{\timeinst=0}^{\funcstop-1} \searchcostsymbol_{\agent}(\stoptime(\belief_{\timeinst}))\right\}, \label{eqn:netobj_search}\\
    & \stoptime_{\agent}(\belief) = \argmax_{\action\in\actionset} \left( \frac{\belief(\action)\revprobsymb}{\searchcostsymbol_{\agent}(\action)}\right) \label{eqn:opt_policy_search}.
\end{align}
Here, $\E_{\stoptime}\{\cdot\}$ denotes expectation parametrized by $\stoptime$ induced by the probability measure $\{\action_{\timeinst},\obs_{\timeinst+1}\}_{\timeinst=1}^{\funcstop-1}$, $\netobjfun(\cdot)$ denotes the expected search cost and $\stoptime$ belongs to the class of stationary search strategies.  
\end{definition}
{\em Remarks.} (1) Note that the minimization in  \eqref{eqn:netobj_search} is over stationary search strategies. It is well known that the optimal search strategy has a threshold structure \citep{VK16}. Since the set of all threshold strategies forms a compact set,  we can replace the  `$\inf$' in (\ref{eqn:opt_stop_time}) for generic optimal stopping problems by `$\min$' in (\ref{eqn:netobj_search}).\\
(2) Since the expected cumulative cost of an agent depends only on the search costs (for constant reveal probabilities), we can set $\searchcostsymbol_{\agent}(1)=1,~\forall \agent\in\agentset$ WLOG.


\subsection{IRL for inverse search. Main result}
In this subsection, we provide an inverse learner-centric view of the Bayesian stopping time problem and the main IRL result for inverse search.
Suppose the inverse learner observes a search agent taking actions over $\numagents$ search environments where the agent performs several independent trials of Protocol \ref{prtcl_decision_search} for Bayesian sequential search in each environment.  
We make the following assumptions about the  inverse learner performing IRL to identify if $\agentset$ comprises an optimal search agent.
\begin{enumerate}[label=(A\arabic*)]
\setcounter{enumi}{\value{assumindex}}
\item \label{asmp:IRL_search_1} The inverse learner knows the dataset
\begin{equation}\label{eqn:dataset_search}
  \datainfsearch =   (\prior,\{\occagent{\agent},\agent\in\agentset\}).
\end{equation}
Here, $\occagent{\agent}$ is the average number of times the agent searches location $\action$ when the target is in $\state$ in environment $\agent$:
\begin{equation}
    \occagent{\agent} = \E_{\stoptime_{\agent}}\left\{\sum_{t=1}^{\funcstop} 
    \mathbbm{1}\{\stoptime_{\agent}(\belief_{\timeinst}) = \action\} |\state
    \right\}.
\end{equation}
We call $\occagent{\agent}$ as the  agent's {\em search action policy} in search environment $\agent$.
\item \label{asmp:IRL_search_2} In dataset $\datainfsearch$, there are at least $\numagents\geq 2$ environments with distinct search costs.
\setcounter{assumindex}{\value{enumi}}
\end{enumerate}

Assumption \ref{asmp:IRL_search_1} is discussed  after the main result. In complete analogy to \ref{asmp:IRL2}, assumption \ref{asmp:IRL_search_2} is needed for identifiability of the search costs. We emphasize that the inverse learner only has the average number of times the agent searches a particular location in any environment. The inverse learner does not know the stopping time or the order in which the agent search the locations. \blue{In completely analogy to Lemma~\ref{lem:relativeoptimality}, 
Lemma~\ref{lem:relativeoptimality_search} below specifies the inverse learner's identifiability of an optimal search agent under assumptions \ref{asmp:IRL_search_1} and \ref{asmp:IRL_search_2}:
\begin{lemma}[IRL identifiability of optimal Bayesian search agent]
\label{lem:relativeoptimality_search} 
The inverse learner identifies the tuple $\optstoptuple$~\eqref{eqn:tuple_search} as an optimal Bayesian search  agent iff \eqref{eqn:relopt_search} holds.
\begin{equation}
     \netobjfun(\stoptime_{\agent},\searchcostagent{\agent})  \leq  \netobjfun(\stoptime_{\agenttwo},\searchcostagent{\agent}),~\forall~\agent,\agenttwo\in\agentset,~\agent\neq\agenttwo. \label{eqn:relopt_search}
\end{equation}
In complete analogy with (\ref{eqn:relopt_stop_time_continue_cost}) in Lemma~\ref{lem:relativeoptimality} for identifying an optimal stopping time agent, $\netobjfun(\cdot)$ in the above equation is the expected cumulative search cost of the agent.
\end{lemma}
}
\blue{We omit the proof of Lemma~\ref{lem:relativeoptimality_search} since it is identical to that of Lemma~\ref{lem:relativeoptimality}.
Eq.\,\ref{eqn:relopt_search} in Lemma~\ref{lem:relativeoptimality_search} is analogous to \eqref{eqn:relopt_stop_time_continue_cost} in Lemma~\ref{lem:relativeoptimality}. The inverse learner simply checks if the expected cumulative search cost for environment $\agent$ is the smallest possible given the finite strategies $\{\stoptime_\agent,\agent\in\agentset\}$.} We are now ready to present our main IRL result for the inverse search problem. The result specifies a set of linear inequalities that are simultaneously necessary and sufficient for a search agent's actions in multiple environments $\agentset$ to be identified as that of an optimal search agent (\ref{eqn:relopt_search}). 

\begin{theorem}[IRL for inverse Bayesian search]\label{thrm:Search}
Consider the inverse learner with dataset $\datainfsearch$ (\ref{eqn:dataset_search}) obtained from a search agent acting in multiple environments $\agentset$. Assume \ref{asmp:IRL_search_1} holds. Then:\\
{\em 1.} \underline{Identifiability:} The inverse learner can identify if the dataset $\datainfsearch$ is generated by an optimal search agent (Definition~\ref{lem:relativeoptimality_search}).\\
{\em 2.} \underline{Existence:} There exists an optimal search agent parameterized by tuple $\optsearchtuple$~\eqref{eqn:tuple_search} if and only if there exists a feasible solution to the following linear (in search costs) inequalities:
\begin{align}
    &\text{Find }\searchcostagent{\agent}\in\reals_+,\searchcostsymbol_{\agent}(1)=1\quad\text{s.t.}\quad\operatorname{NIAC}^{\dagger}(\datainfsearch)\leq 0,~\text{where}\nonumber\\
    &\operatorname{NIAC}^{\dagger}:~\sum_{\state\in\stateset} \prior(\state)(\occagent{\agent}-\occagent{\agenttwo})~\searchcostagent{\agent}< 0~\forall\agent,\agenttwo\in\agentset,~\agent\neq \agenttwo.  \label{eqn:NIAC_dag_def}
\end{align}
{\em 3.} \underline{Reconstruction:} The set-valued IRL estimate of the agent's search costs in environments $\agentset$ is the set of all feasible solutions to the NIAC$^{\dagger}$ inequalities. \qedsymbol
\end{theorem}

The proof of Theorem~\ref{thrm:Search} is in Appendix~\ref{appdx:proof_Search}. Theorem~\ref{thrm:Search} provides a set of linear inequalities whose feasibility is equivalent to identifying the optimality of a Bayesian search agent in multiple environments with different search costs. Note that Theorem~\ref{thrm:Search} uses the search action policies $\{\actselectagent{\agent},\agent\in\agentset\}$ to construct the expected cumulative search costs of the agent in multiple environments and verify if the inequality for identifying optimality (\ref{eqn:relopt_search}) for Bayesian search holds. The key idea for the IRL result is to express the expected cost of the search agent in environment $\agent$ in terms of its chosen search action policy $\occagent{\agent}$~\eqref{eqn:dataset_search}.
Algorithms for linear feasibility such as the simplex method~\citep{BD04} can be used to check feasibility of (\ref{eqn:NIAC_dag_def}) in Theorem~\ref{thrm:Search} and construct a feasible set of search costs for the optimal search agent.

\noindent {\em Discussion of assumption \ref{asmp:IRL_search_1}.}
To motivate \ref{asmp:IRL_search_1}, suppose for each environment $\agent\in\agentset$, the inverse learner records the state $\state_{\trial,\agent}$ and agent actions $\{\action_{1:\funcstop_{\trial,\agent},\trial,\agent}\}$ over $\trial=1,2,\ldots \numtrials$ independent trials. Then, the variable $\occagent{\agent}$ in $\datainfsearch$ (\ref{eqn:dataset_search}) is the limit pmf of the empirical pmf $\occagentemp{\agent}$ as the number of trials $\numtrials\to\infty$.
\begin{equation}
    \label{eqn:act_search_policy}
    \occagentemp{\agent} = \frac{\sum_{\trial=1}^{\numtrials} \sum_{\timeinst=1}^{\funcstop_{\trial,\agent}} \mathbbm{1}\{\state_{\trial,\agent}=\state,\action_{\timeinst,\trial,\agent}=\action\} }{\sum_{\trial=1}^{\numtrials} \mathbbm{1}\{\state_{\trial,\agent}=\state\}}.
\end{equation}
In complete analogy to Sec.\,\ref{sec:disc_assumpt}, almost sure convergence holds by Kolmogorov's strong law of large numbers. 
$\occagent{\agent}$ is the average number of times the agent searches location $\action$ when the target is in location $\state$ in environment $\agent$. More formally, for a fixed state $\state$, $\occagent{\agent}$ is the number of times the posterior belief of the agent visits the region in the unit simplex of pmfs where it is optimal to choose action $\action$. In Appendix~\ref{appdx:proof_Search}, we discuss how the search action policy $\occagent{\agent}$ can be used to express the  agent's cumulative expected search cost~(\ref{eqn:netobj_search}) in the $\agent^{\text{th}}$ environment.

{\em Remark:} Analogous to the action selection policy (\ref{eqn:actionselection}) for stopping problems with multiple stopping actions, the inverse learner uses the search action policy to identify Bayes optimality in stopping problems with multiple continue actions (and single stop action).

\subsection{Numerical example illustrating IRL for inverse search}
\label{sec:Example_Search}
\blue{We now present a  numerical example for inverse search with $3$ search environments and $3$ search locations. The aim of this example is to illustrate the consistency property of Theorem~\ref{thrm:Search}. That is, that the true search costs lie in the set of feasible costs computed by the inverse learner by solving the feasibility test of Theorem~\ref{thrm:Search}.}

{\em Search environments.} We consider $\numagents=3$ search environments with:
\begin{compactitem}
    \item {\em Prior} $ \prior = [1/3~1/3~1/3]'$.
    \item {\em Search locations:} $\numstates=\numactions=3$.
    \item {\em Reveal probability}: $\revprobsymb(1)=0.7,\revprobsymb(2)=0.68,\revprobsymb(3)=0.6$.
    \item {\em Search costs}:\\
    Environment 1: $\searchcostsymbol_{1}(1)=1,~\searchcostsymbol_{1}(2)=3,~\searchcostsymbol_{1}(3)=4$, \\
    Environment 2: $\searchcostsymbol_{2}(1)=1,~\searchcostsymbol_{2}(2)=1,~\searchcostsymbol_{2}(3)=2$, \\
    Environment 3: $\searchcostsymbol_{3}(1)=1,~\searchcostsymbol_{3}(2)=0.5,~\searchcostsymbol_{3}(3)=3$.
\end{compactitem}
(Recall that WLOG the search cost $\searchcostsymbol_{\agent}(1)$ can be set to $1$ for all $\agent\in\{1,2,3\}$.)

{\em Inverse Learner specification.} Next we  consider the inverse learner.
We generate $\numtrials=10^6$ samples for the search agent in all $3$ environments using the above parameters. Recall from Theorem~\ref{thrm:Search} that the inverse learner uses the dataset $\datainf(\Search)$ to perform IRL for search. Here \vspace{-0.25cm}
\begin{equation}
\label{eqn:example_Search_dataset}
    \datainf(\Search) = (\prior, (\occagentemp{\agent},\agent\in\{1,2,3\}),
\end{equation}
where $\numtrials=10^6$, the second term in the dataset is the empirically calculated search action policy (\ref{eqn:act_search_policy}) of the agent in environment $\agent$ from the $10^6$ generated samples. 

\begin{figure}[t]
\centering
\includegraphics[width=0.4\textwidth]{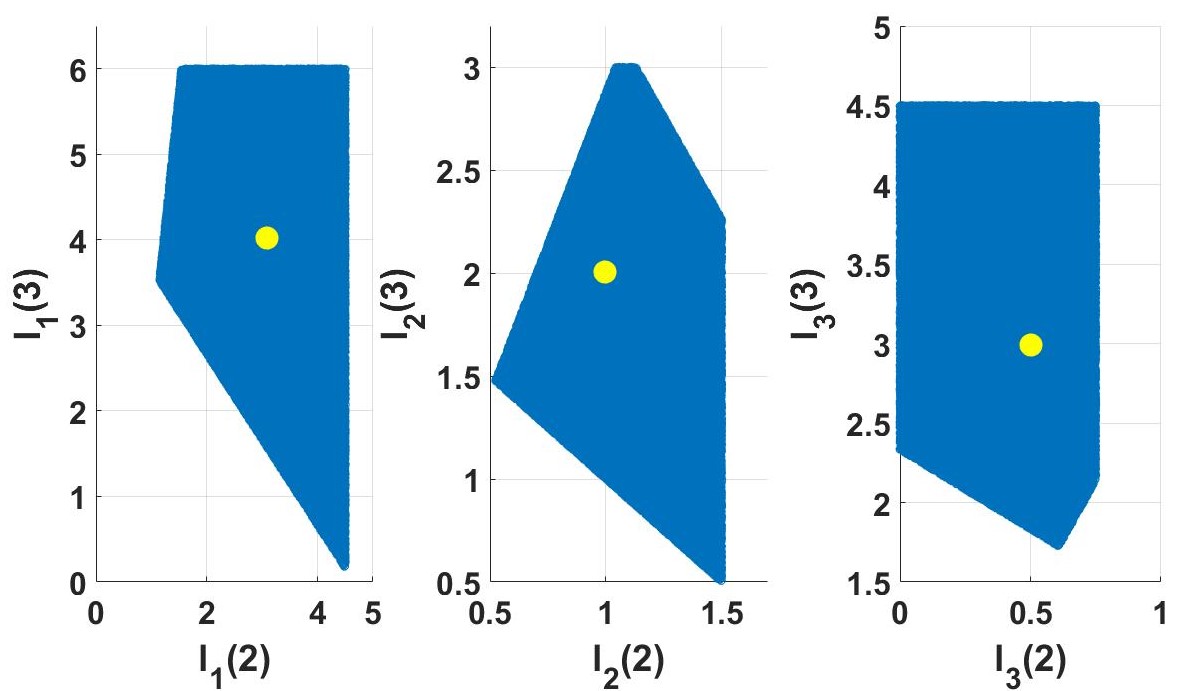}
\caption{Numerical example for inverse search with parameters specified in Sec.\,\ref{sec:Example_Search}. The key observation is that the true search costs (yellow points) lie in the feasible set~(blue region) of costs  computed via Theorem~\ref{thrm:Search}. This follows from the necessity proof of Theorem~\ref{thrm:Search} which says if the agent is an optimal search agent, then the true costs lie in the feasible set of costs that satisfy the NIAC$^{\dag}$ inequalities.
}
\label{fig:results_search}
\end{figure}

{\em IRL Result.} The inverse learner performs IRL by using the dataset $\datainf(\Search)$ (\ref{eqn:example_Search_dataset}) to solve the linear feasibility problem in Theorem~\ref{thrm:Search}. The result of the feasibility test is shown in Fig.\,\ref{fig:results_search}. The blue region is the set of feasible search costs for each environment. 
The feasible set of costs is $\{(\searchcostsymbol_{\agent}(2),\searchcostsymbol_{\agent}(3),\agent\in\{1,2,3\}\}\subseteq \mathbb{R}^6_+$.
For visualization purposes, Fig.\,\ref{fig:results_search} displays the feasible search costs for each environment in a different sub-figure. In complete analogy to Fig.\,\ref{fig:results}, the feasible search costs for each environment are shown in each sub-figure by keeping the search costs of the other $2$ environments fixed at their true values. The true search cost for every environment is highlighted by a yellow point. The key observation is that these true costs belong to the set of feasible costs (blue region) computed via Theorem~\ref{thrm:Search}. Thus, Theorem~\ref{thrm:Search} successfully performs IRL for the search problem and the set of feasible search costs can be reconstructed as the solution to a linear feasibility problem. 

\section{Inverse Optimal Stopping for Predicting YouTube Commenting Behavior}\label{sec:real-world}
\blue{In this section, we illustrate our IRL results for Bayesian stopping time problems on a real-world YouTube dataset. Although we use the same dataset in previous work~\citep{HKP20}, our IRL methodology and experimental results are new. For brevity, we discuss the key differences compared to \citet{HKP20} and justify our choice of Bayesian stopping for modeling user engagement on YouTube in Appendix~\ref{appdx:youtube}.}

\blue{We consider a YouTube dataset comprising approximately $140000$ videos across $25,000$ channels spanning $18$ video categories and over $9$ millions users from April 2007 to May 2015. The diversity of videos in YouTube is immense; it is intuitive to exploit this diversity for understanding how groups of YouTube users exposed to different classes of video content engage differently with YouTube. Hence, by analyzing groups of YouTube users indexed by video category, our aim is to:\\ (1) Identify if YouTube user engagement is consistent with Bayesian optimal stopping, and if so,\\
(2) Reconstruct the stopping costs of user engagement using the IRL results in this paper, and\\
(3) Use the reconstructed costs to predict user engagement in videos. }
 
\blue{Our YouTube dataset does not contain any information (visual cues) about what the human user perceives from the video webpage before choosing to engage on the YouTube platform. Recall from Theorem~\ref{thrm:NIAS_NIAC} that our IRL approach does not depend on the unobserved model dynamics that generate the IRL dataset~\eqref{eqn:IRL_tuple}. This makes our IRL methodology well-suited to scenarios where the parameters of the underlying decision making process are not available in the IRL dataset. Our main conclusions from our IRL analysis of the YouTube dataset can be summarized as:
\begin{compactitem}
\item  YouTube user engagement is consistent with optimal Bayesian stopping. Based on our IRL analysis on
groups of YouTube users, where each group consists of approximately $3500$ viewers,  the YouTube dataset (described below in \eqref{eqn:YT_dataset}) satisfies the NIAS and NIAC feasibility inequalities of Theorem~\ref{thrm:NIAS_NIAC} for optimal Bayesian stopping with a high margin.
\item By choosing two representative points from the feasible set of costs generated by IRL~\eqref{eqn:NIAS_def},~\eqref{eqn:NIAC_def}, namely, max-margin estimate and entropy-regularized estimate defined below, we show our reconstructed IRL costs predict user engagement with high accuracy. Figure~\ref{fig:youtube_results} illustrates the predictive performance of our IRL methodology.
\end{compactitem}
}

\subsection{YouTube Dataset and Model Parameters}\label{sec:YT-parameters}
\begin{figure}[ht]
	\centering
	\vspace{-9pt}
	\includegraphics[width=0.75\textwidth]{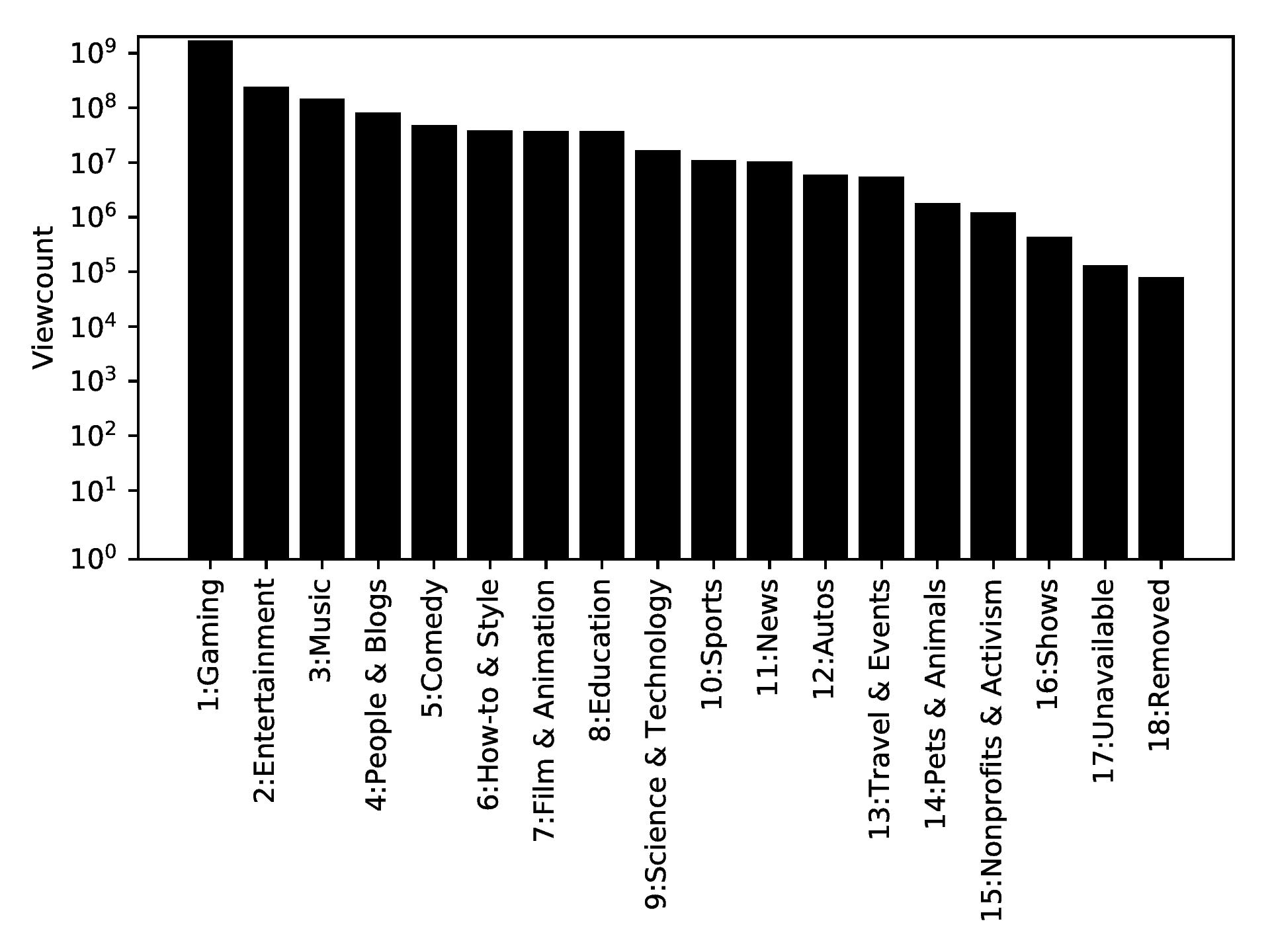}
	\vspace{-10pt}
	\caption{YouTube Dataset Overview. Viewcount summed over all videos (vertical axis) of $\numagents = 18$ video categories. The $18$ categories are listed on the horizontal axis.}
	\label{fig:categoryandviewcount}
\end{figure}

\blue{Categories in YouTube (e.g.\, News, Gaming, Music etc.) are numbered from $1 - 18$ (See Fig.~\ref{fig:categoryandviewcount} for the full listing). The video categories have mean numbers of users ranging from $149$ to $4596$  for high viewcount (greater than $10000$) videos and $8$ to $1801$ for low viewcount videos (less than $10000$). Figure~\ref{fig:categoryandviewcount} lists each video category along with the total number of views. Note that the video categories ``Unavailable'' or ``Removed'' are videos flagged by YouTube as being suspected of violating YouTube's video policies.}

\blue{The YouTube dataset contains the view counts, comment counts, likes, dislikes, thumbnail, title, and category of each video. To relate to our main IRL result of Theorem~\ref{thrm:NIAS_NIAC}, we define the following:\newline
\textit{$1$. Agent}: Group of users interacting with videos in each video segment. User engagement in different video categories can be interpreted as the agent acting in multiple environments. In the rest of the section, we will use the terms `user engagement' and `commenting behavior' interchangeably.\newline 
\textit{$2$. State ($\state$)}: In the YouTube dataset, the state $\state$ of each video is the viewcount $1$ day after the video was published. Specifically, state $\state=1$ is high viewcount (more than $10,000$ views) and $\state=2$ otherwise. In YouTube, video viewcount is the  independent quantity which governs the commenting behavior since videos need to be viewed first before users can comment or rate the video. \newline
\textit{$3.$ Terminal Action ($\action$)}: In the YouTube dataset, the terminal action $\action$ is related to the overall commenting behavior\footnote{By overall commenting behavior in YouTube, we mean both the comment count and the video ratings (likes and dislikes). Another term used in the literature \citep{KH17} is ``user engagement''.} of the users, which is computed using the comment counts, like count, and dislike count $2$ days after the video is published. 
The possible actions are: $\action=1$ denotes low comment count with negative sentiment, $\action=2$ denotes low comment count with neutral sentiment, $\action=3$ denotes low comment count with positive sentiment, $\action=4$ denotes high comment count with negative sentiment, $\action=5$ denotes high comment count with neutral sentiment, and $\action=6$ denotes high comment count with positive sentiment. Here negative sentiment occurs if the difference between the like count and dislike count is less than $-25$, neutral sentiment occurs if the difference lies between $-25, 25$, and positive sentiment occurs if the difference is greater than $25$. A low comment count is said to occur if there are less than $100$ comments, otherwise the comment count is defined to be high. 
\newline
\textit{$4$. Observation ($\obs$)}: The observation $\obs$ for a YouTube user abstracts the visual cues a user perceives that depends on video metadata such as thumbnail, title, category etc. The observation likelihood is indicative of the attention expended by the user on a video. We note that although neither the observations $\obs$ nor the observation likelihood $p(\obs|\state)$ are contained in the YouTube dataset, our IRL algorithm abstracts away these unobserved model parameters, and still yields necessary and sufficient conditions for Bayes optimality.
\\
\textit{$4$. Environment  ($\agent$)}: Environment $\agent$ corresponds to each of the $\numagents = 18$ video categories in our YouTube dataset. Fig.~\ref{fig:categoryandviewcount} lists each video category with the total number of views. Note that the video categories ``Unavailable'' or ``Removed'' are videos flagged by YouTube as being suspected of violating YouTube's video policies\footnote{Refer to~\url{https://www.youtube.com/yt/about/policies/\#community-guidelines} for details}.}

\blue{Recall from Sec.\,\ref{sec:IRL_Stopping} that the inverse learner requires knowledge of the dataset $\datainf =   (\prior,\actselectset)$~\eqref{eqn:IRL_tuple} for identifying optimal Bayesian stopping via Theorem~\ref{thrm:NIAS_NIAC}. In the YouTube context, the variables $\prior,\actselectset=\{\actselectagent{\agent},\agent\in\agentset\}$ dataset  $\datainf$ can be constructed as:\\
\begin{equation}
\begin{split}
\prior(\state) &= \frac{1}{I}\sum_{i=1}^I \mathbbm{1}\{\state_{i} = \state\},~~\actselectagent{\agent} = \frac{\sum_{i=1}^I\mathbbm{1}\{\state_i = \state, \action_i = \action, \operatorname{category}_i=\agent\}}{\sum_{i=1}^{I}\mathbbm{1}\{\state_i=\state, \operatorname{category}_i = \agent\}},
\end{split}
\label{eqn:YT_dataset}
\end{equation}
where $\mathbbm{1}\{\cdot\}$ is the indicator function, variable $i$ indexes the YouTube videos, $I=140000$ is the total number of YouTube videos in the dataset, and environment $\agent\in\{1,2,\ldots,18\}$ indexes the video categories. Also, $\state_i, \action_i, \operatorname{category}_i$ denote the state, action and category of the YouTube video indexed by $i$, where the state and action interpretations for the YouTube videos are discussed above.}

\blue{\subsection{YouTube Data Analysis Results}
We now discuss our experimental findings from our IRL analysis on the YouTube dataset.\footnote{All our numerical results are completely reproducible and can be accessed from the GitHub repository \url{https://github.com/KunalP117/YouTube-Commenting-Analysis}.} Our main task is to predict YouTube's commenting behavior, that is, the action selection policy $\actselectagent{\agent}$ in video category $\agent$ using the IRL algorithms in this paper. Our first observation is that the dataset $\datainf$~\eqref{eqn:YT_dataset} comprising YouTube commenting behavior over $\numagents = 18$ categories  passes the convex feasibility test~\eqref{eqn:NIAS_def} and \eqref{eqn:NIAC_def} of Theorem~\ref{thrm:NIAS_NIAC} with a high margin of $1.85\times 10^{-3}$, where the margin is normalized by the maximum feasible cost $\max_{\agent,\state,\action}\utilityagent{\agent}$. This shows that there exists a Bayesian stopping model that rationalizes YouTube commenting behavior.}

\blue{We now illustrate how well the reconstructed costs from the feasibility test of Theorem~\ref{thrm:NIAS_NIAC} predict the commenting behavior of YouTube videos in different categories. For our prediction task, first, we randomly divided the YouTube dataset into two parts - training data ($80\%$) and testing data ($20\%$). Also, we consider only a subset of the $18$ video categories for which the number of videos exceeds $200$. This extra condition results in $9$ out of $18$ video categories considered for our IRL prediction analysis. For predicting commenting behavior via IRL, we first consider the training data and compute two point-valued estimates of the agent stopping costs that satisfy the NIAS and NIAC inequalities of Theorem~\ref{thrm:NIAS_NIAC}, namely, max-margin IRL and entropy-regularized IRL defined below:
\begin{align}
    &\textit{Max-Margin IRL}:\nonumber\\
    & \{\utilvec_{\operatorname{MM-IRL}},~\epsilon^\ast\}~=~\argmax_{\epsilon\geq 0,\boldsymbol{\utilitysymbolcaps}\geq \mathbf{0}} \epsilon,\text{ such that }\operatorname{NIAS}(\datainf,\boldsymbol{\utilitysymbolcaps}) \leq - \epsilon,~\operatorname{NIAC}(\datainf,\boldsymbol{\utilitysymbolcaps}) \leq - \epsilon,\label{eqn:maxmargin-irl}\\
    &\textit{Entropy-Regularized IRL}:\nonumber\\
    & \utilvec_{\operatorname{Ent-IRL}}~=~\text{Any feasible cost }\utilvec\equiv\{\utilityagent{\agent},\state\in\stateset,\action\in\actionset\}_{\agent=1}^\numagents,~\text{ that satisfies }\label{eqn:entropy-irl}\\
    &~(a)~~\utilitysymbolagent{\agent}(\state,\action_1) = 1,~\forall~\state\in\stateset~(\text{Normalization}),~\text{and}\nonumber\\
    &(b)~~\operatorname{NIAS}(\datainf,\boldsymbol{\utilitysymbolcaps}) \leq 0,~\operatorname{NIAC}(\datainf,\boldsymbol{\utilitysymbolcaps}) \leq 0,~\sumcost(\datainf,\boldsymbol{\utilitysymbolcaps},\{MI(\prior;\actselectagent{\agent})\}_{\agent=1}^\numagents) \leq 0.\nonumber
\end{align}
In \eqref{eqn:maxmargin-irl} and \eqref{eqn:entropy-irl} above, $\boldsymbol{\utilitysymbolcaps}=\{\utilityagent{\agent},\agent\in\agentset,\state\in\stateset,\action\in\actionset\}$ denotes the set of stopping costs over all environments $\agentset$, states $\stateset$ and actions $\actionset$; the NIAS, NIAC and SUMCOST feasibility inequalities are defined in \eqref{eqn:NIAS_def},~\eqref{eqn:NIAC_def} and \eqref{eqn:sumcostfeasible}, respectively. In \eqref{eqn:entropy-irl}, $MI(\prior;\actselectagent{\agent})$ denotes the mutual information between the agent's prior $\prior$ and action selection policy $\actselectagent{\agent}$ defined as:
\begin{equation*}  MI(\prior;\actselectagent{\agent}) = \sum_{\state,\action}\prior(\state)~\actselectagent{\agent} ~\operatorname{log}\left(\frac{\actselectagent{\agent}}{\sum_{\state} \prior(\state)~\actselectagent{\agent}}\right)
\end{equation*}
The intuition behind \eqref{eqn:maxmargin-irl} is clear: choose the stopping costs that pass the feasibility inequalities of Theorem~\ref{thrm:NIAS_NIAC} with the largest margin. In \eqref{eqn:entropy-irl}, we impose the additional constraint that the expected continue cost is the mutual information between the prior and the action selection policy. The inspiration for this information-theoretic cost stems from the seminal work of \cite{Sim03} who modeled human attention as a limited-capacity communication channel, and from Max-Entropy IRL~\citep{ZB08} in IRL literature. Eq.\,\ref{eqn:entropy-irl} yields a softmax structure for the feasible stopping costs~(see Appendix~\ref{appdx:NIAS-NIAC-remarks} for a more detailed explanation); the key idea is that entropy-regularized IRL for Bayesian stopping yields a set of constant stopping costs, constant up to an affine monotone transformation.}

\blue{For predicted cost $\{\utilityagent{\agent},\state\in\stateset,\action\in\actionset,\agent\in\agentset\}$ and action selection policies $\{\actselectagent{\agent},\agent\in\agentset\}$ from the training dataset, the predicted action selection policy $\actselectagentemp{\agent}$ for the test dataset is straightforwardly computed as:
\begin{equation}\label{eqn:predactselect}
\hat{p}_\agent(\action|\state) = \sum_{\action'}~\mathbbm{1}\{\action = \argmax_{b} \sum_{\state} \hat{p}_\agent(\state|\action')~\utilitysymbol_\agent(\state,b)\}~ p_\agent(\action'|\state),\text{ where}
\end{equation}
the probability $\hat{p}_\agent(\state|\action')=\frac{\pi_{0,\text{test}}(\state)~\actselectagent{\agent}} {\sum_{\state} \pi_{0,\text{test}}(\state)~\actselectagent{\agent}}$ is the predicted posterior belief of the state given action $\action'$ for the test dataset. Observe that all terms in the RHS of \eqref{eqn:predactselect} pertain to the training dataset except for the prior $\pi_{0,\text{test}}$ that is empirically computed from the test dataset. Intuitively, \eqref{eqn:predactselect} assumes the observation likelihood for the YouTube user in the test dataset is simply the action selection policy $\actselectagent{\agent}$ from the training dataset. In words, the predicted action selection policy $\actselectagentemp{\agent}$ in \eqref{eqn:predactselect} is obtained by simply summing the likelihoods of all actions $\action'\in\actionset$ for which action $\action$ is optimal given posterior belief $\hat{p}(\state|\action')$. }

\begin{figure}[ht]
    \centering    \includegraphics[width =1 \columnwidth]{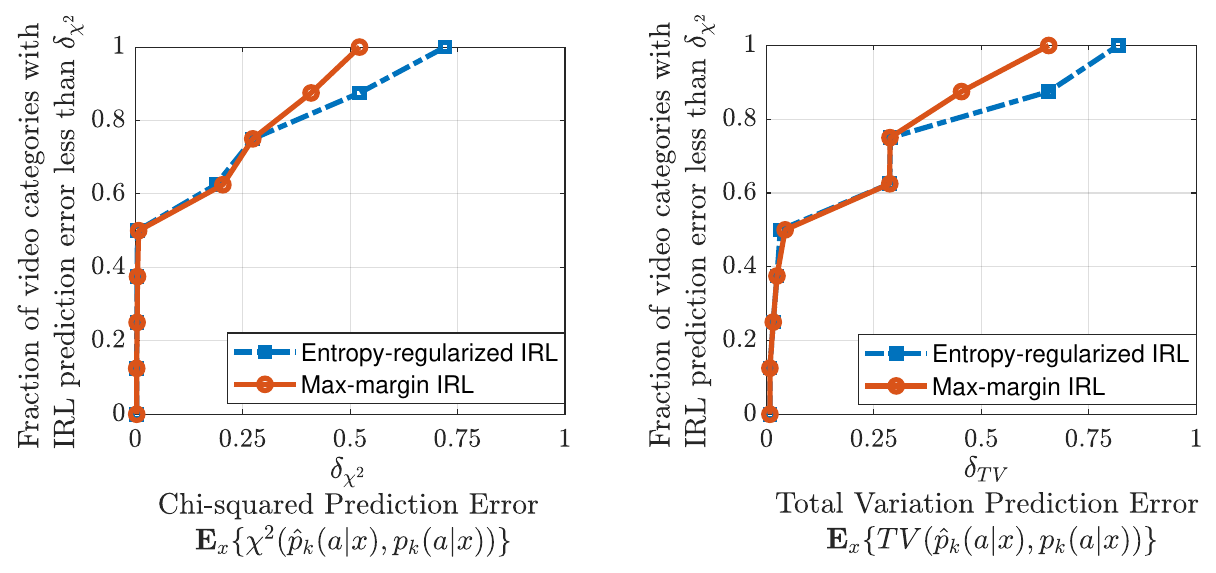}
    \caption{IRL Prediction Error for YouTube Dataset. The main takeaway is that point-valued IRL estimates that satisfy the feasibility test of Theorem~\ref{thrm:NIAS_NIAC} predict YouTube commenting behavior with high accuracy (low statistical distance between true and predicted distributions). For reconstructing the stopping costs, we choose two distinct point-valued stopping costs, namely, entropy-regularized IRL~\eqref{eqn:entropy-irl} and max-margin IRL~\eqref{eqn:maxmargin-irl} and performed our numerical experiments on 9 out of 18 video categories for which the video count exceeded 250. For both sets of estimated stopping costs, we observed that for 7 out of 9 YouTube video categories considered for analysis, both the chi-squared distance and total variation distance between the true and predicted action policy is less than 0.3.}
    \label{fig:youtube_results}
\end{figure}

\blue{Using \eqref{eqn:predactselect}, we obtained two sets of predicted action selection policies, namely, $\hat{\boldsymbol{p}}_{\operatorname{MM-IRL}}=\{\actselectagentemp{\agent},\agent\in\agentset\}_{\operatorname{MM-IRL}}$ and $\hat{\boldsymbol{p}}_{\operatorname{Ent-IRL}}=\{\actselectagentemp{\agent},\agent\in\agentset\}_{\operatorname{Ent-IRL}}$ for the test dataset, corresponding to stopping costs $\utilvec_{\operatorname{MM-IRL}}$~\eqref{eqn:maxmargin-irl} and $\utilvec_{\operatorname{Ent-IRL}}$~\eqref{eqn:entropy-irl}, respectively. To comment on the prediction accuracy, we computed the chi-squared distance and total variation distance between the true and predicted action selection policies for each video category $\agent$. \footnote{Both chi-squared and total variation distance are normalized by definition since they take values in the interval $[0,1]$.} Figure~\ref{fig:youtube_results} shows the IRL prediction results. We observed that for 7 out of the 9 video categories considered for IRL prediction analysis, the chi-squared and total variation distance for both sets of estimated action selection policies lie under $0.3$. Hence, for 7 out of 9 video categories, our IRL algorithm successfully predicts the action selection policies in the test dataset with high accuracy. Another observation from Fig.\,\ref{fig:youtube_results} is that the max-margin IRL estimate is a more accurate predictor compared to the entropy-regularized IRL estimate and outperforms the entropy-regularized IRL in 2 out of 9 video categories.}

\blue{
{\em Summary:} We illustrated the predictive performance of our IRL algorithms~\eqref{eqn:maxmargin-irl},~\eqref{eqn:entropy-irl} on a real-world YouTube dataset. We chose two point-valued IRL estimates of stopping costs from the set of feasible costs that pass the NIAS~\eqref{eqn:NIAS_thrm_CD} and NIAC~\eqref{eqn:NIAC_thrm_CD} inequalities of Theorem~\ref{thrm:NIAS_NIAC}, namely, max-margin IRL~\eqref{eqn:maxmargin-irl} and entropy-regularized IRL~\eqref{eqn:entropy-irl}. We observed that both  these cost estimates accurately predict YouTube commenting behavior (in terms of chi-squared and total variation distance as displayed in Fig.\,\ref{fig:youtube_results}). Moreover, the max-margin IRL estimate yields a more accurate prediction compared to the entropy-regularized estimate.}

\section{Finite sample performance analysis of IRL decision test}\label{sec:finitesample}
Thus far, our IRL framework 
assumes \ref{asmp:IRL1}, namely, that the inverse learner has access to infinite trials of the stopping agent in $\numagents$ environments in order to solve the convex feasibility problem in Theorem~\ref{thrm:NIAS_NIAC}. Suppose the inverse learner  records only a finite number of trials and constructs its IRL dataset~(\ref{eqn:IRL_tuple_compact}) comprising the agent's prior and empirically computed action selection policies in $\numagents$ environments.
In this section, we address the following question: {\em How robust is the IRL decision test in Theorem~\ref{thrm:NIAS_NIAC} to finite sample datasets?} We now view Theorem~\ref{thrm:NIAS_NIAC} as a detector that takes in as input a noisy (empirical) dataset and outputs whether or not the observed agent is identified as an optimal stopping agent. 
Our aim is to provide bounds on the \blue{IRL detector's} error probability in terms of the number of trials recorded by the inverse learner. We then obtain finite sample IRL results for the  examples of inverse SHT and inverse search.


\subsection{Finite sample statistical test for IRL}
Suppose the inverse learner observes the actions of a Bayesian stopping agent in  $\numagents$ environments. In addition to assumption \ref{asmp:IRL2}, we assume the following about the inverse learner for our finite sample result stated in Theorem~\ref{thrm:finite_robustness} below.
\begin{enumerate}[label=(F\arabic*)]
\setcounter{enumi}{0}
\item \label{asmp:IRL_finite_1} The inverse learner knows the {\em finite  dataset}
\begin{equation}\label{eqn:dataset_finitesample}
\begin{split}
    \datafin{\trialset} & = \{\prior,\{\actselectagentemp{\agent},\agent\in\agentset\}\},\text{ where }
    \trialset = \{\numtrials_{\state,\agent},\agent\in\agentset,\state\in\stateset\}.
\end{split}
\end{equation}
In \eqref{eqn:dataset_finitesample}, $\trialset=\{\numtrials_{\state,\agent},\agent\in\agentset,\state\in\stateset\}$, $\numtrials_{\state,\agent}$ is  the number of trials recorded by the inverse learner for environment $\agent$ and state $\state$. $\actselectagentemp{\agent}$ is the empirical action selection policy of the agent in environment $\agent$ computed  for $\numtrials_{\state,\agent}$ trials via (\ref{eqn:actionselection}).
\item \label{asmp:IRL_finite_3} The finite dataset $\datafin{\trialset}$ satisfies the following inequality.
\begin{align}\hspace{-0.5cm}\epsone{\datafin{\trialset}},\epstwo{\datafin{\trialset}}&\geq \left(\sum_{\state,\agent}\frac{\numactions}{2\numtrials_{\state,\agent}}\right) \left( \ln(2\numtrials_{\state,\agent}/\numactions)-\min_{\state,\agent}~ \ln\left(2\numtrials_{\state,\agent}/\numactions\right) \right)\label{eqn:reg_stopping}
\end{align}
\blue{In \eqref{eqn:reg_stopping}, $\numtrials_{\agent}=\sum_{\state}\numtrials_{\state,\agent}$, $\bar{\numtrials}=\numtrials/\funcstop_{\max}^2$ and $\widetilde{\numtrials}=\numtrials^{-1}$. Eq.\,\ref{eqn:reg_stopping} imposes a lower bound on the number of samples needed for our sample complexity result of inverse optimal stopping. \blue{Eq.\,\ref{eqn:reg_stopping} is a sufficient condition for obtaining the constants of the sample complexity bound as the solution of a convex optimization problem; see \eqref{eqn:opt_finite_sht_1} in the Appendix for more details.} Variables $\epsone{\cdot},\epstwo{\cdot}$ are the minimum perturbations needed for the finite dataset $\datafin{\trialset}$ to satisfy and not satisfy, respectively, the NIAS and NIAC inequalities in Theorem~\ref{thrm:NIAS_NIAC}, and defined formally in (\ref{eqn:eps1}), (\ref{eqn:eps2}) for readability.}
\setcounter{assumindex}{\value{enumi}}
\end{enumerate}

For the reader's convenience, we discuss the assumptions \ref{asmp:IRL_finite_1} and \ref{asmp:IRL_finite_3} after the finite sample complexity result, Theorem~\ref{thrm:finite_robustness}.
The feasibility test of Theorem~\ref{thrm:NIAS_NIAC} given a finite number of trials $\trialset$ can be equivalently formulated as a statistical hypothesis detection test that takes as input the finite dataset $\datafin{\trialset}$ and accepts one of the two hypotheses, $H_0$ or $H_1$.
 \begin{compactitem}
 	\item $H_0$: {\em Null hypothesis} that the observed stopping agent is identified as an optimal agent, i.e.\,, the true dataset $\datainf$ is feasible wrt the NIAS and NIAC inequalities (\ref{eqn:NIAS_thrm_CD}), (\ref{eqn:NIAC_thrm_CD}) in Theorem~\ref{thrm:NIAS_NIAC}.
 	\item  $H_1$: {\em Alternative hypothesis} that the observed stopping agent is {\em not} optimal.
\end{compactitem}
\begin{definition}[IRL detector for inverse optimal stopping]
\label{def:IRLstattest}
    Consider the inverse learner with dataset $\datafin{\trialset}$. Assume \ref{asmp:IRL2} and \ref{asmp:IRL_finite_1} hold. The IRL decision test $\testfinite(\cdot)$ for inverse optimal stopping is given by: 
    \begin{equation}
    \testfinite(\datafin{\trialset})=\begin{cases}
    H_0,& \text{if } \IRLoutput(\datafin{\trialset})\neq \emptyset\\
    H_1, & \text{if } \IRLoutput(\datafin{\trialset}) = \emptyset.
    \end{cases} \label{eqn:stattest}
\end{equation}
Here, $\IRLoutput(\datageneric)$ is the set of feasible solutions to the convex NIAS and NIAC inequalities (\ref{eqn:NIAS_thrm_CD}), (\ref{eqn:NIAC_thrm_CD}) given dataset $\datageneric$.
\end{definition}
The statistical test defined above is a  detector that accepts the null hypothesis $H_0$ if the finite dataset passes the feasibility test of Theorem~\ref{thrm:NIAS_NIAC} and accepts the alternative hypothesis $H_1$ it otherwise. Our main result stated below characterizes the performance of the feasibility test in identifying optimality given finite sample constraints, \blue{namely, provide bounds on the detector's Type-I/II error probabilities}.

\subsection{Main result. Finite sample analysis for IRL}\label{sec:finite_sample_main_result_stopping}
Our main result below (Theorem~\ref{thrm:finite_robustness}) 
 characterizes the following error probabilities of the statistical test in Definition~\ref{def:IRLstattest}:
\begin{equation}\label{eqn:post_type_12}
    \hspace{-0.4cm}\text{Type-I error prob.\,: }   \prob(H_0|\IRLoutput(\datafin{\trialset}=\emptyset),~\text{Type-II error prob.\,: } \prob(H_1|\IRLoutput(\datafin{\trialset}\neq\emptyset)
\end{equation}
In \eqref{eqn:post_type_12}, $\datafin{\numtrials})= \emptyset$ means that the finite dataset fails the convex feasibility test for NIAC and NIAS inequalities (\ref{eqn:NIAS_thrm_CD}), (\ref{eqn:NIAC_thrm_CD}) and so the agent is identified as not an optimal agent. Our finite sample result in Theorem~\ref{thrm:finite_robustness} below uses the dataset statistics variables $\epsone{\cdot},\epstwo{\cdot},g(\cdot)$ from the finite dataset $\datafin{\trialset}$ and are defined below. The quantities $\epsone{\cdot}$ and $\epstwo{\cdot}$ are the minimum perturbations needed for the finite dataset to satisfy and not satisfy, respectively, the NIAS and NIAC inequalities in Theorem~\ref{thrm:NIAS_NIAC}, and variable $g$ is the constant for the error probability bounds. 

\subsubsection*{Notation}
Theorem~\ref{thrm:finite_robustness} below uses the following variables:
\begin{align}
&\epsone{\datafin{\trialset}}=\min_{\{\actselectagentempcandsymb{\agent},\agent\in\agentset\}} \sum_{\agent}\|\actselectempsymb{\agent} - \actselectagentempcandsymb{\agent}\|_2^2 \text{ such that } \IRLoutput(\{\pi_0,\{\actselectagentempcand{\agent}\}\}) \neq \emptyset. \label{eqn:eps1}\\
&\epstwo{\datafin{\trialset}}=\min_{\{\actselectagentempcandsymb{\agent},\agent\in\agentset\}} \sum_{\agent}\|\actselectempsymb{\agent} - \actselectagentempcandsymb{\agent}\|_2^2 \text{ such that } \IRLoutput(\{\pi_0,\{\actselectagentempcand{\agent}\}\})= \emptyset.\label{eqn:eps2}\\ 
&g(\datafin{\trialset}) = \left(\numactions\sum_{\state,\agent}\widetilde{\numtrials}_{\state,\agent}\right)\prod_{\state.\agent}\left( \frac{2\numtrials_{\state,\agent}}{\numactions}\right)^{\frac{\widetilde{\numtrials}_{\state,\agent}}{\sum_{\state,\agent}\widetilde{\numtrials}_{\state,\agent}}}, \text{ where } \widetilde{\numtrials}_{\state,\agent} = \numtrials^{-1}_{\state,\agent}\label{eqn:dataset_const}
\end{align}
Having defined our notation for error probability bounds, let us now state our first sample complexity result for the IRL detector~\eqref{eqn:stattest}.
\begin{theorem}[Sample complexity for IRL detector]
\label{thrm:finite_robustness}
        Consider an inverse learner with finite dataset $\datafin{\trialset}$ (\ref{eqn:dataset_finitesample}). The inverse learner aims to detect optimality of the stopping agent's actions using the statistical test in Definition~\ref{def:IRLstattest}. Assume \ref{asmp:IRL2}, \ref{asmp:IRL_finite_1} and \ref{asmp:IRL_finite_3} hold. Then,
        the Type-I and Type-II error probabilities \eqref{eqn:post_type_12} \blue{ of the IRL detector (Definition~\ref{def:IRLstattest}) are bounded as:}
        \begin{align}
            \text{Type-I error probability\,} & \leq g(\datafin{\trialset})~\exp\left(- \trialset_{H}\cdot\epsone{\datafin{\trialset}} \right),\label{eqn:finite_result_1}\\
            \text{Type-II error probability\,} & \leq g(\datafin{\trialset})~\exp\left(-\trialset_{H}\cdot\epstwo{\datafin{\trialset}}\right)\label{eqn:finite_result_2}.
        \end{align}
         \blue{In \eqref{eqn:dataset_const}, $\trialset_{H} = \left(\sum_{\state,\agent} \numtrials_{\state,\agent}^{-1}\right)^{-1}$}, and variables $\epsone{\cdot},\epstwo{\cdot}\text{ and }g$ are defined in \eqref{eqn:eps1},~\eqref{eqn:eps2} and \eqref{eqn:dataset_const}, respectively.\\
\end{theorem}
The proof of Theorem~\ref{thrm:finite_SHT} is in Appendix~\ref{appdx:proof_finite_bound_SHT}. Below, we provide a sketch of the proof. Theorem~\ref{thrm:finite_robustness} characterizes the robustness of the IRL detector in Definition~\ref{def:IRLstattest} to finite sample constraints. It provides an upper bound on the detector's error probabilities in terms of the number of trials recorded by the inverse learner. \blue{Observe that since $\trialset_{H}$ is simply the unnormalized harmonic mean of $\trialset$~\eqref{eqn:dataset_finite_search}, the error rate is exponential in the {\em harmonic mean} of the number of trials recorded over $\numagents$ environments and $\numstates$ states.}

\blue{
The proof of Theorem~\ref{thrm:finite_robustness} uses the two-sided Dvoretzky-Kiefer-Wolfowitz (DKW) concentration inequality~\citep{VAN00,KS07} as the fundamental result to show that these error probabilities can be tightly bound in terms of the sample size $\trialset$ of the finite dataset $\datafin{\trialset}$. The DKW inequality provides a probabilistic bound on the deviation of the empirical cdf from the true cdf for i.i.d\ random variables. The i.i.d\, assumption holds for our detector in Definition~\ref{def:IRLstattest} since the observed actions of the agent for a fixed state are independent and identically distributed over trials for all environments $\agentset$. To obtain our Type-I/II error bounds, we use the Dvoretzky-Kiefer-Wolfowitz (DKW) inequality to probabilistically bound $\norm{\actselect - \actselectemp}_2$, the $L_2$-error between the empirical and true action selection policy for each environment $\agent\in\agentset$ and state $\state\in\stateset$,
followed by the union bound to bound the sum of $L_2$-errors due to finite sample size over all states and environments.}

\begin{comment}
\blue{More formally, suppose $\IRLoutput(\datafin{\trialset})\neq\emptyset$. This implies the null hypothesis $H_0$ holds if the event $E$ is true, where the event $E$ is defined as
\begin{equation*}
    \{\{\actselectagent{\agent},\agent\in\agentset\} |~ \sum_{\agent}\|p_{\agent} - \actselectempsymb{\agent}\|_2^2\leq \epstwo{\datafin{\trialset}}\},
\end{equation*}
In the above definition, $\{\actselectagent{\agent},\agent\in\agentset\}$ are the action selection policies of the true asymptotic dataset $\datainf$ (\ref{eqn:IRL_tuple_compact}). In other words, if $\datainf$ lies within an $\epstwo{\datafin{\trialset}}$ distance of the finite dataset $\datafin{\trialset}$, then the observed stopping agent's actions satisfy the NIAS and NIAC inequalities of Theorem~\ref{thrm:NIAS_NIAC} and the agent is identified as an optimal stopping agent.  Thus, $\prob(E)$ upper bounds the probability that the null hypothesis holds given the finite dataset passes the convex feasibility test of Theorem~\ref{thrm:NIAS_NIAC}. Consequently, $1-\prob(E)$ lower bounds the posterior Type-II probability of error. Using a similar argument for $\epsone{\datafin{\trialset}}$, we can upper bound the posterior Type-I error probability.}\vspace{0.2cm}\\
\end{comment}
\noindent {\em Discussion of Assumptions.}\\
\ref{asmp:IRL_finite_1}: Given the finite dataset $\datafin{\trialset}$ in (\ref{eqn:dataset_finitesample}), \ref{asmp:IRL_finite_1} says that the inverse learner checks if the convex feasibility test of Theorem~\ref{thrm:NIAS_NIAC} has a feasible solution to detect an optimal stopping agent.\\ 
\ref{asmp:IRL_finite_3}: Abstractly, \ref{asmp:IRL_finite_3} says that the inverse learner observes sufficiently many trials of the agent over all environments $\agentset$ such that the condition (\ref{eqn:reg_stopping}) is met.\vspace{0.1cm}\\ First, for a given dataset $\datageneric$, note that only one out of $\epsone{\datageneric},\epstwo{\datageneric}$ is non-zero and positive. Hence, (\ref{eqn:reg_stopping}) involves only the non-zero variable out of $\epsone{\datafin{\trialset}},\epstwo{\datafin{\trialset}}$. Some words about the RHS of (\ref{eqn:reg_stopping}).  $q(\trialset)-j(\trialset)$ is a measure of how far  is $\numtrials_{\min}=\min_{\state,\agent}\numtrials_{\state,\agent}$  from the remaining elements in $\trialset\symbol{92}\numtrials_{\min}$. Since the RHS terms are of the form $\ln(z)/z$, it is easy to check that $q(\trialset)-j(\trialset)$ decreases as the elements of $\trialset$ increase uniformly. As the number of samples go to infinity, the RHS in (\ref{eqn:reg_stopping}) tends to $0$, hence the condition is almost surely satisfied for infinite samples. For finite $\trialset$, checking if (\ref{eqn:reg_stopping}) holds requires the inverse learner to solve an optimization problem (for the LHS) and perform $\numagents\numstates$ multiplication operations and $\numagents\numstates$ addition operations to compute the RHS of (\ref{eqn:reg_stopping}). \blue{As a practical estimate, for the inverse SHT task in Sec.\,\ref{sec:example_regul_IRL_SHT} for $100$ SHT environments, we observed that the inequality in \eqref{eqn:reg_stopping} is satisfied if the samples exceeded $\sim 10^3$ for each environment.
}

\begin{comment} 
{\em Structure of Proof:}
There are three key steps in the proof of Theorem~\ref{thrm:finite_robustness}. We will discuss this in the context of the posterior Type-I probability of error. For the first step, notice that the IRL detector produces the same output for both the finite and asymptotic dataset if the true action selection policies lie anywhere in the $\epsone{\datafin{\trialset}}$ neighborhood of the finite dataset (\ref{eqn:dataset_finitesample}). The posterior probability $\prob(H_0|\IRLoutput(\datafin{\trialset})\neq \emptyset)$ can be lower bounded by the probability that the true dataset belongs in this region which can be further bounded by using the DKW concentration inequality. 

The second step involves making this lower bound  tight, i.e.\,, maximizing this bound subject to the constraint that the maximum deviation between the true and finite dataset is less than the minimum perturbation $\epsone{\datafin{\trialset}}$. This results in (\ref{eqn:finite_result_1}). A similar procedure applies to the Type-II error. 

Finally, we invoke \ref{asmp:IRL_finite_2} (assumption on the learner's prior belief about the optimality of the agent) to express the Type-I and Type-II error probability in terms of the posterior error bounds computed above. By using the implicit knowledge that these probabilities are less than $1$, it is straightforward to show that (\ref{eqn:finite_result_3}), (\ref{eqn:finite_result_4}) hold. 
\end{comment}

\subsection{Example 1. Finite sample effects for IRL in inverse SHT}
\label{sec:finite_SHT}
We next turn to a finite sample analysis of IRL for inverse sequential hypothesis testing (SHT).
Recall from Theorem~\ref{thrm:classic_SHT} that
identifying optimality of SHT is equivalent to feasibility of the linear inequalities NIAS and NIAC$^{\ast}$.
The inverse learner's SHT dataset comprises both the agent action selection policies and the expected stopping times to perform IRL compared to only the action selection policies for inverse optimal stopping. Hence, in addition to the DKW inequality, our main result, Theorem~\ref{thrm:finite_SHT} also uses the Hoeffding's inequality~\citep{BLM13} to account for the finite sample effect\footnote{Hoeffding's inequality applies to bounded r.v.s\,, and is true for SHT since the stopping time $\funcstop$ is finite almost surely.} on the computation of the expected stopping time.

{\em Assumptions and Detection Test.} Suppose the inverse learner observes the actions of the Bayesian stopping agent over $\numagents$ SHT environments. We assume the following about the inverse learner for our finite sample result stated below for the inverse SHT problem.
\begin{enumerate}[label=(F\arabic*)]
\setcounter{enumi}{\value{assumindex}}
\item \label{asmp:IRL_finite_SHT_1} The inverse learner uses the {\em finite SHT dataset}
\begin{equation}
\label{eqn:dataset_finite_SHT}
    \datafin{\trialset} = \{\prior,\{\actselectagentemp{\agent},\hat{\sumruncostsymbol}_{\agent},\agent\in\agentset\}\}
\end{equation}
to detect if the stopping agent is an optimal SHT agent or not. The variable $\trialset$ defined in (\ref{eqn:dataset_finitesample}) is the number of trials recorded by the inverse learner, $\hat{\sumruncostsymbol}_{\agent}$ is the sample average of the agent's stopping time in the $\agent^{\text{th}}$ environment. $\actselectagentemp{\agent}$ is the agent's empirical action selection policy computed for $\numtrials_{\state,\agent}$ trials via (\ref{eqn:actionselection}) in the $\agent^{\text{th}}$ environment.
\item \label{asmp:IRL_finite_SHT_2}The inverse learner knows $\tau_{\max} = \inf~\{\timeinst>0~|~\prob(\funcstop\leq t)=1,\forall \agent\in\agentset\}$, an upper bound on the stopping time of the SHT strategies chosen by the agent in all environments $\agentset$.  
\item \label{asmp:IRL_finite_SHT_3}The finite dataset $\datafin{\trialset}$ satisfies the following inequality.
\begin{equation}\label{eqn:reg_SHT}
\begin{split}
&\epsone{\datafin{\trialset}},\epstwo{\datafin{\trialset}} \geq q(\trialset)-j(\trialset),\text{ where}\\
    &q(\trialset) = \sum_{\state,\agent}\frac{\ln(2\numtrials_{\state,\agent}/\numactions)}{2\numtrials_{\state,\agent}/\numactions} + \sum_{\agent} \frac{\ln(2\bar{\numtrials}_{\agent})}{2\bar{\numtrials}_{\agent}},\\
    &j(\trialset)=\min_\agent \left(\min_{\state} \operatorname{ln}\left(\frac{2\numtrials_{\state,\agent}}{\numactions}\right), \operatorname{ln}\left(\frac{\bar{\numtrials}_{\agent}}{\funcstop^2_{\max}}\right) \right) \left( \numactions\sum_{\state,\agent}\frac{\numtrials^{-1}_{\state,\agent}}{2}+\sum_{\agent}\frac{\numtrials^{-1}_{\agent}}{2}\right)
    \end{split}
\end{equation}
\setcounter{assumindex}{\value{enumi}}
\end{enumerate}
In \eqref{eqn:reg_SHT}, $\numtrials_{\agent}=\sum_{\state}\numtrials_{\state,\agent}$, $\bar{\numtrials}=\numtrials/\funcstop_{\max}^2$ and $\widetilde{\numtrials}=\numtrials^{-1}$.
\blue{Analogous to \eqref{eqn:reg_stopping} in assumption \ref{asmp:IRL_finite_3} for finite sample complexity of IRL for optimal stopping, \eqref{eqn:reg_SHT} imposes a lower bound on the number of samples needed for our sample complexity result of inverse SHT. \blue{Eq.\,\ref{eqn:reg_SHT} is a sufficient condition for obtaining the constants of the sample complexity bound as the solution of a convex optimization problem.} $\epsone{\cdot},\epstwo{\cdot}$ are the minimum perturbations needed for the finite dataset to satisfy and not satisfy, respectively, the linear NIAS and NIAC$^\ast$ inequalities in Theorem~\ref{thrm:classic_SHT}, and defined formally in \ref{eqn:eps_SHT}). The quantities $q(\cdot),j(\cdot)$ are decreasing functions of the sample size~$\trialset$.} For the reader's convenience, we discuss the assumptions \ref{asmp:IRL_finite_SHT_1}-\ref{asmp:IRL_finite_SHT_3} after the finite sample complexity result, Theorem~\ref{thrm:finite_SHT}. Analogous to Definition~\ref{def:IRLstattest}, the statistical detection test for the inverse SHT problem is defined below. It takes in as input a finite (noisy) dataset and outputs one of the two hypotheses- $H_0$ (agent is an optimal SHT agent) or $H_1$ (agent is not an optimal SHT agent).

\begin{definition}[IRL decision test for inverse SHT]
\label{def:IRLstattest_SHT}
    Consider the inverse learner with dataset $\datafin{\trialset}$ (\ref{eqn:dataset_finite_SHT}). Assume \ref{asmp:IRL2} \ref{asmp:SHT2}, \ref{asmp:SHT3} and \ref{asmp:IRL_finite_SHT_1} hold. The IRL detector $\testfinite(\cdot)$ for the inverse SHT problem is given by: 
    \begin{equation}\label{eqn:IRL_detector_SHT}
    \testfinite(\datafin{\trialset})=\begin{cases}
    H_0,& \text{if } \IRLoutput_{\SHT}(\datafin{\trialset})\neq \emptyset\\
    H_1, & \text{if } \IRLoutput_{\SHT}(\datafin{\trialset}) = \emptyset.
    \end{cases}
\end{equation}
Here, $\IRLoutput_{\SHT}(\datageneric)$ is the set of feasible solutions to the linear NIAS and NIAC$^{\ast}$ inequalities (\ref{eqn:NIAS_def}), (\ref{eqn:NIAC_ast_def}) in Theorem~\ref{thrm:classic_SHT} given dataset $\datageneric$.
\end{definition}

 {\em Main Result. Finite Sample analysis for inverse SHT}\\
We now present our finite sample result for IRL of the inverse SHT problem. It provides bounds for the Type-I/II error probabilities of the IRL detector~\eqref{eqn:IRL_detector_SHT} in terms of the sample size of $\datafin{\trialset}$~\eqref{eqn:dataset_finite_SHT}.

{\em Notation.} Theorem~\ref{thrm:finite_SHT} below uses the following variables:
\begin{equation}
\begin{split}
   &\epsone{\datafin{\trialset}}:\min_{\{\actselectagentempcandsymb{\agent},\hat{\sumruncostsymbol}_\agent'\}} \sum_{\agent}\|\actselectempsymb{\agent} - \actselectagentempcandsymb{\agent}\|_2^2 +  (\hat{\sumruncostsymbol}_{\agent} - \hat{\sumruncostsymbol}_\agent') ^ 2,~\IRLoutput_{\SHT}(\{\prior,\{\actselectagentempcandsymb{\agent},\sumruncostagent{\agent}'\}\}) \neq \emptyset\\
    &\epstwo{\datafin{\trialset}}:\min_{\{\actselectagentempcandsymb{\agent},\hat{\sumruncostsymbol}_\agent'\}} \sum_{\agent}\|\actselectempsymb{\agent} - \actselectagentempcandsymb{\agent}\|_2^2 + (\hat{\sumruncostsymbol}_{\agent} - \hat{\sumruncostsymbol}_\agent') ^ 2,~\IRLoutput_{\SHT}(\{\prior,\{\actselectagentempcandsymb{\agent},\sumruncostagent{\agent}'\}\})= \emptyset\\
    &i(\datafin{\trialset}) = \trialset_{H}(\SHT)\cdot h(\datafin{\trialset}),~\text{where }\trialset_{H}(\SHT)=\numactions\sum_{\state,\agent}\numtrials_{\state,\agent}^{-1} + \tau_{\max}^2\sum_{\agent}~\numtrials_{\agent}^{-1}\text{, and}\\
    &~h(\datafin{\trialset}) = \prod_{\agent}\bigg( \left(2\bar{\numtrials}_{\agent}\right)^{\bar{\numtrials}_{\agent}^{-1}}\prod_{\state}\left(\frac{2\bar{\numtrials}_{\state,\agent}}{\numactions}\right)^{\numactions\widetilde{\numtrials}_{\state,\agent}}\bigg)^{\trialset_{H}^{-1}(\SHT)}.
\end{split}
\label{eqn:eps_SHT}
\end{equation} 
In \eqref{eqn:eps_SHT}, $\numtrials_{\agent}=\sum_{\state}\numtrials_{\state,\agent}$, $\bar{\numtrials}=\numtrials/\funcstop_{\max}^2$ and $\widetilde{\numtrials}=\numtrials^{-1}$. Analogous to the finite sample result for inverse optimal stopping, $\epsone{\cdot},~\epstwo{\cdot}$ defined above are the minimum perturbations needed for the finite SHT dataset to  satisfy and not satisfy, respectively, the NIAS and NIAC$^\ast$ inequalities in Theorem~\ref{thrm:classic_SHT}. Compared to the minimum perturbations defined in \eqref{eqn:eps1} and \eqref{eqn:eps2} for inverse optimal stopping, the key distinction is that $\epsone{\cdot}$ and $\epstwo{\cdot}$ in  \eqref{eqn:eps_SHT} also involve  perturbations in the expected continue cost of the agent. \blue{The variable $i$ in \eqref{eqn:eps_SHT} is the error constant for the finite sample error bounds for inverse SHT; variable $\trialset_{H}(\SHT)$ can be interpreted as a weighted harmonic mean of the recorded trials $\trialset$~\eqref{eqn:dataset_finite_SHT}.}
\begin{theorem}[Sample complexity for inverse SHT]
\label{thrm:finite_SHT}
        Consider an inverse learner with dataset $\datafin{\trialset}$ (\ref{eqn:dataset_finitesample})
        detecting if the agent acting in multiple environments $\agentset$ is an optimal SHT agent using the statistical test in Definition~\ref{def:IRLstattest_SHT}. Assume 
        \ref{asmp:IRL_finite_SHT_1}-\ref{asmp:IRL_finite_SHT_3} hold. Then,
        the Type-I  and Type-II  error probabilities of the IRL detector (Definition~\ref{def:IRLstattest_SHT})
        are bounded as:
        \begin{align}
            & \text{Type-I error probability} \leq  i(\datafin{\trialset})\, \exp\left(-2~\trialset_{H}(\SHT)\cdot\epsone{\datafin{\trialset}}\right) ,\label{eqn:finite_result_SHT_1}\\
            &  \text{Type-II error probability} \leq i(\datafin{\trialset})\, \exp\left(-2~\trialset_{H}(\SHT)\cdot\epstwo{\datafin{\trialset}}\right) \label{eqn:finite_result_SHT_2}.
        \end{align}
\end{theorem}
The proof of Theorem~\ref{thrm:finite_SHT} is in Appendix~\ref{appdx:proof_finite_bound_SHT}. Theorem~\ref{thrm:finite_SHT} characterizes the robustness of the linear feasibility test in Theorem~\ref{thrm:classic_SHT} to finite sample constraints. \blue{Compared to Theorem~\ref{thrm:finite_robustness}, the finite sample result of Theorem~\ref{thrm:finite_Search} requires accounting for the empirical estimate of the agent's expected continue cost. Hence, in addition to the DKW inequality, the proof of Theorem~\ref{thrm:finite_SHT} uses Hoeffding's inequality to bound the empirical estimation error for the expected continue cost.}

{\em Discussion of Assumptions.} \\
\ref{asmp:IRL_finite_SHT_1}: \ref{asmp:IRL_finite_SHT_1} specifies the inverse learner's dataset for the inverse SHT problem computed from a finite number of trials.\\ \ref{asmp:IRL_finite_SHT_2} says the inverse learner knows the upper bound of the agent's stopping times over all environments. This assumption is crucial for our main result since the Hoeffding's inequality (for bounding the finite sample effect of the expected stopping time) requires this knowledge. \\
\ref{asmp:IRL_finite_SHT_3}: The condition (\ref{eqn:reg_SHT}) in \ref{asmp:IRL_finite_SHT_3} is  analogous to assumption \ref{asmp:IRL_finite_3} for the finite sample result for IRL of optimal stopping. \ref{asmp:IRL_finite_SHT_3} admits a close form expression for the error bounds in Theorem~\ref{thrm:finite_SHT}. Abstractly, \ref{asmp:IRL_finite_SHT_3} says that the number of samples recorded by the inverse learner is sufficiently large so that the condition (\ref{eqn:reg_SHT}) is satisfied.

\subsection{Example 2. Finite sample effects for IRL in inverse search} \label{sec:finite_Search}
We now analyze the finite sample effect of IRL for inverse search. Recall from Theorem~\ref{thrm:Search} that optimal search is equivalent to feasibility of the linear NIAC$^{\dag}$ inequalities. Our main result below, namely, Theorem~\ref{thrm:finite_Search}, characterizes the robustness of the feasibility test (wrt the NIAC$^{\dag}$ inequality) for detecting optimal search under finite sample constraints. It turns out that  Theorem~\ref{thrm:finite_Search} is a special case of Theorem~\ref{thrm:finite_robustness}, our finite sample complexity result for IRL of inverse optimal stopping.   

{\em Main assumptions and detection test.}
Suppose the inverse learner observes the actions of a Bayesian stopping agent. We assume the following about the inverse learner:
\begin{enumerate}[label=(F\arabic*)]
\setcounter{enumi}{\value{assumindex}}
    \item \label{asmp:IRL_finite_Search} Instead of \ref{asmp:IRL_search_1}, the inverse learner uses the {\em finite  dataset}
    \begin{equation}
        \label{eqn:dataset_finite_search}
        \datafinsearch{\trialset} = \{\prior,\{\occagentemp{\agent},\agent\in\agentset\}\}
    \end{equation}
    to detect if the agent performs optimal search or not. $\occagentemp{\agent}$ is the empirical search action policy of the agent defined in (\ref{eqn:act_search_policy}) and $\trialset=\{\numtrials_{\state,\agent},\state\in\stateset,\agent\in\agentset\}$ denotes the number of trials recorded by the inverse learner in state $\state$, where $\agent$ indexes the search environment.
    \item \label{asmp:IRL_finite_Search_2}The prior belief of the targets $\prior$ is a uniform prior, i.e.\,, $\prior(\state)=1/\numstates$. Also, the reveal probability $\revprob$ is the same for all actions $\action\in\actionset$, i.e.\,, $\revprob= \revprobsymb$. Although the variable $\revprobsymb$ is unknown to the inverse learner, it satisfies the following inequality. 
    \begin{equation}\label{eqn:revprob_constraint}
        \revprobsymb \geq \max\left\{\revprobprime,1 - \frac{\min_{\action\in\actionset} \searchcostagent{\agent}}{\max_{\action\in\actionset} \searchcostagent{\agent}},~\forall \agent\in\agentset \right\}
    \end{equation}
\setcounter{assumindex}{\value{enumi}}
\end{enumerate}

For the reader's convenience, we discuss the assumptions \ref{asmp:IRL_finite_Search} and \ref{asmp:IRL_finite_Search_2} after the finite sample complexity result, Theorem~\ref{thrm:finite_Search}. We now define the statistical detection test for the inverse search problem. It takes in as input the finite (noisy) dataset $\datafinsearch{\trialset}$ (\ref{eqn:dataset_finite_search}) and detects one of the two hypotheses- $H_0$ (agent performs optimal search) or $H_1$ (agent does {\em not} perform optimal search).
\begin{definition}[IRL decision test for inverse search]
\label{def:IRLstattest_Search}
    Consider the inverse learner with dataset $\datafinsearch{\trialset}$ (\ref{eqn:dataset_finite_search}). Assume \ref{asmp:IRL_search_2}, \ref{asmp:IRL_finite_Search} holds. The IRL detector $\testfinite(\cdot)$ for the inverse search problem is given by: 
    \begin{equation}
    \testfinite(\datafinsearch{\trialset})=\begin{cases}
    H_0,& \text{if } \IRLoutput_{\Search}(\datafinsearch{\trialset})\neq \emptyset\\
    H_1, & \text{if } \IRLoutput_{\Search}(\datafinsearch{\trialset}) = \emptyset.
    \end{cases}
\end{equation}
Here, $\IRLoutput_{\Search}(\datageneric)$ is the set of feasible solutions to the linear NIAC$^{\dagger}$ inequalities (\ref{eqn:NIAC_dag_def}) in Theorem~\ref{thrm:Search} given dataset $\datageneric$.
\end{definition}

{\em Main Result. Finite Sample Result for Inverse Search}.\\
We now present Theorem~\ref{thrm:finite_Search}, our finite sample result for IRL of the inverse search problem. 
Theorem~\ref{thrm:finite_Search} provides bounds for the Type-I/II and posterior Type-I/II error probabilities of the IRL detector in Definition~\ref{def:IRLstattest_Search} in terms of the sample size of the finite search dataset and uses the following variables.
\begin{align*}
    \epsone{\datafinsearch{\trialset}} &= \min_{\{\occagentcandsymb{\agent},\agent\in\agentset\}} \sum_{\agent} \|\occagentempsymb{\agent}-\occagentcandsymb{\agent}\|_2^2,~ \IRLoutput_{\Search}(\{\prior,\occagentcand{\agent},\agent\in\agentset\}) \neq \emptyset.\\
    \epstwo{\datafinsearch{\trialset}} &= \min_{\{\occagentcandsymb{\agent},\agent\in\agentset\}} \sum_{\agent} \|\occagentempsymb{\agent}-\occagentcandsymb{\agent}\|_2^2,~ \IRLoutput_{\Search}(\{\prior,\occagentcand{\agent},\agent\in\agentset\})= \emptyset\nonumber.
\end{align*}
The variables $\epsone{\cdot},~\epstwo{\cdot}$ are the minimum perturbations needed for the finite search dataset to satisfy and not satisfy, respectively, the linear NIAC$^{\dag}$ inequalities~\eqref{eqn:NIAC_dag_def} in Theorem~\ref{thrm:Search}. 
\begin{theorem}[Sample complexity for  inverse search]
\label{thrm:finite_Search}
        Consider an inverse learner with dataset $\datafinsearch{\trialset}$ (\ref{eqn:dataset_finitesample})
        detecting if a Bayesian stopping agent is performing optimal search by using the statistical test in Definition~\ref{def:IRLstattest_Search}. Assume \ref{asmp:IRL_finite_Search} and \ref{asmp:IRL_finite_Search_2}  hold. Then, the Type-I  and Type-II  error probabilities for the IRL detector (Definition~\ref{def:IRLstattest_Search}) are bounded as:
        \begin{align}
             &\text{Type-I error probability} \leq\frac{(1-\revprobprime)\numactions}{\epsone{\datafinsearch{\trialset}}(\revprobprime)^2}\left(\sum_{\state,\agent} \numtrials_{\state,\agent}^{-1/2}\right)^2, \label{eqn:finite_result_Search_1}\\
            & \text{Type-II error probability} \leq \frac{(1-\revprobprime)\numactions}{\epstwo{\datafinsearch{\trialset}}(\revprobprime)^2}\left(\sum_{\state,\agent} \numtrials_{\state,\agent}^{-1/2}\right)^2.\label{eqn:finite_result_Search_2}
        \end{align}
\end{theorem}
The proof of Theorem~\ref{thrm:finite_Search} is in Appendix \ref{appdx:proof_Search}. 
Theorem~\ref{thrm:finite_Search} characterizes the robustness of the linear feasibility test in Theorem~\ref{thrm:Search} to finite sample constraints. It upper bounds the probability of incorrectly detecting the Bayesian agent as an optimal search agent or not an optimal search agent, in terms of the number of trials recorded by the inverse learner.

{\em Discussion of assumptions.} \\
\ref{asmp:IRL_finite_Search}: Assumption \ref{asmp:IRL_finite_Search} specifies the inverse learner's dataset for the inverse search problem computed from a finite number of trials. \\
\ref{asmp:IRL_finite_Search_2}: \ref{asmp:IRL_finite_Search_2} says that the agent has the same reveal probability for all locations for all environments and the inverse learner knows this reveal probability is greater than a certain value. \blue{This assumption can be viewed as an analogy of having the same instantaneous continue cost for the agent solving the SHT problem.}
The condition (\ref{eqn:revprob_constraint}) results in the optimal search strategy of the agent to be periodic for all environments - the agent searches each location exactly once in a particular order (depends on the agent's search costs) and repeats this cycle till the target is located. This allows the search action policy $\occagent{\agent}$ to be written in terms of the conditional pdf of the stopping time $\prob_{\stoptime_{\agent}}(\funcstop|\state)$ (see Appendix~\ref{appdx:proof_finite_bound_search}). By analyzing the finite sample effects of the stopping time due to the added structure, Theorem~\ref{thrm:finite_Search} results. Note that Theorem~\ref{thrm:finite_Search} does not require the inverse learner to have information about the true stopping time of the agent, but only the empirical search action policy.

\subsection*{Summary}
For finite sample observations,
this section presented an IRL detector for optimality of a Bayesian stopping agent (Definition~\ref{def:IRLstattest}) and provided error bounds of the detector  (Theorems~\ref{thrm:finite_robustness}). We also presented finite sample IRL detectors for optimality in SHT and Bayesian search (Definitions~\ref{def:IRLstattest_SHT}, \ref{def:IRLstattest_Search})  and obtained error bounds of the detector in terms of the sample size (Theorem~\ref{thrm:finite_SHT}, \ref{thrm:finite_Search}). \blue{The key idea behind the sample complexity results is the construction of a probabilistic bound on the minimum perturbation needed to satisfy and not satisfy, respectively, the feasibility inequalities for optimality to bound the Type-I and Type-II error probabilities of the IRL detector, respectively.}



\section{Discussion and extensions}

\blue{This paper has proposed Bayesian revealed preference methods for inverse reinforcement learning (IRL)  in partially observed environments.} Specifically, we considered  IRL for a Bayesian agent performing multi-horizon sequential stopping. 
\blue{The results in this paper achieve IRL under the following restrictions on the inverse learner: (1) The inverse learner does not know if the decision maker is an optimal Bayesian stopping agent (2) The inverse learner does not know the agent's observation likelihood and (3) IRL for noisy datasets.} \blue{Our IRL algorithms first identify if the agent is behaving in an  optimal manner, and if so, estimate their stopping costs. The inverse learner can at best identify optimality of an agent's strategy wrt to its strategies chosen in other environments, a notion intuitively explained in the introduction and defined formally in Lemma~\ref{lem:relativeoptimality}.} 
To illustrate our IRL approach, we considered two examples of sequential stopping problems, namely, sequential hypothesis testing (SHT) and Bayesian search and provided algorithms to estimate their misclassification/search costs.

Our  main results were:
\begin{compactenum}
\item Specifying necessary and sufficient conditions for the decisions taken by a Bayesian agent in multiple environments to be identified as optimal sequential stopping and generating set-valued estimates of their stop costs (Theorem~\ref{thrm:NIAS_NIAC}). \blue{To the best of our knowledge, our IRL results for Bayesian stopping time problems when the inverse learner has no knowledge of the agent's dynamics is novel.}
\item Constructing convex feasibility based IRL algorithms for set-valued estimation of misclassification for an SHT agent (Theorem~\ref{thrm:classic_SHT}) and search costs for a search agent (Theorem~\ref{thrm:Search}) when decisions from infinite trials of the agent are available in multiple SHT and search environments, respectively. These results are special cases of Theorem~\ref{thrm:NIAS_NIAC} due to additional structure of the SHT and search problem compared to generic sequential stopping problems.
\item Proposing IRL detection tests for detecting optimality of sequential stopping, SHT and search when only a finite number of agent decisions are observed. 
\item Providing sample complexity bounds on the Type-I/II and posterior Type-I/II error probabilities of the above detection tests under finite sample constraints (Theorems~\ref{thrm:finite_robustness}, \ref{thrm:finite_SHT} and \ref{thrm:finite_Search}).
\end{compactenum}


\subsubsection*{Extensions}
This paper identifies optimal stopping behavior in a Bayesian agent by observing their actions without external interference. A natural extension is to consider the controlled IRL setting where the inverse learner is \blue{an {\em active} entity} that can influence/control the actions of the agent. This leads to the  question: {\em How to influence the agent's actions so as to better identify optimal stopping behavior and estimate the agent costs more efficiently?}\\ 
Another question is: {\em How to formulate conditions under which the set-valued cost estimates for an agent in finitely many environments tends to a point-valued estimate as the number of environments tend to infinity?} In classical revealed preference theory, the papers  \cite{RN15,MSCL78} characterize properties of utility functions that rationalize infinite datasets. It is worthwhile generalizing these results to a Bayesian IRL setting.\\
\blue{Recent advances in deep IRL~\citep{deepIRL-1,deepIRL-2} use deep neural networks as function approximators for the underlying feature space. Our current research aims to extend the results in this paper to deep-IRL for inverse optimal stopping where the inverse learner does not know the underlying state space and relies on neural networks for feature space approximation.}\\
\blue{The IRL methodology of the paper assumes the analyst has no knowledge of the agent's observation likelihood. If the inverse learner knows {\em a priori} that the agent must choose its observation likelihood from a finite set, the inverse learner cannot rely on NIAS and NIAC in Theorem~\ref{thrm:NIAS_NIAC} for checking optimal Bayesian stopping. Instead, one must adapt adaptive search techniques for identifying the optimal observation likelihood that is (a) consistent with the inverse learner's dataset and (b) optimizes the agent's objective. If the agent's observation likelihood is known to be multi-variate Gaussian, then one can use the tree search approach that has seen success in applications such as adversarial tracking~\citep{LS13} and motion planning~\citep{BR11}. Extending the IRL results in this paper to tree-based adaptive search techniques is a subject of current research.
}

\blue{Finally, it is worthwhile  exploring IRL for stopping time problems using the iterative update approach of \cite{AB04} and the MCMC based sampling approach of \cite{ZB08}.}

\newpage
\appendix

\section{Context and Perspective. IRL for Bayesian stopping problems}\label{appdx:context}
\subsection{Literature and Applications. IRL in Bayesian stopping problems}
  \label{sec:intro_applications}
\blue{
IRL methods have been successfully applied to areas like robotics~\citep{KR16}, user engagement in multimedia social networks such as YouTube \citep{HKP20}, autonomous navigation by \citet{AB04,ZB08} and \cite{SH16} and inverse cognitive radar \citep{VK20,VK201}. Below we discuss several real-world examples where an analyst aggregates data from a Bayesian stopping time agent, and has no knowledge of the agent's observation likelihood for solving the IRL problem.
\begin{compactitem}
\item {\em Consumer Insights and Ad Design Research:} Online multimedia are sequential Bayesian decision makers~\citep{RC04,KRA10}; they accumulate evidence sequentially from audio-visual cues on the screen and then take an action (for example, playing a video, clicking on an ad etc.\,). In advertisement design, an analyst observes how an online user (stopping time agent) reacts to a pop-up advertisement in multiple environments, where the environment is characterized by the current web-page, content and position of the ad etc. In consumer research for online movie platforms, the analyst observes whether an online user clicks on a movie thumbnail or not in multiple scenarios, where the scenario depends on factors like user's past history of movies watched and neighboring movie thumbnails. The decision process of the online user (forward learner) in both these examples can be embedded into an SHT framework, where the sensing cost is the cost of attention to visual cues and the stopping (terminal) cost measures the online user's preferences for viewing the advertisement/movie.
By characterizing the content reactivity of online users in different multimedia platforms, IRL for stopping time problems is useful for targeted ad-design and content recommendation.
\item {\em Electronic counter-countermeasures in electronic Warfare:} Sequential Bayesian jamming models are extensively used in Electronic Counter Measures (ECM) for mitigating radar systems; see \citet{ARI15,SO16} and \citet{AR06} for details. The proposed  IRL algorithms  can be used for Electronic Counter Counter Measures (ECCM) by the radar system to reverse engineer the adversarial ECM algorithms and avoid performance mitigation, hence extending the paper \cite{VK201} to the Bayesian case. For instance, suppose an adversarial radar uses Bayesian search to identify valuable targets like in \citet{searchradar}.  Using IRL for inverse search, a radar analyst can use the estimated search costs of the adversarial radar for effectively designing the targets to avoid being easily detected. \blue{\citet{RYA19,XUE21} develop inverse optimal control (IOC) based IRL algorithms for reconstructing adversary intent for tracking control. Our work complements \citet{XUE21} since it allows one to still achieve IRL without knowledge of model dynamics, as is common to assume in the literature.}  
\item {\em Interpretable ML for Smart Healthcare:} Recently, sequential Bayesian models for assisting medical diagnoses have been aggressively used in smart healthcare like in \citet{NI18,ON13,EX13,JA12} and \citet{TH94}. These trained models are usually only accessible in an abstracted black-box format in an executable software application. {\em Our IRL algorithm provides an interpretable Bayesian decision model for these assistive algorithms.} Interpretability in AI-enabled healthcare~\citep{AHM18} facilitates informed decisions for the debugging and improvement of assistive diagnoses.
\end{compactitem}
}

\subsection{Related works in IRL}
\label{sec:related_lit}
We now summarize the key IRL works in the literature and compare them to our paper.


\blue{{\em (a) IRL in fully observed environments:}} Traditional IRL~\citep{NG00,AB04} aims to estimate an unknown deterministic reward function of an agent by observing the optimal actions of the agent in a Markov decision process (MDP) \blue{setting. The key assumption is the existence of an optimal policy. Our convex feasibility approach for IRL in stopping time problems can be viewed as a generalization of the feasibility inequalities in \citet[Theorem 3]{NG00}. \citet[Theorem 3]{NG00} compute a feasible set of rewards that ensure the agent's policy outperforms all other policies. Since the set of policies for an MDP is finite, \citet[Theorem 3]{NG00} comprises a finite set of linear inequalities. In comparison, the set of policies for a partially observed MDP (POMDP) is infinite. From the feasible set of rewards, \cite{NG00,RAT06} choose the max-margin reward, i.e.\,, the reward that maximizes the regularized sum of differences between the performance of the observed policy and all other policies. In Sec.\,\ref{sec:Example_SHT}, we compute a regularized max-margin estimate of costs for inverse SHT and plot the reconstruction error.
\cite{AB04} achieve IRL by devising iterative algorithms for estimating the agent's reward function. Abstractly, the key idea is to terminate the iterative process once the value function of the rewards converges to an $\epsilon$ interval.}
\vspace{0.1cm}\\
\cite{ZB08} use the principle of maximum entropy for achieving IRL of an optimal agent, \blue{wherein the agent's policy is subject to a Shannon mutual information regularization. This regularization facilitates expressing the optimal policy in closed form; the optimal policy turns out to be softmax in terms of the Q-function of the MDP. \citet{JSB20} extend \cite{ZB08} to a more general regularization setup that also admits a closed form solution to the optimal policy in terms of strongly convex functions for regularization, for examples, the Tsallis entropy~\citep{LKL20} that generalizes Shannon entropy.
}
\blue{Solving the IRL task with zero dynamics knowledge has also been explored in the literature. \citet{HGW16} append the IRL task with simultaneous learning of model dynamics, specifically, the agent's transition kernel. The key idea in the approach is to maximize the log-likelihood of sampled trajectories wrt the appended parameter space that parametrizes the agent's rewards and transition kernel. Our problem setting differs from \citet{HGW16} in that we operate in the non-parametric partially observed setting regime where the observation likelihood of the agent is unknown and not necessarily parametrizable. Indeed, our results can be specialized to parameter families of observation likelihood known to the analyst, and is a subject of current research.
}

\cite{LV12} generalize IRL to continuous space processes and circumvent the problem of finding the optimal policy for candidate reward functions. Recently \cite{FU17,WF15,SH16} and \citet{FN16} used deep neural networks for IRL to estimate agent rewards  parametrized by complicated non-linear functions. \cite{DR07} achieve IRL when the agent's rewards are sampled from a prior distribution and the demonstrator's trajectories update the posterior belief of the reward. \blue{Building on the seminal work of \citet{RUST94}, \citet{KIM21,CAO21} study identifiability of parameters for structure MDPs in IRL. In analogy, in this paper, we provide identifiability conditions for a subset of POMDPs, namely, Bayesian stopping problems.
}

\blue{{\em (b) IRL in partially observed environments:} 
The influential works of \citet{CH09,CH11} and \citet{MK12} are the first works on IRL in a POMDP setting.
They extend traditional IRL~\citep{NG00,AB04} to an infinite state space (space of posterior beliefs of the agent). \cite{MK12} extend Bayesian IRL~\citep{DR07} for MDPs to the POMDP setting. In analogy to Bayesian IRL, the aim is to compute the posterior distribution of reward functions given an observation dataset. The assumption of a softmax action policy suffices to compute the likelihood of the observation dataset given a reward function, and hence, bypasses the need to compare the agent's performance with respect to other candidate policies. \vspace{0.1cm}\\
Since our work is  closely related to \cite{CH09,CH11}, we briefly review their approach. In \cite{CH09,CH11}, the inverse learner first checks if the agent chooses the optimal action given a particular posterior belief, for {\em finitely many beliefs} aggregated from the observed trajectories of belief-action pairs. This is analogous to our NIAS condition~\eqref{eqn:NIAS_def} in Theorem~\ref{thrm:classic_SHT}, where we check if the agent's terminal action is optimal given its terminal belief. Next, the inverse learner check if the agent's policy is optimal with respect to a {\em finite set of policies} that deviate from the observed policy by a single step. This resembles our NIAC condition~\eqref{eqn:NIAC_ast_def} in Theorem~\ref{thrm:NIAS_NIAC} where we check for optimality of the Bayesian decision maker's actions in multiple environments. \vspace{0.1cm}\\
As \cite{CH09,CH11} mention, this approach to checking for optimality only gives rise to a necessary condition, and not a necessary and sufficient condition like in \cite{NG00,AB04}, where the number of policies are finite. In other words, without prior information about the nature of the Q-function given a policy, it is impossible to check for global optimality, that is, find a reward function that outperforms {\em all} other policies (infinitely many policies).}

\blue{Our Bayesian revealed preference based approach is {\em complementary} to \cite{CH11}. While \cite{CH11} develop IRL methods for POMDPs with no assumption on problem structure, we consider a subset of POMDPs, namely, Bayesian stopping time problems. Due to the structure of stopping time problems, we show that our IRL algorithms {\em do not} require knowledge of the observation likelihood of the decision maker, nor require solving a POMDP. Indeed, IRL for generic POMDPs is non-identifiable if the inverse learner does not know the model dynamics, nor can solve a POMDP. To test for optimality, our IRL algorithms rely on the decision maker's strategies from multiple environments, where every environment differs in the terminal cost. Decision strategy in multiple environments can be viewed as a surrogate for performance wrt different policies. To summarize, our work builds on the seminal work of \cite{CH11} with the key discerning features of our IRL methodology being}:
(1) Unobservability of agent dynamics,~(2) No assumptions on decision optimality, and (3) IRL generalization for empirical (noisy) datasets \blue{with performance guarantees via finite sample complexity}.

\blue{
{\em (c) Inverse Rational Control (IRC).} IRC~\citep{KW20} is a closely related field to IRL in partially observed environments. IRC models sub-optimality in decision makers as a misspecified reward function and aims to estimate this reward. The IRC task comprises two sub-tasks:\\
First, the inverse learner constructs a map from a continuous space of reward functions parameterized by $\theta$ to the reward's optimal policy. \\Second, based on a finite observation dataset $\dataset$, the underlying hyperparameter $\theta$ is estimated as the maximum likelihood estimate $\argmax_{\theta} \prob(\dataset|\theta)$.\vspace{0.1cm}\\
In comparison, our approach bypasses the first sub-task in IRC 
by
checking the feasibility of a finite set of convex inequalities. Given the information available to the inverse learner, these inequalities are both necessary and sufficient conditions for identifying optimality of a decision maker's decisions in multiple environments.
Indeed, increasing the number of environments in which the decision maker's actions are observed decreases the size of the feasible set of rationalizing rewards, and hence increases the precision of our set-valued IRL cost estimate.\footnote{Revealed preference micro-economics~\citep{MSCL78,RN15} have studied the consistency of the set-valued approach to eliciting agent rewards. \cite{MSCL78} specifies conditions under which the feasible set of utility functions reconstructed from a dataset of agent actions converges to a point for infinite datasets. \cite{RN15} constructs a quasi-concave utility function that rationalizes an infinite dataset.}}


{\em (d) Revealed Preference.} The key formalism used in this  paper to achieve IRL is {\em Bayesian revealed preferences} studied in microeconomics by~\citet{MR14,CM15,CD15,CD19}. 
Non-parametric estimation of cost functions given a finite length time series of decisions and budget constraints is the central theme in the area of classical (non-Bayesian) revealed preferences in microeconomics, starting with \cite{AF67,SM38} where necessary and sufficient conditions for constrained utility maximization are given; see also \cite{VR82,VAR12,Woo12} and more recently  in machine learning~\citep{LO09}.


{(e) Examples of Bayesian stopping time problems.} After constructing an IRL framework for general stopping time problems, this paper discusses  two important examples, namely,
inverse sequential hypothesis testing and inverse Bayesian search. Below we briefly motivate these examples.\\
{\em Example 1. Inverse Sequential Hypothesis Testing (SHT).} 
Sequential hypothesis testing (SHT)~\citep{Poo93,RS14} is  widely studied in detection theory.
The inverse SHT problem of
estimating misclassification costs by observing the decisions of an SHT agent has not been addressed. Estimating SHT misclassification costs is useful in adversarial inference problems. For example, by observing the actions of an adversary, an inverse learner can estimate the adversary's utility and predict its future decisions.

\noindent{\em Example 2. Inverse Bayesian Search}.
In Bayesian  search, each agent sequentially searches  locations until a stationary (non-moving) target is found. Bayesian search~\citep{RS14} is used in vehicular tracking~\citep{WN05}, image
processing~\citep{PL08} and cognitive radars~\citep{GD07}. IRL for Bayesian search requires the inverse learner to estimate the search costs by observing the search actions taken by a Bayesian search agent in multiple environments with different search costs. 

Bayesian search is a special case of the Bayesian multi-armed bandit problem \citep{Git89,BS12}. \blue{\em A promising extension of our IRL approach would be to solve inverse Bayesian bandit problems, namely, estimate the Gittins indices of the bandit arms.}
Regarding the literature in inverse bandits, \cite{CH19} propose a real-time assistive procedure for a human performing a bandit task based on the history of actions taken by the human. \cite{NTG21} solve the inverse bandit problem by assuming the inverse learner knows the variance of the stochastic reward; in comparison out setup assumes no knowledge of the rewards.

\section{Proof of Lemma~\ref{lem:relativeoptimality}}\label{appdx:lemrelopt}
\blue{{\em Proof.} Suppose $\datainf$ is generated by a Bayesian agent performing optimal stopping~(Definition~\ref{def:absoptimality}) in $\numagents$ environments. By definition, the following conditions hold:
\begin{equation}\label{eqn:absoptimality-lemma}
\begin{split}
    \stoptime_{\agent}(\belief,\funcstop)= \argmin_{\action\in\actionset} \belief'\utilityvec_{\agent,\action},~\netobjfun(\stoptime_{\agent},\utilitysymbolagent{\agent})=\inf_{\stoptime\in\stoptimeset}\netobjfun(\stoptime,\utilitysymbolagent{\agent}),
    \end{split}
\end{equation}
where $\netobjfun(\cdot)$~\eqref{eqn:opt_stop_objfun} is the expected cumulative cost comprising the expected stopping cost $\grosspayoff(\cdot)$ and expected cumulative continue cost $\sumruncostsymbol(\cdot)$
Since the set of chosen strategies $\boldsymbol{\stoptime}_{\agentset}\subset\boldsymbol{\stoptime}$, the set of {\em all} admissible policies, the feasibility of \eqref{eqn:absoptimality-lemma} implies the following conditions hold:
\begin{align}
    \stoptime_{\agent}(\belief,\funcstop)&= \argmin_{\action\in\actionset} \belief'\utilityvec_{\agent,\action},\nonumber\\
    \netobjfun(\stoptime_{\agent},\utilitysymbolagent{\agent})&=\min_{\stoptime\in\boldsymbol{\stoptime}_{\agentset}}\netobjfun(\stoptime,\utilitysymbolagent{\agent}).\label{eqn:relopt_stop_time}
\end{align}
Since $\boldsymbol{\stoptime}_{\agentset}$ is finite, the `inf' in \eqref{eqn:absoptimality-lemma} can be replaced with `min' in \eqref{eqn:relopt_stop_time}. The second condition in \eqref{eqn:relopt_stop_time} is simply a reformulation of \eqref{eqn:relopt_stop_time_continue_cost}. Hence, the `if' statement of Lemma~\ref{lem:relativeoptimality} is proved.}

\blue{We now prove the `only if' direction. Suppose the inverse learner has access to $\datainf$ aggregated from a Bayesian agent's actions in $\numagents$ environments. Specifically, the inverse learner only knows the agent's incurred expected costs finitely many policies $\stoptime_\agent\in\boldsymbol{\stoptime}_{\agentset}$. Alternatively, the sole knowledge of $\datainf$ implies that the inverse learner {\em does not know} the agent's expected stopping cost, nor the expected cumulative continue cost if the agent chooses any policy $\stoptime\in\boldsymbol{\stoptime}\backslash \boldsymbol{\stoptime}_{\agentset}$. Condition \eqref{eqn:opt_stop_act} is independent of the agent's policy, and only depends on the agent's stopping belief. However, \eqref{eqn:opt_stop_objfun} requires the inverse learner to compare the expected cost of the agent's strategy $\stoptime_\agent$ in environment $\agent$ against infinitely many strategies $\stoptime\in\boldsymbol{\stoptime}$. Due to inverse learner's limited knowledge, the best the inverse learner can do to check if \eqref{eqn:opt_stop_objfun} holds is to check the feasibility of \eqref{eqn:relopt_stop_time}.} \hfill \qedsymbol

\section{Proof of Theorem~\ref{thrm:NIAS_NIAC}}
\label{appdx:NIAS_NIAC}
We first introduce an observation likelihood $\fictobslike_{\agent}$ over a fictitious observation space $\fictobsset$ with generic element $\obsfict$ for stopping strategy $\stoptime_{\agent},\agent\in\agentset$:
\begin{equation}
\fictobslike_{\agent}(\obsfict|\state) = \sum_{\bar{\obs}:\belief_{\funcstop}=\belief}\left(\prod_{\timeinst=1}^{\funcstop}\oprob(\obs_{\timeinst},\state)\right)\label{eqn:att_fun_basic}
\end{equation}
Here $\bar{\obs}$ denotes a sequence of observations $\obs_1,\obs_2,\ldots$ and $\funcstop$ is the random stopping time for strategy $\stoptime_{\agent}$ defined in (\ref{eqn:def_stop_time}). $\fictobslike_{\agent}(\obsfict|\state)$ is the likelihood of all observation sequences $\bar{\obs}$ such that given true state $\tstate=\state$ and stopping strategy $\stoptime_{\agent}$, the agent's belief state at the stopping time $\funcstop$ is $\belief$.
Equivalently, $\fictobslike_{\agent}(\obsfict|\state)$ is the conditional probability density of the agent's stopping belief for strategy $\stoptime_{\agent}$. By definition, the mapping from $\obsfict$ to stopping belief $\belief$ is one-to-one. Hence, $|\fictobsset|=|\Delta(\stateset)|$, where $\Delta(\stateset)$ denotes the $\numstates-1$ dimensional unit simplex of pmfs.

Next, we re-formulate the expected stopping cost $\grosspayoff(\stoptime_{\agent},\utilitysymbolagent{\agent})$ defined in (\ref{eqn:opt_stop_objfun}) for stopping cost $\utilitysymbolagent{\agent}$ in terms of the fictitious observation likelihood defined in (\ref{eqn:att_fun_basic}).
\begin{align}
   \grosspayoff(\stoptime_{\agent},\utilitysymbolagent{\agent})&=\E_{\stoptime_{\agent}}\left\{ \belief_{\funcstop}'\utilityvec_{\agent,\action_{\funcstop}}\right\}=\int_{\fictobsset}\underbrace{\left(\sum_{\state\in\stateset}\fictobslikeagent{\agent}\prior(\state)\right)}_{\text{Marginal distribution of }\obsfict}~\min_{\action\in\actionset}\belief'\utilityvec_{\agent,\action}~d\obsfict\label{eqn:stoppingcost-reformulation}
\end{align}
In the above equation, the summation within the parentheses is the unconditional probability density of the stopping belief $\belief$ given stopping strategy $\stoptime_{\agent}$. Also, as described above, $\fictobslike_{\agent}(\obsfict|\state)$ is the likelihood of all observation sequences $\bar{\obs}$ such that given true state $\tstate=\state$ and stopping strategy $\stoptime_{\agent}$, the agent's belief state at the stopping time $\funcstop$ is $\belief$. We are now ready to prove necessity and sufficiency of the NIAS, NIAC inequalities (\ref{eqn:NIAS_thrm_CD}), (\ref{eqn:NIAC_thrm_CD}) in Theorem~\ref{thrm:NIAS_NIAC} for identifying an optimal stopping agent (Lemma~\ref{lem:relativeoptimality}).

\subsection{Necessity of NIAS, NIAC inequalities}
Recall from Theorem~\ref{thrm:NIAS_NIAC} that the analyst knows the agent's action selection policy in multiple environments. The action selection policy $\actselectagent{\agent}$ in $\datainf$ is a stochastically garbled version of $\fictobslike_{\agent}(\obsfict|\state)$ defined in (\ref{eqn:att_fun_basic}) and $\revpos{\agent}$ is a stochastic garbling of the agent's stopping belief $\belief$ when the stop action is $\action$. The action selection policy can be rewritten as follows
\begin{align}
    \actselectagent{\agent}& = \int_{\fictobsset} p_{\agent}(\action|\obsfict)\,\fictobslikeagent{\agent}~d\obsfict \\
    \implies \revpos{\agent}&  = \int_{\fictobsset}\frac{ p_{\agent}(\action|\obsfict)\fictobslikeagent{\agent}\prior(\state)}{p_{\agent}(\action)}~d\obsfict= \int_{\fictobsset} p_{\agent}(\obsfict|\action)~\belief(\state)~d\obsfict,\label{eqn:re_act_select}
\end{align}
where $\belief$ is the agent's stopping belief and $p_{\agent}(\obsfict|\action)$ is the probability density of the fictitious observations $\obsfict$ conditioned on the stop action $\action$.
\subsubsection{NIAS}
Let action $\action$ be the optimal stop action (\ref{eqn:opt_stop_act}) for stopping belief $\belief$ of the $\agent^{\text{th}}$ agent in $\actionset$. Then, 
\begin{align}
    & \sum_{\state\in \stateset} \belief(\state) (\utilityagent{\agent} - \utilitysymbolagent{\agent}(\state,b))\leq 0,~\forall \action,\actiontwo\in\actionset\label{eqn:ineq_1}\\
 	\implies & \int_{\fictobsset}\sum_{\state\in \stateset} \belief(\state) (\utilityagent{\agent} - \utilitysymbolagent{\agent}(\state,b)) ~p_{\agent}(\obsfict|\action)d\obsfict\leq 0 \label{eqn:ineq_2}\\
 	\implies & \sum_{\state\in \stateset}\left( \int_{\fictobsset} p_{\agent}(\obsfict|\action)~\belief(\state)~d\obsfict \right)(\utilityagent{\agent} - \utilitysymbolagent{\agent}(\state,b))\leq 0 \label{eqn:ineq_3}\\
 	\implies & \boxed{\sum_{\state\in \stateset} p_{\agent}(\state|\action) (\utilityagent{\agent} - \utilitysymbolagent{\agent}(\state,b))\leq 0.}\label{eqn:nias_nec}
\end{align}
Eq.\,\ref{eqn:ineq_1} says that the expected stop cost given belief $\belief$ is minimum for stop action $\action$. Here, the expectation is taken over the finite state set $\stateset$. The LHS of (\ref{eqn:ineq_2}) is the expected value of the LHS of (\ref{eqn:ineq_1}) taken over the space of fictitious observations $\fictobsset$ wrt the probability density $p_{\agent}(\obsfict|\action)$. Since $|\utilityagent{\agent}-\utilityagent{\agenttwo}|$ is bounded, the integral on the LHS of (\ref{eqn:ineq_2}) is finite. Hence, by Fubini's theorem, we can change the order of summation to get (\ref{eqn:ineq_3}). The first term in the integral of (\ref{eqn:ineq_3}) is equal to $p_{\agent}(\state|\action)$ from (\ref{eqn:re_act_select}) which results in the final NIAS inequality (\ref{eqn:nias_nec}).
\subsubsection{NIAC} 
Define $\grosspayoffsurr_{\agent,\agenttwo}$ as expected stopping cost when the fictitious observation likelihood is $\actselectagent{\agent}$ and stopping cost is $\utilityagent{\agenttwo}$. Then:
\begin{equation}
\label{eqn:exp_stopcost_surr}
    \grosspayoffsurr_{\agenttwo,\agent} = \sum_{\action\in\actionset}\left(\sum_{\state\in\stateset}\actselectagent{\agenttwo}\prior(\state)\right)\min_{b\in\actionset}\sum_{\state\in\stateset} p_{\agent}(\state|\action)\utilitysymbolagent{\agent}(\state,b).
\end{equation}
It follows from Blackwell dominance~\citep{BL53} that: 
\begin{align}
\grosspayoffsurr_{\agenttwo,\agent} \geq \grosspayoff(\stoptime_{\agenttwo},\utilitysymbolagent{\agent}) \text{ for all }\agent,~\agenttwo,\label{eqn:proof_nec_NIAC_1}\vspace{-0.3cm}
 \end{align}
 since the kernel $p_\agent(\obs_{1:\funcstop(\stoptime_\agent)}|\state)$ Blackwell dominates the action selection policy $p_\agent(\action|\state)$. A key observation is that, in \eqref{eqn:proof_nec_NIAC_1}, equality holds for $\agent=\agenttwo$ and is straightforward to show using Jensen's inequality. To summarize, we have the following inequality:
 \begin{equation}\label{eqn:}
     \grosspayoff_{\agent,\agent} = \grosspayoffsurr_{\agent,\agent},~\grosspayoff_{\agenttwo,\agent}\leq\grosspayoffsurr_{\agenttwo,\agent}
 \end{equation}
For any set of environment indices $\{\agent_1,\agent_2,\ldots \agent_I\}\subset\agentset,~(\agent_{I+1}=\agent_1)$, we have the following inequality from (\ref{eqn:relopt_stop_time_continue_cost}) in Lemma~\ref{lem:relativeoptimality}:
\begin{equation*}
\sum_{i=1}^I \grosspayoff(\stoptime_{\agent_{i+1}},\utilitysymbolagent{\agent_{i}}) -  \grosspayoff(\stoptime_{\agent_{i}},\utilitysymbolagent{\agent_i}) \geq \sum_{i=1}^I \sumruncostsymbol(\stoptime_{\agent_i}) - \sumruncostsymbol(\stoptime_{\agent_{i+1}})=0
\end{equation*}
Combining the inequalities (\ref{eqn:proof_nec_NIAC_1}) with the above inequality, we get the NIAC inequality:\vspace{-0.3cm}
\begin{equation}
\label{eqn:proof_NIAC_1}
    \sum_{i=1}^I \grosspayoffsurr_{\agent_{i+1},\agent_{i}} -  \grosspayoffsurr_{\agent_{i},\agent_i} \geq 0
\end{equation}

\subsection{Sufficiency of NIAS, NIAC inequalities for Bayes optimal stopping}
The inverse learner only has access to the agent's prior $\prior$ over the state space $\stateset$ and action selection policy $\actselectagent{\agent}$ induced by the agent's policy in environment $\agent$. For a finite set of fictitious observations, the sufficiency proof assumes that there exists a one-to-one correspondence between the fictitious observation $\obsfict$ to the terminal action $\action$. If the observation space $\obsset$ is continuous-valued, the sufficiency proof assumes there exist disjoint subsets $\fictobsset(\action)\subset\fictobsset$ and pdfs $f_\action$ with support $\fictobsset(\action)$ such that the fictitious observation likelihood $\fictobslikeagent{\agent}$ can be expressed as:
\begin{equation}\label{eqn:pre-image}
    \fictobslikeagent{\agent} = f_\action(\obsfict)~p_\agent(\action|\state),~\forall \state\in\stateset,\obsfict\in\fictobsset(\action),~\action\in\actionset.
\end{equation}
It follows straightforwardly from the fictitious observation likelihood expression in \eqref{eqn:pre-image} that, for all $\obsfict\in\fictobsset(\action)$, the belief is simply $p(\cdot|\action)$:
\begin{equation}
    p_\agent(\state|\obsfict) \underset{(\text{from \eqref{eqn:pre-image}})}{=} \frac{f_\action(\obsfict)~p_\agent(\action|\state)\prior(\state)}{\sum_{\state'}f_\action(\obsfict)~p_\agent(\action|\state')\prior(\state')} = \frac{p_\agent(\action|\state)\prior(\state)}{\sum_{\state'}p_\agent(\action|\state')\prior(\state')} = p_\agent(\state|\action)
\end{equation}
The important but subtle consequence of this assumption is that the expected stopping cost~\eqref{eqn:relopt_stop_time_continue_cost}
$\grosspayoff_{\agent,\agenttwo}$ is {\em equal} to the surrogate cost $\grosspayoffsurr_{\agent,\agenttwo}$~\eqref{eqn:exp_stopcost_surr} for all $\agent,\agenttwo\in\{1,2,\ldots,\numagents\}$.

\subsubsection{NIAS}
 Suppose NIAS inequality holds, that is, for all $\agent\in\agentset$,
 \begin{align}
    &\action=\argmin_{b\in\actionset} \sum_{\state\in\stateset} p_{\agent}(\state|\action) \utilitysymbolagent{\agent}(\state,b),~\forall \action\in\actionset.\label{eqn:proof_opt_stop_act_1}
 \end{align}
 Since the set $\{p_{\agent}(\state|\action),\action\in\actionset\}$ constitutes the set of all stopping beliefs when $\fictobslikeagent{\agent}=\actselectagent{\agent}$, the following condition holds from (\ref{eqn:proof_opt_stop_act_1}). 
 \begin{equation}
     \stoptime_{\agent}(\belief,\funcstop) =
    \argmin_{\action\in\actionset}\belief'\utilityvec_{\agent,\action}\label{eqn:proof_opt_stop_act}.
 \end{equation}
 Eq.\,\ref{eqn:proof_opt_stop_act} is precisely (\ref{eqn:opt_stop_act}), which says the agent chooses the stop action that minimizes its stopping cost given its stopping belief. Hence, it only remains to show that (\ref{eqn:relopt_stop_time_continue_cost}) in Lemma~\ref{lem:relativeoptimality} holds to complete our sufficiency proof.
 \subsubsection{NIAC}
Assuming the NIAS condition~\eqref{eqn:nias_nec} holds, we use the concept of KKT multipliers from duality theory~\cite[Sec.\,5.5]{BD04} to show that  NIAC~\eqref{eqn:proof_NIAC_1} is sufficient for \eqref{eqn:relopt_stop_time_continue_cost} in Lemma~\ref{lem:relativeoptimality} for optimal Bayesian stopping to hold. 
To do so, we use Lemma~\ref{lem:NIAC-variational} below for linear assignment problems to show the existence of scalars $\sumruncostagent{\agent}$ that satisfy \eqref{eqn:relopt_stop_time_continue_cost}; the feasibility inequality of interest is stated in \eqref{eqn:NIAC-feasibility-form} below. We now state Lemma~\ref{lem:NIAC-variational} which can be viewed as a variational form of the NIAC inequality:
\begin{lemma}\label{lem:NIAC-variational}
Suppose NIAC~\eqref{eqn:proof_NIAC_1} holds. Then:\\
(a) The solution of the following linear assignment problem is the identity map, that is, the optimal assignment map $x^\ast_{\agent,\agenttwo}$ is given by $x_{\agent,\agenttwo}^\ast=1$ if $\agent=\agenttwo$, and 0 otherwise:
\begin{align}\label{eqn:alt_opt_NIAC}
 &\operatorname{minimize}_{x_{\agent,\agenttwo}}~\sum_{\agent,\agenttwo=1}^{\numagents}~x_{\agent,\agenttwo}~\grosspayoffsurr_{\agent,\agenttwo},~\text{subject to:}\\
 & \sum_{\agenttwo}~x_{\agent,\agenttwo} \geq 1,~\sum_{\agent}~x_{\agent,\agenttwo} \geq 1,~ x_{\agent,\agenttwo} \geq 0~~\forall~\agent,\agenttwo \in\{1,2,\ldots,\numagents\}\nonumber.
\end{align}
\noindent (b) The KKT multipliers corresponding to the solution of the above assignment problem solve the feasibility condition of \eqref{eqn:relopt_stop_time_continue_cost} in Lemma~\ref{lem:relativeoptimality}.
\end{lemma}

\noindent {\em Proof.}\\ 
(a) Let $x_{\agent,\agenttwo}^\ast$ denote the optimal solution to the optimization problem~\eqref{eqn:alt_opt_NIAC}. Indeed, since \eqref{eqn:alt_opt_NIAC} is an LP, $x^\ast_{\agent,\agenttwo}\in\{0,1\}$. We can prove by contradiction that if NIAC~\eqref{eqn:proof_NIAC_1} holds, then the optimal assignment variables $x^\ast_{\agent,\agenttwo}$ is Kronecker delta, that is,
$x^\ast_{\agent,\agenttwo} = 1$ if $\agent=\agenttwo$ and $0$:\\
Choose any arbitrary feasible $x_{\agent,\agenttwo}\in\{0,1\}$. Consider the sequence of indices $I \equiv\{1,h_x(1),~h_x\circ h_x(1),~\ldots,~(h_x\circ)^{\numagents-2} h(1)\}$, where $h_x(\agent)=\agent'$ is the unique (due to assignment constraints in \eqref{eqn:alt_opt_NIAC}) index $\agent'\in\agentset$ for which $x_{\agent,\agent'}=1$ and `$\circ$' denotes the function composition operator. From invoking NIAC~\eqref{eqn:proof_NIAC_1} on the index sequence $I$, we observe that $\sum_{\agent,\agenttwo=1}^\numagents x_{\agent,\agenttwo} \grosspayoffsurr_{\agent,\agenttwo} \leq \sum_{\agent} \grosspayoffsurr_{\agent,\agent} =  \sum_{\agent,\agenttwo=1}^\numagents x^\ast_{\agent,\agenttwo} \grosspayoffsurr_{\agent,\agenttwo}$, where $x^\ast_{\agent,\agenttwo}=1$ if $\agent = \agenttwo$ and $0$ otherwise. Hence, the identity map solves the assignment problem~\eqref{eqn:alt_opt_NIAC}.\vspace{0.15cm}\\
(b) We now write down the Karush-Kuhn-Tucker (KKT) conditions~\citep[pg.\, 121]{BD04} for the assignment problem \eqref{eqn:alt_opt_NIAC} at the optimal solution $\{x^\ast_{\agent,\agenttwo}\}_{\agent,\agenttwo=1}^\numagents$ that are first-order necessary conditions for optimality:
\begin{equation}
\begin{split}
&\text{There exist scalars } \lambda_{1,\agent},~\lambda_{2,\agent},~\lambda_{3,\agent,\agenttwo}\geq 0,~\agent,\agenttwo\in\{1,2,\ldots\numagents\},\text{ such that:}\\
    &\text{(i) For } \agenttwo=\agent:~ \grosspayoffsurr_{\agent,\agent} = \lambda_{1,\agent} + \lambda_{2,\agent},~\text{(ii) For }\agenttwo\neq \agent:~ \grosspayoffsurr_{\agenttwo,\agent} =  \lambda_{1,\agent} + \lambda_{2,\agenttwo} + \lambda_{3,\agenttwo,\agent}.
\end{split}
\label{eqn:KKT-LP}
\end{equation}
The scalars $\lambda_{1,\agent}$ and $\lambda_{2,\agenttwo}$ in \eqref{eqn:KKT-LP} correspond to KKT multipliers associated with the inequality constraints $(-\sum_{\agenttwo}~x_{\agent,\agenttwo}) \leq -1$ and $(-\sum_{\agent}~x_{\agent,\agenttwo}) \leq -1$ in \eqref{eqn:alt_opt_NIAC}, respectively. We note that both sets of inequalities are active at $\{x^\ast_{\agent,\agenttwo}\}_{\agent,\agenttwo=1}^{\numagents}$, the solution of \eqref{eqn:alt_opt_NIAC}. The scalar $\lambda_{\agent,\agenttwo}$ is the KKT multiplier associated with the inequality constraint $(-x_{\agent,\agenttwo})\leq 0$, where the inequality is active only for $x^\ast_{\agent,\agenttwo},~\agent \neq \agenttwo$. For any pair of environments $\agent,\agenttwo\in\{1,2,\ldots,\numagents\}$, the following inequalities result due to the KKT conditions in \eqref{eqn:KKT-LP}:
\begin{align}
    &\grosspayoffsurr_{\agent,\agent} - \lambda_{2,\agent} = \grosspayoffsurr_{\agenttwo,\agent} - \lambda_{2,\agenttwo} - \lambda_{3,\agenttwo,\agent} \leq  \grosspayoffsurr_{\agenttwo,\agent} - \lambda_{2,\agenttwo}~~~(\text{since }\lambda_{3,\agenttwo,\agent}\geq 0)\nonumber\\
    \Leftrightarrow~& \grosspayoffsurr_{\agent,\agent} + (\max_{\agent'} \lambda_{2,\agent'}- \lambda_{2,\agent}) \leq  \grosspayoffsurr_{\agenttwo,\agent} + (\max_{\agent'} \lambda_{2,\agent'} - \lambda_{2,\agenttwo})\nonumber\\
    \Leftrightarrow~&\grosspayoffsurr_{\agent,\agent} + \sumruncostagent{\agent} \leq \grosspayoffsurr_{\agenttwo,\agent} + \sumruncostagent{\agenttwo}\label{eqn:NIAC-feasibility-form}\\
     &(\text{by replacing $(\max_{\agent'} \lambda_{2,\agent'} - \lambda_{2,\agent})$ with the variable $\sumruncostagent{\agent}$ for all $\agent\in\agentset$})\nonumber
     \end{align}

We now reconstruct an estimate of the agent's expected continue cost $\widehat{\sumruncostagent{}}$ below and show \eqref{eqn:relopt_stop_time_continue_cost} holds for Bayes optimal stopping. With $\boldsymbol{p}_{\stoptime}=p_{\stoptime}(\action|\state)$ denoting the action selection policy induced by a stopping strategy $\stoptime$, consider the following reconstructed estimate of the agent's expected continue cost $\widehat{\sumruncostagent{}}(\stoptime)$ in terms of the feasible variables $\{\sumruncostagent{\agent}\}_{\agent=1}^\numagents$~\eqref{eqn:NIAC-feasibility-form}:
\begin{equation}\label{eqn:reconstructed-RI-cost}
\begin{split}
    \widehat{\sumruncostagent{}}(\stoptime) & = \max_{\agent=1,2,\ldots,\numagents}~\left\{
     \sumruncostagent{\agent} + \grosspayoff_{\agent,\agent} - \grosspayoffsurr(\stoptime,\utilitysymbolagent{\agent})
     \right\},\text{ where}\\
     \grosspayoffsurr(\stoptime,\utilitysymbolagent{\agent}) & =\sum_{\action\in\actionset}\left(\sum_{\state\in\stateset}p_{\stoptime}(\action|\state)\prior(\state)\right)\min_{b\in\actionset}\sum_{\state\in\stateset} p_{\stoptime}(\state|\action)\utilitysymbolagent{\agent}(\state,b)
\end{split}
\end{equation}
In \eqref{eqn:reconstructed-RI-cost}, $\grosspayoffsurr(\stoptime,\utilitysymbolagent{\agent})$ denotes the expected stopping cost of the Bayesian agent with strategy $\stoptime$ and stopping costs $\utilityagent{\agent}$ {\em assuming a one-to-one map between the set of observations $\obs_{1:\funcstop(\stoptime)}$ to action $\action$}.\footnote{While it may seem counter-intuitive to assume a one-to-one mapping from the  fictitious observation space to action space, one can show for convex costs like entropic costs (Shannon-Gibbs, R{\'e}nyi and Tsallis) that the {\em optimal} mapping is one-to-one. The key idea is to show that having a many-to-one map with the same expected stopping cost is sub-optimal in that the agent incurs a strictly larger expected continue cost; see \citet[Lemma~1]{MM15} for a more detailed explanation.
} The variable $p_\stoptime(\state|\action)$ is the posterior belief of the state computed using Bayes rule as:
\begin{equation*}
    p_\stoptime(\state|\action) = \frac{\prior(\state)p_\stoptime(\action|\state)}{\sum_{\state'}\prior(\state')p_\stoptime(\action|\state')}
\end{equation*}
Indeed, if the mapping from the fictitious observation set $\fictobsset$ to the action set $\actionset$ is assumed to be one-to-one, the expected stopping cost can be expressed in terms of the action selection policy $\boldsymbol{p}_{\stoptime}$ induced by the stopping strategy $\stoptime$. From \eqref{eqn:NIAC-feasibility-form}, it is straightforward to show that $\sumruncostagent{}(\stoptime_\agent) = \sumruncostagent{\agent}$. Hence, replacing $\sumruncostagent{\agent}$ in \eqref{eqn:NIAC-feasibility-form} with $\widehat{\sumruncostagent{}}(\stoptime_\agent)$ yields the following inequalities:
\begin{equation} \label{eqn:reconstruction-cost-feasible}
\begin{split}
&\grosspayoffsurr_{\agent,\agent} + \widehat{\sumruncostagent{}}(\stoptime_\agent) \leq \grosspayoffsurr_{\agenttwo,\agent} + \widehat{\sumruncostagent{}}(\stoptime_\agenttwo)\\
    \Leftrightarrow~&\text{A cumulative running cost can be reconstructed~\eqref{eqn:reconstructed-RI-cost} such that condition \eqref{eqn:relopt_stop_time_continue_cost} in Lemma~\ref{lem:relativeoptimality}}\\
    &\text{holds with expected stopping costs } \grosspayoffsurr_{\agenttwo,\agent},~\agent,\agenttwo \in \{1,2,\ldots,\numagents\}.~\text{\qedsymbol}
\end{split}
\end{equation}
In words, for a feasible set of stopping costs $\{\utilitysymbolagent{\agent}\}_{\agent=1}^\numagents$ such that NIAS and NIAC hold, the Bayesian agent's (unobserved) strategies satisfy optimal Bayesian stopping~\eqref{eqn:relopt_stop_time_continue_cost}. Moreover, for every feasible set of costs $\{\utilitysymbolagent{\agent}\}_{\agent=1}^\numagents$, the term $(\max_{\agent'} \lambda_{2,\agent'} - \lambda_{2,\agent})$ denotes the expected continue cost incurred by the Bayesian agent due to choosing strategy $\stoptime_\agent$, and $\grosspayoffsurr_{\agent,\agenttwo}$ denotes the agent's incurred expected stopping cost in environment $\agenttwo$ if it chooses strategy $\stoptime_\agent$.

\subsection{Remarks}\label{appdx:NIAS-NIAC-remarks}
\blue{
 1. {\em IRL for inverse SHT.} For the inverse SHT problem discussed in Sec.\,\ref{sec:SHT}, the inverse learner knows $\sumruncostsymbol_{\agent}$, the expected cumulative continue cost for the agent in environment $\agent$. Hence, the inverse learner can identify optimal SHT simply by checking if the NIAS inequality~\eqref{eqn:nias_nec} and the following inequality is feasible:
 \begin{equation}\label{eqn:niac_dag_origin}
 \grosspayoffsurr_{\agent,\agent}-\grosspayoffsurr_{\agenttwo,\agent}\leq \sumruncostsymbol_{\agenttwo}-\sumruncostsymbol_{\agent},~\forall \agent,\agenttwo\in\agentset,
 \end{equation}
 where $\grosspayoffsurr(\cdot)$ is the expected stop cost defined in \eqref{eqn:exp_stopcost_surr} and $\sumruncostsymbol_{\agent}$ is the expected continue cost of the agent in environment $\agent$ now known to the inverse learner. Due to \ref{asmp:SHT3},  $\grosspayoffsurr_{\agent,\cdot} = \grosspayoff_{\agent,\cdot}$. Hence,~\eqref{eqn:niac_dag_origin} is equivalent to \eqref{eqn:relopt_stop_time_continue_cost} in Lemma~\ref{lem:relativeoptimality}. We term the inequality in \eqref{eqn:niac_dag_origin} as NIAC$\ast$ and use it in Theorem~\ref{thrm:classic_SHT} for IRL for inverse SHT.\\
 2. {\em Different observation likelihoods for different environments.} Theorem~\ref{thrm:NIAS_NIAC} is a purely {\em data-centric} approach for IRL that makes no assumptions on the agent's observation likelihood. If the NIAS and NIAC inequalities have a feasible solution, then the inverse learner's dataset $\datainf$ can be rationalized by a Bayesian agent that acts optimally (in the sense of Lemma~\ref{lem:relativeoptimality}) and has a fixed observation likelihood over all $\numagents$ environments. It may very well be the case that the Bayesian agent has a different observation likelihood in different environments, but Theorem~\ref{thrm:NIAS_NIAC} is opaque to this condition.}
 
 \blue{If the inverse learner knows {\em a priori} that the Bayesian agent uses a different observation likelihood for different environments, we need stronger conditions to achieve IRL. A sufficient condition for identifying optimal Bayesian stopping with distinct observation likelihoods in different environments is to assume the expected cumulative continue cost of the agent is independent of the observation likelihood. One example that satisfies this assumption is the entropic continue cost:
\begin{equation}\label{eqn:entropy_continue}
	\runcostinst_{\timeinst} = \lambda~(H(\pi_{\timeinst}) - \E  \{ H(\pi_{\timeinst+1}) \vert \pi_\timeinst \} ),~\timeinst\geq 0,~\lambda>0
\end{equation}
where $H(p) = -\sum_{i} p_i \operatorname{log}(p_i)$ is the entropy of pmf $p$. The above choice of continue cost has two advantages:\\
(i) The expected {\em cumulative} continue cost for agent $\agent$ is simply $H(\prior) - \E_{\action} \{ H(\actselectagent{\agent})\}$, and is independent of the observation likelihood; see \citet[Lemma~1]{MM15} for a discussion on how conditioned on state $\state$, the optimal mapping from the space of fictitious observations $\obsfict$~\eqref{eqn:att_fun_basic} to the space of actions $\actionset$ is one-to-one due to the convexity of the entropic cost.\\
(ii) The inverse learner can test for `absolute optimality'~\eqref{eqn:opt_stop_act},~\eqref{eqn:opt_stop_time} of the Bayesian agent's decisions and does not require observing the agent's behavior in multiple environments. Using the method of Lagrange multipliers, it is straightforward to show that for environment $\agent$, the following relation holds between the agent's stopping costs and its observed decisions for optimal Bayesian stopping:
\begin{equation}\label{eqn:optstopcostentropy}
p_\agent(\action|\state) = \frac{p_\agent(\action)~\exp (-\utilityagent{\agent}/\lambda)}{\sum_{\actiontwo\in\actionset}p_\agent(\actiontwo)~\exp (-\utilitysymbolagent{\agent}(\state,\actiontwo)/\lambda) },~\forall~\action,\state,\agent,
\end{equation}
where $\lambda>0$ is a feasible variable that parametrizes the continue cost~\eqref{eqn:entropy_continue}, and $p_\agent(\action)$ is the marginal distribution of the action $\action$ in environment $\agent$. IRL is achieved by checking for the feasibility condition of \citet[Eq.~3,~Proposition~1]{CD19} and solving the above set of equations \eqref{eqn:optstopcostentropy} for $\utilityagent{\agent}$; observe that there is no assumption of a fixed observation likelihood for the Bayesian agent across environments and the IRL estimate returns an ordinal estimate of the agent's stopping costs.
}

\section{Proof of Theorem~\ref{thrm:Search}}\label{appdx:proof_Search}
 We will show (\ref{eqn:NIAC_dag_def}) is equivalent to the condition for identifying search optimality (\ref{eqn:relopt_search}) in two steps. For a fixed stationary search strategy~$\stoptime:\belief\to\action$ and search cost $\{\searchcostsymbol(\action),\action\in\actionset\}$, we first express the expected cumulative search cost in terms of the search action policy $\occsymb(\state,\action)$ (\ref{eqn:dataset_search}) and the prior $\prior$.
 \begin{align*}
     \netobjfun(\stoptime,\searchcostsymbol)& = \E_{\stoptime}\left\{\sum_{\timeinst=1}^{\funcstop} \searchcostsymbol(\stoptime(\belief_{\timeinst}))\right\}= \E_{\stoptime}\left\{\sum_{\action\in\actionset}\searchcostsymbol(\action)\left(\sum_{\timeinst=1}^{\funcstop} \mathbbm{1}\{\stoptime(\belief_{\timeinst})=\action\}\right)\right\}\\
     & = \sum_{\action\in\actionset}\searchcostsymbol(\action)\sum_{\state\in\stateset}\prior(\state)\E_{\stoptime}\left\{\sum_{\timeinst=1}^{\funcstop} \mathbbm{1}\{\stoptime(\belief_{\timeinst})=\action\}|\state\right\} = \sum_{\state\in\stateset,\action\in\actionset}\prior(\state)\occsymb(\state,\action)\searchcostsymbol(\action).
 \end{align*}

Now, consider the set of search strategies $\{\stoptime_{\agent},\agent\in\agentset\}$.
\begin{align*}
&(\ref{eqn:relopt_search})\equiv~\stoptime_{\agent} \in \argmin_{\{\stoptime_{\agenttwo},\agenttwo\in\agentset\}} \netobjfun(\stoptime_{\agenttwo},\searchcostsymbol_{\agent})\Longleftrightarrow  \netobjfun(\stoptime_{\agent},\searchcostsymbol_{\agent}) -\netobjfun(\stoptime_{\agenttwo},\searchcostsymbol_{\agent})\leq 0,~\agent,\agenttwo\in\agentset.\\
&\Longleftrightarrow  \sum_{\state\in\stateset,\action\in\actionset} \prior(\state)(\occagent{\agent}-\occagent{\agenttwo})~\searchcostagent{\agent}\leq 0\equiv(\ref{eqn:NIAC_dag_def}). 
\end{align*}
\hfill \qedsymbol

\section{Proof of Theorem~\ref{thrm:finite_SHT}}
\label{appdx:proof_finite_bound_SHT}
We divide the proof of Theorem~\ref{thrm:finite_robustness} into 4 steps:\\
\noindent {\em Step 1. Using Dvoretzky-Kiefer-Wolfowitz (DKW) inequality~\citep{KS07} to bound the deviation of the empirically computed action selection policy $\actselectagentemp{\agent}$ from $\actselectagent{\agent}$.}\\ 
The DKW inequality~\citep{VAN00} provides a \blue{finite sample characterization of the asymptotic result} of Glivenko-Cantelli theorem by  quantifying the convergence rate of the empirical cdf to the true cdf. Let $\actcdfagent{\agent}$ and $\actcdfagentemp{\agent}$ denote the cdfs of $\actselectagent{\agent}$ and $\actselectagentemp{\agent}$, respectively. From the two-sided DKW inequality, the following inequalities result:
\begin{align*}
    &1- 2\exp\left(-2\numtrials_{\state,\agent}\eps^2\right)\leq\prob\left(\max_{\action\in\actionset}|\actcdfagentemp{\agent}-\actcdfagent{\agent}|<\eps\right)
    \leq\prob\left(\max_{\action\in\actionset}|\actcdfagentemp{\agent}-\actcdfagent{\agent}|\leq\eps\right)\\
    &\leq \prob\left(|\actselectagent{\agent}-\actselectagentemp{\agent}|\leq \eps,\forall\action\right)\leq \prob\left(\sum_{\action\in\actionset}|\actselectagent{\agent}-\actselectagentemp{\agent}|^2\leq \numactions\eps^2\right).
\end{align*}
For a fixed state $\state$ and environment $\agent$, let $\eps_{\state,\agent}$ bound the error $|\actselectagentemp{\agent}-\actselectagent{\agent}|,~\forall \action\in\actionset$. With $\eps_{\max}^2 = \numactions (\sum_{\state,\agent}\eps_{\state,\agent}^2)$, we have the following probabilistic bound on the $L_2$-error between the true and empirical action selection policies, summed over all states, actions and environments:
\begin{align}
    &\boxed{\prob\left(\sum_{\action,\state,\agent}|\actselectagent{\agent}-\actselectagentemp{\agent}|^2\leq\eps_{\max}^2 \right)\geq \prod_{\state,\agent}1-2\exp\left(-2\numtrials_{\state,\agent}\eps_{\state,\agent}^2\right)}.\label{eqn:sht_finite_1}
\end{align}
\vspace{0.15cm}

\noindent {\em Step 2. Using Hoeffding's Inequality \citep{BLM13} to bound the deviation of the sample average of the SHT stopping times $\hat{\sumruncostsymbol}_{\agent}$ from the true value $\sumruncostagent{\agent}$.}\\
The inverse learner knows the agent's stopping time $\funcstop\in[1,\funcstop_{\max}]$ for all $\numagents$ environments. Analogous to \eqref{eqn:sht_finite_1}, for a fixed environment $\agent$, let $\eta_{\agent}$ bound the error $|\sumruncostagentemp{\agent} - \sumruncostagent{\agent}|$. With $\eta_{\max}^2 = \sum_{\agent}\eta_{\agent}^2$, we have the following probabilistic bound on the $L_2$-error between the true and empirical expected stopping times of the agent, summed over all $\numagents$ environments via the two-sided Hoeffding's inequality:
\begin{align}
    &\prob(|\sumruncostagentemp{\agent} - \sumruncostagent{\agent}|\leq \eta_{\agent})\geq 1-2\exp\left(-2\numtrials_{\agent}\eta_{\agent}^2/\funcstop_{\max}^2\right)\nonumber\\
    \implies~&\prob(\sum_{\state,\agent}|\sumruncostagentemp{\agent} - \sumruncostagent{\agent}|^2\leq \eta_{\max}^2)\geq \prob(|\sumruncostagentemp{\agent} - \sumruncostagent{\agent}|\leq \eta_{\agent},~\forall\agent\in\agentset)\nonumber\\
    \implies~&\boxed{\prob(\sum_{\state,\agent}|\sumruncostagentemp{\agent} - \sumruncostagent{\agent}|^2\leq \eta_{\max}^2)\geq \prod_{\agent} 1-2\exp\left(-\frac{2\numtrials_{\agent}\eta_{\agent}^2}{\funcstop_{\max}^2}\right)},~\text{where }\numtrials_{\agent}=\sum_{\state}\numtrials_{\state,\agent}\label{eqn:sht_finite_2}
\end{align}\vspace{0.15cm}

{\em Step 3. Using the union bound on error bounds from steps 1 and 2 to bound the cumulative deviation of empirically computed action selection policies and expected stopping times.}\\
Our aim is to construct a tight bound on the probability of the event $E_{pert}(\delta_{\max})$, where $E_{pert}(\delta_{\max})$ is defined as:
\begin{equation}
  E_{pert}(\delta_{\max}) \equiv \{~\{\actselectagentemp{\agent},\sumruncostagentemp{\agent}\}~\big| \sum_{\state,\agent}|\actselectagent{\agent}-\actselectagentemp{\agent}|^2 + \sum_{\state}|\sumruncostagent{\agent}-\sumruncostagentemp{\agent}|^2\leq\delta_{\max}^2\},\label{eqn:e_pert}
\end{equation}
We note that $\prob(E_{pert}(\delta_{\max}))$ bounds the Type-I and Type-II IRL error probabilities for suitable choices of $\delta_{\max}$. The Type-I error probability is bounded by $1-\prob(E_{pert})$ when $\delta_{\max}^2$ in (\ref{eqn:e_pert}) is set to $\epsone{\datafin{\trialset}}$. Also, the Type-II error probability is bounded by $1-\prob(E_{pert})$ when $\delta_{\max}^2$ in (\ref{eqn:e_pert}) is set to $\epstwo{\datafin{\trialset}}$. \blue{Recall from \eqref{eqn:eps1},~\eqref{eqn:eps2} that $\epsone{\datafin{\trialset}}$ and $\epstwo{\datafin{\trialset}}$ correspond to the minimum $L_2$-perturbation needed for the {\em finite} IRL dataset $\datafin{\trialset}$~\eqref{eqn:dataset_finitesample} to pass and fail, respectively, the NIAS and NIAC conditions of Theorem~\ref{thrm:NIAS_NIAC} for inverse optimal Bayesian stopping.}

For a fixed error tuple $\{\eps_{\state,\agent},\eta_\agent\}$, consider the surrogate event $E(\{\eps_{\state,\agent},\eta_{\agent}\})$ defined as:
\begin{align}
E(\{\eps_{\state,\agent},\eta_\agent\})=&\{~\{\actselectagentemp{\agent},\sumruncostagentemp{\agent}\}~\big|~\actselectagentemp{\agent}-\actselectagent{\agent}|\leq\eps_{\state,\agent},|\sumruncostagentemp{\agent}-\sumruncostagent{\agent}|\leq\eta_{\agent},~\forall~\action,\state,\agent\}.\label{eqn:event_1}
\end{align}
Clearly, $E(\{\eps_{\state,\agent},\eta_\agent\})~\subseteq~E_{pert}(\delta_{\max})$ if $\delta_{\max}^2$ in \eqref{eqn:e_pert} is equal to $\numactions\sum_{\state,\agent}\eps_{\state,\agent}^2 +\sum_{\agent} \eta_{\state,\agent}^2$. Combining the error bounds in (\ref{eqn:sht_finite_1}) and (\ref{eqn:sht_finite_2}) via a union bound to bound $\prob(E(\{\eps_{\state,\agent},\eta_{\agent}\}))$ yields the following inequality:
\begin{align}
    & \prob(E_{pert}(\delta_{\max}))\geq \prob(E(\{\eps_{\state,\agent},\eta_{\agent}\}))\nonumber\\
    \geq~& \prod_{\state,\agent}1-2\exp\left(-2\numtrials_{\state,\agent}\eps_{\state,\agent}^2\right)\prod_{\agent} 1-2\exp\left(-2\numtrials_{\agent}\eta_{\agent}^2/\funcstop_{\max}^2\right)\label{eqn:sht_markov}\\
    \geq~& 1-\sum_{\state,\agent}2\exp\left(-2\numtrials_{\state,\agent}\eps_{\state,\agent}^2\right) -\sum_{\agent}2\exp\left(-2\numtrials_{\agent}\eta_{\agent}^2/\funcstop_{\max}^2\right)\label{eqn:sht_finite_3}\\
    \implies & \boxed{\prob(E_{pert}(\delta_{\max})) \geq 1-\sum_{\state,\agent}2\exp\left(-2\numtrials_{\state,\agent}\eps_{\state,\agent}^2\right) -\sum_{\agent}2\exp\left(-2\numtrials_{\agent}\eta_{\agent}^2/\funcstop_{\max}^2\right)},\label{eqn:sht_finite_union_approx}
\end{align}
where $\delta_{\max}^2 = \numactions\sum_{\state,\agent}\eps_{\state,\agent}^2 + \sum_{\agent}\eta_{\agent}^2$. The inequality in \eqref{eqn:sht_markov} is simply a union bound on the error bounds in \eqref{eqn:sht_finite_1} and \eqref{eqn:sht_finite_2}. The inequality in \eqref{eqn:sht_finite_3} holds due to Assumption~\ref{asmp:IRL_finite_SHT_3} that says the analyst observes the agent's stopping action over sufficiently many trials. If the expected stopping time is known accurately, that is, $\sumruncostagentemp{\agent} = \sumruncostagent{\agent}$ for all $\agent\in\agentset$, Assumption~\ref{asmp:IRL_finite_SHT_3} specializes to Assumption~\ref{asmp:IRL_finite_3}.

\noindent {\em Step 4. Obtaining a tight bound on the error probability computer in step 3.}\\
 Eq.\,\ref{eqn:sht_finite_union_approx} provides a probabilistic bound on the perturbation of the empirical dataset $\datafin{\trialset}$ from the asymptotic dataset $\datainf$ in terms of the sample size $\trialset = \{\numtrials_{\state,\agent},\state\in\stateset,\agent\in\agentset\}$, for a fixed error sequence $\{\eps_{\state,\agent},\eta_\agent\}$. Our final step is to maximize the RHS in~\eqref{eqn:sht_finite_union_approx} subject to the constraint $\numactions(\sum_{\state,\agent}\eps_{\state,\agent}^2)+\sum_{\agent}\eta_{\agent}^2 = \delta_{\max}^2$ to obtain the tightest bound on $\prob(E_{pert}(\delta_{\max}))$~\eqref{eqn:sht_finite_union_approx}. Equivalently, our aim is to minimize the following objective function:
 \begin{align}
\hspace{-0.4cm}&\boxed{\min_{\{\eps_{\state,\agent}^2,\eta_{\agent}^2\}\geq 0} \sum_{\state,\agent}2\exp\left(-2\numtrials_{\state,\agent}\eps_{\state,\agent}^2\right) +\sum_{\agent}2\exp\left(-2\numtrials_{\agent}\eta_{\agent}^2/\funcstop_{\max}^2\right),~\text{s.t. } \numactions\sum_{\state,\agent}\eps_{\state,\agent}^2+\sum_{\agent}\eta_{\agent}^2=\delta_{\max}^2}.\label{eqn:opt_finite_sht_1}
\end{align}
We observe that the terms $\exp\left(-2\numtrials_{\state,\agent}\eps_{\state,\agent}^2\right)$ and $\exp\left(-2\numtrials_{\agent}\eta_{\agent}^2/\funcstop_{\max}^2\right)$ are convex in $\eps_{\state,\agent}^2$ and $\eta_{\agent}^2$, respectively. Also, Assumption \ref{asmp:IRL_finite_SHT_2} ensures the values of the terms $\eps_{\state,\agent}^2,\eta_{\agent}^2$ are bounded away from $0$ at the local optimum of \eqref{eqn:opt_finite_sht_1} computed via the method of Lagrange multipliers. Assumption \ref{asmp:IRL_finite_SHT_3} ensures the $\eps_{\state,\agent}^2,\eta_{\agent}^2$ in (\ref{eqn:def_opt_finite}) satisfy the Slater's condition for regularity. Since the equality constraint is linear, and the objective function is convex in the feasible variables $\eps_{\state,\agent}^2$ and $\eta_{\agent}^2$, \eqref{eqn:opt_finite_sht_1} constitutes a convex optimization problem whose solution can be computed using the method of Lagrange multipliers (since Assumption~\ref{asmp:IRL_finite_SHT_2} ensures inactive inequality constraints at the optimal solution).
Finally, the solution of the optimization problem~\eqref{eqn:opt_finite_sht_1} can be expressed as:
\begin{align}
    &~\boxed{\eps_{\state,\agent}^2 = (\numactions\widetilde{\numtrials}_{\state,\agent}/2)\left(\ln(\lambda)+\ln(2\numtrials_{\state,\agent}/\numactions)\right),~\eta_{\state,\agent}^2= (\widetilde{\bar{\numtrials}}_{\agent}/2)(\ln(\lambda)+\ln(2\bar{\numtrials}_{\agent}))},~\text{where}\nonumber\\
    &~\ln(\lambda)=\frac{\delta_{\max}^2-\numactions\sum_{\state,\agent}\ln((2\numtrials_{\state,\agent}/\numactions)^{\numactions\widetilde{\numtrials}_{\state,\agent}/2}) - \sum_{\agent} \ln(2\bar{\numtrials}_{\agent}^{\widetilde{\bar{\numtrials}}_{\agent}/2}) }{(\numactions\sum_{\state,\agent}\widetilde{\numtrials}_{\state,\agent}+ \sum_{\agent\in\agentset}\widetilde{\bar{\numtrials}}_{\agent})/2}~~.\label{eqn:def_opt_finite}
\end{align}
In the above equations, $\bar{\numtrials}_{(\cdot)}=\numtrials_{(\cdot)}/\funcstop_{\max}^2$,~$\widetilde{\numtrials}=\numtrials^{-1}$.
Subtracting the objective function of (\ref{eqn:opt_finite_sht_1}) evaluated at the optimal values of $\eps_{\state,\agent}^2$ and $\eta_\agent^2$~\eqref{eqn:def_opt_finite} from $1$, and setting $\delta_{\max}^2$ to $\epsone{\datafin{\trialset}}$ and $\epstwo{\datafin{\trialset}}$, respectively, yield lower bounds for Type-I and Type-II error probabilities of the IRL detector, respectively.\hfill\qedsymbol

{\em Remark.} The proof of Theorem~\ref{thrm:finite_robustness} for finite sample complexity of Theorem~\ref{thrm:NIAS_NIAC} is identical to the above proof structure (except that there is no step 2) and hence, omitted for brevity.

\section{Proof of Theorem~\ref{thrm:finite_Search}}
\label{appdx:proof_finite_bound_search}
To prove Theorem~\ref{thrm:finite_Search}, we first state and prove an auxiliary result, namely, Proposition~\ref{prop:Search_bound}, below. Theorem~\ref{thrm:finite_Search} is a special case of Proposition~\ref{prop:Search_bound} as discussed below.

\begin{proposition} 
\label{prop:Search_bound}Given dataset $\datafinsearch{\trialset}$ and \ref{asmp:IRL_finite_Search_2}, the 
deviation of the finite sample search action policy $\occagentemp{\agent}$ and the true search action policy $\occagent{\agent}$
can be bounded in terms of the number of samples $\trialset=\{\numtrials_{\state,\agent}\}$ as follows.
 \begin{equation}
 \label{eqn:search_bound}
        \prob\left(\sum_{\action,\state,\agent} |\occagent{\agent}-\occagentemp{\agent}|^2 \leq \epsilon\right)
        \geq 1 - \sum_{\state,\agent} \frac{u(\datafinsearch{\trialset})}{\epsilon\numtrials_{\state,\agent}^{1/2}}, 
    \end{equation}
    where $u(\datafinsearch{\trialset})=\frac{(1-\revprobprime)}{(\revprobprime)^2}\numstates\sum_{\state,\agent}\numtrials_{\state,\agent}^{-1/2}$ and $\occagentemp{\agent}$ is the sample average of the number of times the agent searches location $\action$ given state $\state$ in environment $\agent$. 
\end{proposition}
{\bf Proof.} Assumption \ref{asmp:IRL_finite_Search_2} implies that for any environment $\agent\in\agentset$, given prior $\prior$, the agent's optimal search sequence $\action_0,\action_1,\action_2,\ldots$ is periodic, i.e.\,, $\action_{\timeinst}=\action_{\timeinst + \numactions}$. In other words, in any interval of $\numactions$ time steps, the agent searches each location exactly once in a particular (unknown to the inverse learner) order.

Consider the agent's search policy  $\occagent{\agent}$ defined in (\ref{eqn:dataset_search}). Below we express $\occagent{\agent}$ in terms of the pdf of the stopping time of the search process.
\begin{align*}
    \occagent{\agent}& = \E_{\stoptime_{\agent}}\left\{\sum_{t=1}^{\funcstop}
    \mathbbm{1}\{\stoptime_{\agent}(\belief_{\timeinst}) = \action\} |\tstate=\state
    \right\}= \sum_{\timeinst=1}^{\infty}\prob_{\stoptime_{\agent}}\left( \funcstop=\timeinst|\tstate=\state\right)(\floor{\timeinst/\numstates} + r(\state,\action)).\\
    & = \sum_{\timeinst=1}^{\infty}\floor{\timeinst/\numstates}\prob_{\stoptime_{\agent}}\left( \funcstop=\timeinst|\tstate=\state\right) + r(\state,\action) = \E_{\stoptime_{\agent}}\{\floor{\funcstop/\numstates}~|\state\} + r_{\agent}(\state,\action)=\frac{1}{\revprobsymb} + r_{\agent}(\state,\action).
\end{align*}
Here, $\revprobsymb$ denotes the reveal probability of the agent and $\floor{\cdot}$ denotes the floor function. $r(\state,\action)=1$ if agent  searches location $\action$ prior to location $\state$ in one search cycle from time $\timeinst=0\to\numstates-1$ in environment $\agent$, and $0$ otherwise. The final equality follows  from the fact that conditioned on the true state $\tstate=\state$, the random variable $\floor{\funcstop/\numstates}$ follows a geometric distribution with parameter $\revprobsymb$ (unknown) due to \ref{asmp:IRL_finite_Search_2}.

Consider now the quantity $|\occagentemp{\agent}-\occagent{\agent}|$. Define $\hat{\E}_{\stoptime_{\agent}}\{\funcstop/\numstates\}=\frac{\sum_{\trial=1}^{\numtrials_{\state,\agent}} \floor{\funcstop_{\state,\agent,\trial}/\numstates} }{\numtrials_{\state,\agent}}$, the sample average of the normalized stopping time $\floor{\funcstop/\numstates}$ computed from $\datafinsearch{\trialset}$. Then,
\begin{align*}
    &|\occagentemp{\agent}-\occagent{\agent}|= |\E_{\stoptime_{\agent}}\{\floor{\funcstop/\numstates}~|\state\} + r_{\agent}(\state,\action) - \hat{\E}_{\stoptime_{\agent}}\{\floor{\funcstop/\numstates}~|\state\} - r_{\agent}(\state,\action)|\\
    =& \left|\frac{1}{\revprobsymb} -\hat{\E}_{\stoptime_{\agent}}\{\floor{\funcstop/\numstates}~|\state\} \right|~(\text{equal for all $a$ for a fixed $\state$}).
\end{align*}
$\hat{\E}_{\stoptime}\{\floor{\funcstop/\numstates}|\state\}$ is an unbiased estimator of $\E_{\stoptime}\{\floor{\funcstop/\numstates}|\state\}$ with 
variance $(1-\revprobsymb)/\numtrials_{\state,\agent}\revprobsymb^2$. Using Chebyshev's inequality for random variables with finite variance to bound $|\occagentemp{\agent}-\occagent{\agent}|$ for  fixed $\action,\state,\agent$, the following inequality results
\begin{equation}
    \prob\left( |\occagentemp{\agent}-\occagent{\agent}|\leq \eps \right)\geq 1- \frac{(1-\revprobsymb)}{\numtrials_{\state,\agent}(\revprobsymb\eps)^2}
\end{equation}
For any set of positive reals $\{\eps_{\state,\agent},\state\in\stateset,\agent\in\agentset\}$ s.t. $|\occagent{\agent}-\occagentemp{\agent}|\leq \eps_{\state,\agent}$ and  $(\numactions\sum_{\state,\agent}\eps_{\state,\agent}^2)\leq \eps_{\max}^2$, we have 
\begin{align}
    &\prob(\sum_{\state,\agent}|\occagentemp{\agent}-\occagent{\agent}|^2\leq \eps_{\max}^2)\geq \prod_{\state,\agent}1-\frac{(1-\revprobsymb)}{\numtrials_{\state,\agent}(\revprobsymb\eps_{\state,\agent})^2}\nonumber\\
    \geq& 1- \sum_{\state,\agent} \frac{(1-\revprobsymb)}{\numtrials_{\state,\agent}(\revprobsymb\eps_{\state,\agent})^2}\geq 1- \sum_{\state,\agent} \frac{(1-\revprobprime)}{\numtrials_{\state,\agent}(\revprobprime\eps_{\state,\agent})^2}.\label{eqn:finite_search_bound}
\end{align}
Since $\frac{(1-\revprobsymb)}{\numtrials_{\state,\agent}(\revprobsymb\eps_{\state,\agent})^2}$ is decreasing in $\eps_{\state,\agent}$, the tightest lower bound is achieved for the above inequality when $\sum_{\state,\agent}\eps_{\state,\agent}^2= \eps_{\max}^2$ and is the solution to the following constrained optimization problem.
\begin{align}
    \min_{\{\eps_{\state,\agent},\state\in\stateset,\agent\in\agentset\}} \sum_{\state,\agent} \frac{ (1-\revprobprime)}{\numtrials_{\state,\agent}(\revprobprime\eps_{\state,\agent})^2}\text{ s.t. } \numactions\sum_{\state,\agent}\eps_{\state,\agent}^2 = \eps_{\max}^2.\label{eqn:opt_finite_search}
\end{align}
Moreover, since the objective function in (\ref{eqn:opt_finite_search}) is convex in $\eps_{\state,\agent}^2$ and constraint is affine in $\eps_{\state,\agent}^2$,  the method of Lagrange multipliers~\citep{BD04} yields necessary and sufficient conditions for an optimal solution to the above optimization problem if the solution obtained is positive for all $\state\in\stateset,\agent\in\agentset$. The optimal value of $\eps_{\state,\agent}^2=\frac{\eps_{\max}^2 \numtrials_{\state,\agent}^{-1/2}}{\numactions\sum_{\state,\agent}\numtrials_{\state,\agent}^{-1/2}}>0$ and thus minimizes the objective function in (\ref{eqn:opt_finite_search}). Plugging this value in (\ref{eqn:finite_search_bound}) and setting $\eps_{\max}^2=\epsilon$ yields the bound in the RHS of (\ref{eqn:search_bound}) and completes the proof for Proposition~\ref{prop:Search_bound}. 

To obtain the error bounds (\ref{eqn:finite_result_Search_1}), note that setting 
$\eps=\epsone{\datafinsearch{\trialset}}$ in (\ref{eqn:search_bound}) and subtracting the objective function from 1 bounds from below the Type-I error probability (see Sec.\,\ref{sec:finite_sample_main_result_stopping} for a detailed explanation). Similarly, setting $\eps=\epstwo{\datafinsearch{\trialset}}$ in (\ref{eqn:search_bound}) and subtracting the objective function from $1$ bounds from below the Type-II error probability of the IRL detector which completes the proof. \hfill\qedsymbol

\section{Context. IRL For Predicting YouTube Commenting Behavior}\label{appdx:youtube}
\blue{Our previous work~\citep{HKP20} analyzes YouTube user engagement  from a behavioral economics viewpoint. Although we use the same dataset for our numerical experiments in this paper, we emphasize that the IRL approach in this paper is new and differs from \cite{HKP20} as:\\
(1) In \cite{HKP20}, we check if YouTube engagement is consistent with rationally inattentive utility maximization behavior~\cite{CM15}, a {\em static} decision model studied widely in behavioral and information economics. In comparison, our aim here is to test if the YouTube dataset satisfies Bayes optimal stopping, a {\em dynamic} decision model.\\
(2) The inference algorithms in \cite{HKP20} considers {\em pairs} of video categories to reconstruct the underlying utility function of the YouTube user. In this paper, our IRL approach considers {\em all $18$ YouTube video categories  (described in Sec.\,\ref{sec:YT-parameters} below) simultaneously in the feasibility test} for reconstructing the underlying stopping costs of the YouTube  user, and hence, fully exploits the diversity in engagement behavior.\\
(3) In \cite{HKP20}, we perform a naive prediction analysis of YouTube user engagement using a {\em maximum a posteriori} (MAP) approach. In this paper, we predict the distribution of user engagement behavior via two representative point estimates of the recovered stopping costs and show the statistical similarity of the predicted distribution to the true engagement distribution.}

\noindent \blue{ {\em YouTube user engagement and Bayesian stopping.}\\
YouTube is a social multimedia platform where human users interact with video content on YouTube channels by posting comments and rating videos. Empirical studies (\cite{KH17,HG17,ABCH15,AK17}) show that the comments and ratings from users are  influenced by the thumbnail, title, category, and perceived popularity of each video. Models for human decision making in the context of online multimedia platforms have been studied extensively in the literature. Two widely-used classes of models that motivate us to understand YouTube user engagement from the lens of Bayesian stopping are `parallel constraint satisfaction models' and `evidence accumulation models'. {\em Parallel constraint satisfaction models}~(\cite{GCK08,MC89}) assume that information is screened sequentially to highlight salient alternatives and final choice is made when the decision maker reaches sufficient internal coherence. {\em Evidence accumulation models}~(\cite{KRA10,RC04}) model consumers' attention by drift-diffusion models that accumulate evidence based on whether they are fixating their gaze on either the product or its price. The decision is taken when any of the alternatives' evidence threshold level is achieved. }

\blue{Both classes of models described above have one aspect in common - the decision maker makes a final choice {\em after} sequentially accumulating information, and naturally fits our Bayesian stopping time framework. In terms of YouTube webpage parameters, we hypothesize the YouTube user is a Bayesian agent that sequentially consumes webpage cues such as thumbnail, title and perceived popularity and incurs a cost of attention, followed by engaging on the YouTube platform and incurring a terminal cost. Our IRL aim in this section is to identify using the YouTube dataset, if YouTube users engage `optimally' in a Bayesian stopping sense.}

\bibliography{iqd}
\end{document}

%% file: macrodefs.tex
\newcommand{\testfinite}{\operatorname{Test}_{\operatorname{IRL}}}
\newcommand{\posintegers}{\mathbb{Z}^+}
\newcommand{\norm}[1]{\lVert#1\rVert} 
\newcommand{\stateset}{\ensuremath{\mathcal{X}}} 
\newcommand{\dataset}{\ensuremath{\mathcal{D}}} 
\newcommand{\actionset}{\ensuremath{\mathcal{A}}} 
\newcommand{\prior}{\ensuremath{\pi_0}} 

\newcommand{\reals}{\ensuremath{\mathbb{R}}} 
\newcommand{\grosspayoff}{\ensuremath{G}}
\newcommand{\grosspayoffsurr}{\ensuremath{\tilde{G}}}
\newcommand{\E}{\mathbb{E}}
\newcommand{\prob}{\mathbb{P}}

\newcommand{\IRLoutput}{\operatorname{IRL}}

\newcommand{\action}{a} 
\newcommand{\state}{x}
\newcommand{\tstate}{\state^o}
\newcommand{\obs}{y}
\newcommand{\obsfict}{\tilde{\obs}_{\belief}}
\newcommand{\obsset}{\mathcal{Y}}
\newcommand{\qedsymbol}{$\blacksquare$}

\newcommand{\SHT}{\operatorname{SHT}}
\newcommand{\Search}{\operatorname{Search}}

\newcommand{\agentset}{\mathcal{M}}

\newcommand{\oprob}{\ensuremath{B}}

\newcommand{\datageneric}{\mathcal{D}}
\newcommand{\datainf}{\mathcal{D}_{M}}
\newcommand{\trialset}{\mathcal{\numtrials}}
\newcommand{\datafin}[1]{\widehat{\mathcal{D}}_{M}(#1)}

\newcommand{\numtrials}{K}
\newcommand{\mc}{\bar{L}}
\newcommand{\numactions}{A}
\newcommand{\runcost}{\mathcal{\sumruncostsymbol}}
\newcommand{\runcostinst}{c}
\newcommand{\runcostinstvec}{\bar{\runcostinst}}
\newcommand{\sumruncostsymbol}{C}
\newcommand{\sumruncostagent}[1]{\sumruncostsymbol_{#1}}
\newcommand{\utilityagent}[1]{\utilitysymbol_{#1}(\state,\action)}
\newcommand{\utilityagentaltact}[1]{\utilitysymbol_{#1}(\state,\action')}

\newcommand{\utilitysymbol}{s}
\newcommand{\utilitysymbolcaps}{S}
\newcommand{\utilvec}{\boldsymbol{\utilitysymbolcaps}}
\newcommand{\utilityvec}{\bar{\utilitysymbol}}
\newcommand{\actselectagent}[1]{p_{#1}(\action|\state)}
\newcommand{\actselectagentemp}[1]{\hat{p}_{#1}(\action|\state)}

\newcommand{\actselectset}{\boldsymbol{p}}

\newcommand{\actselectagentempcand}[1]{\hat{p}'_{#1}(\action|\state)}
\newcommand{\actselectagentempcandsymb}[1]{\hat{p}'_{#1}}

\newcommand{\obslikesymbol}{p}
\newcommand{\actselect}{p(\action|\state)}
\newcommand{\actselectemp}{\hat{p}(\action|\state)}
\newcommand{\actselectempsymb}[1]{\hat{p}_{#1}}

\newcommand{\obslike}{\obslikesymbol(\obs|\tstate)}
\newcommand{\obslikestoppingagent}[1]{p(\obs_{1:\funcstop(\stoptime_{#1})}|x)}

\newcommand{\funcstop}{\tau}
\newcommand{\stoptime}{\mu}

\newcommand{\belief}{\pi}

\newcommand{\trial}{k}
\newcommand{\timeinst}{t}
\newcommand{\agent}{m}
\newcommand{\numagents}{M}

\newcommand{\tstatedata}{\tstate_{\trial,\agent}}
\newcommand{\actiondata}{\action_{\trial,\agent}}
\newcommand{\stoptimedata}{\funcstop_{\trial}(\stoptime_{\agent})}

\newcommand{\agenttwo}{n}
\newcommand{\stoptimeset}{\boldsymbol{\stoptime}}

\newcommand{\utilitysymbolset}{\boldsymbol{\utilitysymbol}}
\newcommand{\utilitysymbolagent}[1]{\utilitysymbol_{#1}}
\newcommand{\stoptuple}{\Xi}
\newcommand{\optstoptuple}{\stoptuple_{opt}}

\newcommand{\netobjfun}{J}
\newcommand{\sumcost}{\operatorname{SUMCOST}}
\newcommand{\numstates}{X}
\newcommand{\fictobslike}{\alpha}
\newcommand{\fictobslikeagent}[1]{\fictobslike_{#1}(\obsfict|\state)}
\newcommand{\fictobsset}{\obsset_{\belief}}
\newcommand{\revpos}[1]{p_{#1}(\state|\action)}

\definecolor{pigment}{rgb}{0.2, 0.2, 0.6}
\definecolor{darkblue}{rgb}{0.0, 0.0, 0.55}
\newcommand{\blue}[1]{\textcolor{black}{#1}}

\newcommand{\epsone}[1]{\varepsilon_1(#1)}
\newcommand{\epstwo}[1]{\varepsilon_2(#1)}

\newcommand{\revprob}{\revprobsymb(\action)}
\newcommand{\revprobsymb}{\alpha}
\newcommand{\optsearchtuple}{\stoptuple_{opt}}
\newcommand{\eps}{\varepsilon}

\newcommand{\datainfsearch}{\mathcal{D}_{\numagents}(\Search)}
\newcommand{\datafinsearch}[1]{\widehat{\mathcal{D}}_{\numagents}(#1)}

\newcommand{\searchcostsymbol}{l}

\newcommand{\actiontwo}{b}

\newcommand{\searchcostagent}[1]{\searchcostsymbol_{#1}(\action)}
\newcommand{\revprobset}{\boldsymbol{\revprobsymb}}
\newcommand{\occsymb}{g}
\newcommand{\occagent}[1]{\occsymb_{#1}(\action,\state)}
\newcommand{\occagentemp}[1]{\hat{\occsymb}_{#1}(\action,\state)}
\newcommand{\occagentempsymb}[1]{\hat{\occsymb}_{#1}}

\newcommand{\revprobprime}{\revprobsymb^{\ast}}
\newcommand{\occagentcand}[1]{\hat{\occsymb}_{#1}'(\action,\state)}
\newcommand{\occagentcandsymb}[1]{\hat{\occsymb}_{#1}'}
\newcommand{\cdfsymb}{F}
\newcommand{\actcdfagent}[1]{\cdfsymb_{#1}(\action|\state)}
\newcommand{\actcdfagentemp}[1]{\hat{\cdfsymb}_{#1}(\action|\state)}
\newcommand{\sumruncostagentemp}[1]{\hat{\sumruncostsymbol}_{#1}}

\newcommand{\optstoptimeset}{\stoptimeset^\ast}
\newcommand{\optstoptime}{\stoptime^\ast}